# Real-Time Multilingual
# Sign Language Processing

Amit Moryossef

Department of Computer Science

Ph.D. Thesis

Submitted to the Senate of Bar-Ilan University

Ramat Gan, Israel                         December 2023

To those who always believed in me,
and to the ones I have loved and lost.

# Acknowledgements

First and foremost, my sincerest appreciation to my supervisor, Yoav Goldberg, for his guidance, support, and encouragement. The autonomy he granted and his candid feedback significantly contributed to my growth as a scientist.

This thesis owes its very existence to Maayan Gazuli (Hazut), who first introduced me to the world of sign language. It was a casual conversation on a car ride that sparked my interest and set me on this journey. I am immensely thankful for Maayan's unwitting role as the catalyst of this entire endeavor.

My deepest gratitude goes to Valerie Sutton, the inventor of Sutton Sign-Writing, for her pioneering work in creating a linguistic transcription system for signed languages. Her system served as a vital tool in the research and analysis underpinning this thesis and profoundly influenced its outcome. Without her contributions, this thesis would not have been feasible. I am honored to build upon her groundbreaking work and grateful for her commitment to making signed languages accessible to all.

I also extend my thanks to Ioannis Tsochantaridis for hosting me as an intern at Google during the COVID-19 pandemic, and to Sarah Ebling, Annette Rios, and Mathias Müller for their unwavering support and for hosting me at the University of Zurich. Their kindness and hospitality played a crucial role in maintaining my mental well-being, for which I am deeply grateful.

To my Bar-Ilan University labmates, particularly Yanai Elazar, Valentina Pyatkin, Hila Gonen, Shauli Ravfogel, and Amir David Nissan Cohen, I thank you for being my academic social circle during this challenging period.

To the Computational Linguistics Ph.D. students at the University of Zurich, especially Noëmi Aepli, Tannon Kew, and Anastassia Shaitarova, your warm welcome and inclusive atmosphere made my time there not only productive but also enjoyable. You've made a meaningful impact on both my professional development and personal well-being, and for that, I am genuinely grateful.

Special recognition goes to the team at the University of Hamburg Institute of German Sign Language and Communication of the Deaf for their openness and outreach. Thanks to Thomas Hanke for granting me access to the DGS corpus and providing continual support; Maria Kopf for teaching me how to read and write HamNoSys; and Annika Herrmann for filling gaps in my knowledge with interesting linguistic information. Your collective wisdom has widely contributed to this thesis.

A big shoutout to the students I supervised during this thesis — Zifan Jiang from the University of Zurich and Rotem Shalev Arkushin from Reichman University. Your willingness to heed my advice, coupled with the courage to challenge my ideas when they were off the mark, made this journey intellectually enriching for all parties involved. Thank you for keeping me on my toes.

I would also like to acknowledge Rebecca Norton and Hannah Neuser for their encouragement and support, which gave me the motivation and confidence I needed to persevere through difficult times and complete this work.

Lastly, I wish to express my heartfelt gratitude to my family for their love, support, and encouragement throughout my journey.

# Funding

This work was financially supported by **Bar-Ilan University** (the European Research Council (ERC) under the European Union's Horizon 2020 research and innovation programme, grant agreement No. 802774 (iEXTRACT)); by **the University of Zurich** (the European Union's Horizon 2020 research and innovation programme (grant number 101016982, EASIER), and the Swiss Innovation Agency (Innosuisse) flagship IICT (PFFS-21-47)); and by **Google**.

# Preface

## Publications

Portions of this thesis are joint work and have been published elsewhere.

- Chapter 2, "Background" includes material published as "Sign Language Processing" in `https://sign-language-processing.github.io`, (Moryossef and Goldberg, 2021).

- Section 3.1, "pose-format: Library for Viewing, Augmenting, and Handling *.pose* Files" appeared in `https://github.com/sign-language-processing/pose`, (Moryossef et al., 2021a).

- Section 3.2, "Sign Language Datasets" appeared in `https://github.com/sign-language-processing/datasets`, (Moryossef and Müller, 2021).

- Section 3.3, "3D Hand Pose Benchmark" appeared in `https://github.com/sign-language-processing/3d-hands-benchmark`, (Moryossef, 2022).

- Section 5.1, "Activity Detection" includes material that appeared in SLRTP 2020: The Sign Language Recognition, Translation & Production Workshop, in "Real-Time Sign-Language Detection using Human Pose Estimation" (Moryossef et al., 2020).

- Section 5.2, "Isolated Recognition" includes material that appeared in Proceedings of the IEEE/CVF Conference on Computer Vision and Pattern

Recognition, in "Evaluating the immediate applicability of pose estimation for sign language recognition" (Moryossef et al., 2021b).

- Section 5.3, "Gloss Translation" includes material that appeared in Proceedings of the 1st International Workshop on Automatic Translation for Signed and Spoken Languages (AT4SSL 2021), in "Data Augmentation for Sign Language Gloss Translation" (Moryossef et al., 2021c).

- Section 6.1, "Segmentation" includes material that appeared in Proceedings of the Findings of the 2023 Conference on Empirical Methods in Natural Language Processing, in "Linguistically Motivated Sign Language Segmentation" (Moryossef et al., 2023a).

- Section 6.3, "Translation" includes material under submission to Proceedings of the 2024 Joint International Conference on Computational Linguistics, Language Resources and Evaluation, in "SignBank+: Multilingual Sign Language Translation Dataset" (Moryossef and Jiang, 2023).

- Section 7.1, "Baseline" includes material that appeared in Proceedings of the 2nd International Workshop on Automatic Translation for Signed and Spoken Languages (AT4SSL 2023), in "An Open-Source Gloss-Based Baseline for Spoken to Signed Language Translation" (Moryossef et al., 2023b). This work was further presented at the Meeting of Computational Linguistics in The Netherlands (CLIN 2023).

- Chapter 8, "Sign Language Translation Application" includes material that appeared in "sign.mt: Real-Time Multilingual Sign Language Translation Application" (Moryossef, 2023c). ⋆ *Outstanding demo paper award* ⋆

- Chapter 9, "Implications for Spoken Languages" includes material that appeared in "Addressing the Blind Spots in Spoken Language Processing" (Moryossef, 2023b).

## Highlighted Collaborations

- "Including Signed Languages in Natural Language Processing" appeared in Proceedings of the 59th Annual Meeting of the Association for Computational Linguistics (ACL 2021, Yin et al. (2021)). ⋆ *Best paper award* ⋆

- "Machine Translation between Spoken Languages and Signed Languages Represented in SignWriting" appeared in Proceedings of the 17th Conference of the European Chapter of the Association for Computational Linguistics (EACL 2023, Jiang et al. (2023a)).

- "Ham2Pose: Animating Sign Language Notation into Pose Sequences" appeared in Proceedings of the IEEE/CVF Conference on Computer Vision and Pattern Recognition 2023 (CVPR 2023, Arkushin et al. (2023)).

- "Considerations for meaningful sign language machine translation based on glosses" appeared in Proceedings of the 61th Annual Meeting of the Association for Computational Linguistics (ACL 2023, Müller et al. (2023)). ⋆ *Outstanding paper award* ⋆

# Contents

# Abstract

Signed languages serve as a vital means of communication for millions of deaf and hard-of-hearing individuals worldwide. Utilizing a visual-gestural modality, they convey complex linguistic structures through manual articulations combined with non-manual elements like facial expressions and body movement. Despite their linguistic richness and cultural importance, signed languages have often been marginalized by the latest advances in text-centric artificial intelligence technologies, such as Machine Translation and Large Language Models. This marginalization restricts access to these technologies for a significant population, leaving them behind in the rapid advancements in language-based AI.

Sign Language Processing (SLP) is an interdisciplinary field comprised of Natural Language Processing (NLP) and Computer Vision. It is focused on the computational understanding, translation, and production of signed languages. Traditional approaches have often been constrained by the use of gloss-based systems that are both language-specific and inadequate for capturing the multidimensional nature of sign language. These limitations have hindered the development of technology capable of processing signed languages effectively.

This thesis aims to revolutionize the field of SLP by proposing a simple paradigm that can bridge this existing technological gap. We propose the use of SignWiring, a universal sign language transcription notation system, to serve as an intermediary link between the visual-gestural modality of signed languages and text-based linguistic representations.

Unlike gloss-based approaches, our paradigm using SignWriting is designed to accurately capture the multidimensional and language-independent aspects of signed languages. This allows for the creation of a unified and scalable framework that can accommodate the rich linguistic diversity found in various signed languages across the globe.

We contribute foundational libraries and resources to the SLP community, thereby setting the stage for a more in-depth exploration of the tasks of sign language translation and production. These tasks encompass the translation of sign language from video to spoken language text and vice versa. Through empirical evaluations, we establish the efficacy of our transcription method as a pivot for enabling faster, more targeted research, that can lead to more natural and accurate translations across a range of languages.

Our paradigm establishes a clear boundary between NLP and Computer Vision within the broader context of SLP. This division mirrors the existing separation between NLP and Signal Processing in the realm of spoken language technologies. By doing so, we open the door for more specialized research efforts in each sub-discipline, thereby enriching the ecosystem of technologies and methodologies available for SLP.

The universal nature of our transcription-based paradigm also paves the way for real-time, multilingual applications in SLP, thereby offering a more inclusive and accessible approach to language technology. This is a significant step toward universal accessibility, enabling a wider reach of AI-driven language technologies to include the deaf and hard-of-hearing community.

In summary, this thesis presents a new approach to Sign Language Processing, one that aims to set a new standard for inclusive, real-time, and multilingual language technologies. By bridging the existing gap between text-centric AI and the visual-gestural world of signed languages, we substantially contribute toward making language-based AI universally accessible.

# Part I

# Introduction to Sign Language Processing

"we do these things not because they are easy,
but because we thought they were going to be."

— Pinboard, The Programmers' Credo

# Chapter 1

# Introduction

Signed languages (also known as sign languages) are languages that use the visual-gestural modality to convey meaning through manual articulations in combination with non-manual elements like the face and body. They serve as the primary means of communication for numerous deaf and hard-of-hearing individuals. Similar to spoken languages, signed languages are natural languages governed by a set of linguistic rules (Sandler and Lillo-Martin, 2006), both emerging through an abstract, protracted aging process and evolving without deliberate meticulous planning. Signed languages are not universal or mutually intelligible, despite often having striking similarities among them. They are also distinct from spoken languages—i.e., American Sign Language (ASL) is not a visual form of English but its own unique language.

Sign Language Processing (Bragg et al., 2019; Yin et al., 2021) is an emerging field of artificial intelligence concerned with the automatic processing and analysis of sign language content. While research has focused more on the visual aspects of signed languages, it is a subfield of both Natural Language Processing (NLP) and Computer Vision (CV). Challenges in sign language processing often include machine translation of sign language videos into spoken language text (sign language translation), from spoken language text (sign language production), or sign language recognition for sign language understanding.

Unfortunately, the latest advances in language-based artificial intelligence,

like machine translation and personal assistants, expect a spoken language input (text or transcribed speech), excluding around 200 to 300 different signed languages (United Nations, 2022) and up to 70 million deaf people (World Health Organization, 2021; World Federation of the Deaf, 2022).

Throughout history, Deaf communities fought for the right to learn and use signed languages and for the public recognition of signed languages as legitimate ones. Indeed, signed languages are sophisticated communication modalities, at least as capable as spoken languages in all aspects, both linguistic and social. However, in a predominantly oral society, deaf people are constantly encouraged to use spoken languages through lip-reading or text-based communication. The exclusion of signed languages from modern language technologies further suppresses signing in favor of spoken languages. This exclusion disregards the preferences of the Deaf communities who strongly prefer to communicate in signed languages both online and for in-person day-to-day interactions, among themselves and when interacting with spoken language communities (Padden and Humphries, 1988; Glickman and Hall, 2018). Thus, it is essential to make signed languages accessible.

To date, a large amount of research on Sign Language Processing (SLP) has been focused on the visual aspect of signed languages, led by the Computer Vision (CV) community, with little NLP involvement. This focus is not unreasonable, given that a decade ago, we lacked adequate CV tools to process videos for further linguistic analyses. However, similar to spoken languages, signed languages are fully-fledged systems exhibiting all the fundamental characteristics of natural languages, and existing SLP techniques do not adequately address or leverage the linguistic structure of signed languages. Signed languages introduce novel challenges for NLP due to their visual-gestural modality, simultaneity, spatial coherence, and lack of written form. The lack of a written form makes the spoken language processing pipelines - which often start with audio transcription before processing - incompatible with signed languages, forcing researchers to work directly on the raw video signal.

Furthermore, SLP is not only intellectually appealing but also an important research area with significant potential to benefit signing communities. Bene-ficial applications enabled by signed language technologies include improved documentation of endangered sign languages; educational tools for sign language learners; tools for query and retrieval of information from signed language videos; personal assistants that react to signed languages; real-time automatic sign language interpretations; and more. Needless to say, in addressing this research area, researchers should work *alongside* and *under the direction of* deaf communities, and to benefit the signing communities' interest above all (Harris et al., 2009).

## 1.1   (Brief) History of Signed Languages and Deaf Culture

Throughout modern history, spoken languages were dominant, so much so that signed languages struggled to be recognized as languages in their own right, and educators developed misconceptions that signed language acquisition might hinder the development of speech skills. For example, in 1880, a large international conference of deaf educators called the "Second International Congress on Education of the Deaf" banned teaching signed languages, favoring speech therapy instead. It was not until the seminal work on American Sign Language (ASL) by Stokoe Jr (1960) that signed languages started gaining recognition as natural, independent, and well-defined languages, which inspired other researchers to further explore signed languages as a research area. Nevertheless, antiquated attitudes that placed less importance on signed languages continue to inflict harm and subject many to linguistic neglect (Humphries et al., 2016). Several studies have shown that deaf children raised solely with spoken languages do not gain enough access to a first language during their critical period of language acquisition (Murray et al., 2020). This language deprivation can lead to life-long consequences on the cognitive, linguistic, socioemotional, and academic development of the deaf (Hall et al., 2017).

Signed languages are the primary languages of communication for the Deaf[1]

---

[1]When capitalized, "Deaf" refers to a community of deaf people who share a language and

and are at the heart of Deaf communities. In the past, the failure to recognize signed languages as fully-fledged natural language systems in their own right has had detrimental effects, and in an increasingly digitized world, NLP research should strive to enable a world in which all people, including the Deaf, have access to languages that fit their lived experience.

## 1.2   Thesis Overview

The dissertation asserts that for progress to be made in the area of sign language processing, it is vital to adopt a written phonetic lexical representation for sign language, as an intermediary stage for any subsequent tasks.

This section provides an overview of the thesis structure and content, as well as the contributions made to the field of Sign Language Processing.

**Part I**   Introduces the field of Sign Language Processing to the reader.

1. Chapter 1 situates this field within the broader context of artificial intelligence and machine learning research, outlines the main problem addressed, and introduces signed languages and Deaf culture.

2. Chapter 2 introduces the linguistic aspects of signed languages as natural languages, explains and demonstrates their representation, overviews the existing types of available resources, and covers the various tasks involved. Some of these tasks are further described in background sections within relevant chapters.

3. Chapter 3 discusses some of the preliminary work carried out in preparation for this thesis, with a focus on libraries designed to be widely used in sign language processing research.

---

a culture, whereas the lowercase "deaf" refers to the audiological condition of not hearing. We follow the more recent convention of abandoning a distinction between "Deaf" and "deaf", using the latter term also to refer to (deaf) members of the sign language community (Napier and Leeson, 2016; Kusters et al., 2017).

**Part II**   Explores the use of universal phonetic sign language written forms as an intermediate representation for downstream sign language processing tasks, such as translation and production.

1. Chapter 4 introduces the idea of utilizing written sign language representations as an intermediate stage.

2. Chapter 5 discusses some of the preliminary work carried out at the beginning of this thesis, with a focus on relevant work published in sign language research venues.

3. Chapter 6 explores the application of *lexical* written representations as an intermediate phase for translation signed-to-spoken language translation.

4. Chapter 7 compliments Chapter 6 by examining the usability of such an intermediate representation in the production process and assesses its effectiveness in comparison to the use of *semantic* forms.

**Part III**   Wraps up the thesis by integrating the different components into a working demonstration, and discussing key insights and contributions.

1. Chapter 8 presents a sign language translation application that can translate from and to multiple signed languages in real-time and offline. It delves into the engineering, design, and development of this application.

2. Chapter 9 discusses the implications of the findings in this thesis as they relate to the field of Spoken Language Processing, and proposes a way for the two fields to benefit from each other.

3. Chapter 10 concludes the thesis by summarizing the main findings and contributions, outlining potential avenues for future research directions in Sign Language Processing, and highlighting the potential impact of this work on the deaf community.

# Chapter 2

# Background (Moryossef and Goldberg, 2021)

## 2.1  (Brief) Sign Language Linguistics Overview

Signed languages consist of phonological, morphological, syntactic, and semantic levels of structure that fulfill the same social, cognitive, and communicative purposes as other natural languages. While spoken languages primarily channel the oral-auditory modality, signed languages use the visual-gestural modality, relying on the signer's face, hands, body, and space around them to create distinctions in meaning. We present the linguistic features of signed languages[1] that researchers must consider during their modeling.

**Phonology**  Signs are composed of minimal units that combine manual features such as hand configuration, palm orientation, placement, contact, path movement, local movement, as well as non-manual features including eye aperture, head movement, and torso positioning (Liddell and Johnson, 1989; Johnson and Liddell, 2011; Brentari, 2011; Sandler, 2012). Not all possible phonemes are realized in both signed and spoken languages, and inventories of two languages' phonemes/features may not overlap completely. Different languages

---

[1]We mainly refer to ASL, where most research has been conducted, but not exclusively.

are also subject to rules for the allowed combinations of features.

**Simultaneity** Though an ASL sign takes about twice as long to produce than an English word, the rates of transmission of information between the two languages are similar (Bellugi and Fischer, 1972). One way signed languages compensate for the slower production rate of signs is through simultaneity: Signed languages use multiple visual cues to convey different information simultaneously (Sandler, 2012). For example, the signer may produce the sign for "cup" on one hand while simultaneously pointing to the actual cup with the other to express "that cup." Similarly to tone in spoken languages, the face and torso can convey additional affective information (Liddell et al., 2003; Johnston and Schembri, 2007). Facial expressions can modify adjectives, adverbs, and verbs; a head shake can negate a phrase or sentence; gaze can help indicate referents.

**Referencing** The signer can introduce referents in discourse either by pointing to their actual locations in space or by assigning a region in the signing space to a non-present referent and by pointing to this region to refer to it (Rathmann and Mathur, 2011; Schembri et al., 2018). Signers can also establish relations between referents grounded in signing space by using directional signs or embodying the referents using body shift or eye gaze (Dudis, 2004; Liddell and Metzger, 1998). Spatial referencing also impact morphology when the directionality of a verb depends on the location of the reference to its subject and/or object (de Beuzeville, 2008; Fenlon et al., 2018): For example, a directional verb can move from its subject's location and end at its object's location. While the relation between referents and verbs in spoken language is more arbitrary, referent relations are usually grounded in signed languages. The visual space is heavily exploited to make referencing clear.

Another way anaphoric entities are referenced in sign language is by using classifiers or depicting signs that help describe the characteristics of the referent (Supalla, 1986; Wilcox and Hafer, 2004; Roy, 2011). Classifiers are typically one-handed signs that do not have a particular location or movement assigned to them, or derive features from meaningful discourse (Liddell et al., 2003), so they can be used to convey how the referent relates to other entities, describe its movement, and give more details. For example, to tell about a car swerving and crashing, one might use the hand classifier for a vehicle, move it to indicate swerving, and crash it with another entity in space.

To quote someone other than oneself, signers perform *role shift* (Cormier et al., 2015), where they may physically shift in space to mark the distinction and take on some characteristics of the people they represent. For example, to recount a dialogue between a taller and a shorter person, the signer may shift to one side and look up when taking the shorter person's role, shift to the other side and look down when taking the taller person's role.

**Fingerspelling**   results from language contact between a signed language and a surrounding spoken language (Battison, 1978; Wilcox, 1992; Brentari and Padden, 2001; Patrie and Johnson, 2011). A set of manual gestures corresponds with a written orthography or phonetic system. This phenomenon, found in most signed languages, is often used to indicate names, places, or new concepts from the spoken language, but has often become integrated into the language as another linguistic strategy (Padden, 1998; Montemurro and Brentari, 2018).

## 2.2   Representation

Representation is a significant challenge for SLP. Unlike spoken languages, signed languages have no widely adopted written form. As signed languages are conveyed through the visual-gestural modality, video recording is the most straightforward way to capture them. However, as videos include more information than needed for modeling and are expensive to record, store, and transmit, a lower-dimensional representation has been sought after.

Figure 2.1 illustrates various signed language representations. In this demonstration, we deconstruct the video into its individual frames to exemplify the alignment of the annotations between the video and representations.

Figure 2.1: Representations of an American Sign Language phrase with video frames, pose estimations, SignWriting, HamNoSys and glosses. English translation: "What is your name?" (Yin et al., 2021)

**Videos** are the most straightforward representation of a signed language and can amply incorporate the information conveyed through signing. One major drawback of using videos is their high dimensionality: They usually include more information than needed for modeling and are expensive to store, transmit, and encode. As facial features are essential in sign, anonymizing raw videos remains an open problem, limiting the possibility of making these videos publicly available (Isard, 2020).

**Skeletal Poses** reduce the visual cues in videos to skeleton-like wireframes or meshes, representing the location of joints. This representation has been extensively used in computer vision, estimating human pose from video data, and determining the spatial configuration of the body at each point in time. Although high-quality pose estimation can be achieved using motion capture equipment, such methods are often expensive and intrusive. As a result, estimating pose from videos has become the preferred method in recent years (Pishchulin et al., 2012; Chen et al., 2017; Cao et al., 2019; Güler et al., 2018). Compared to video representations, accurate skeletal poses have a lower complexity and provide a semi-anonymized representation of the human body, while observing relatively low information loss. However, they remain a continuous, multidimensional representation that is not adapted to most NLP models.

**Written notation systems** represent signs as discrete visual features. Some systems are written linearly, and others use graphemes in two dimensions. While various universal (Sutton, 1990; Prillwitz and Zienert, 1990) and language-specific notation systems (Stokoe Jr, 1960; Kakumasu, 1968; Bergman, 1977) have been proposed, no writing system has been adopted widely by any sign language community, and the lack of standards hinders the exchange and unification of resources and applications between projects. Figure 2.1 depicts two universal notation systems: SignWriting (Sutton, 1990), a two-dimensional pictographic system, and HamNoSys (Prillwitz and Zienert, 1990), a linear stream of graphemes designed to be machine-readable.

**Glosses** are the transcription of signed languages sign-by-sign, with each sign having a unique semantic identifier. While various sign language corpus projects have provided guidelines for gloss annotation (Mesch and Wallin, 2015; Johnston and De Beuzeville, 2016; Konrad et al., 2018), a standardized gloss annotation protocol has yet to be established. Gloss IDs offer a more precise approach by assigning unique identifiers to each form of a sign, not just its meaning, capturing unspecified phonetic variations. Linear gloss annotations have been criticized for their imprecise representation of signed language. These annotations fail to capture all the information expressed simultaneously through different cues, such as body posture, eye gaze, or spatial relations, leading to a loss of information that can significantly affect downstream performance on SLP tasks (Yin and Read, 2020a; Müller et al., 2023).

Table 2.1 additionally exemplifies the various representations for more isolated signs. For this example, we use SignWriting as the notation system. Note that the same sign might have two unrelated glosses, and the same gloss might have multiple valid spoken language translations.

| Video | Pose Estimation | SignWriting | Gloss | English Translation |
|---|---|---|---|---|
| 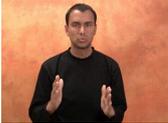 | 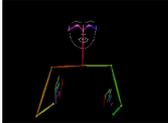 | 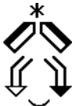 | HOUSE | House |
| 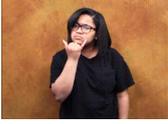 | 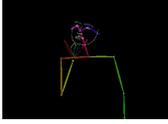 | 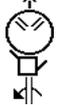 | WRONG-WHAT | What's the matter? What's wrong? |
| 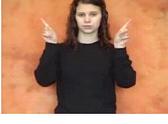 | 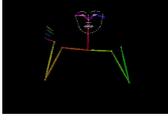 | 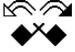 | DIFFERENT BUT | Different But |

Table 2.1: Example of isolated signs represented in multiple ways.

## 2.3   Annotation Tools

**ELAN - EUDICO Linguistic Annotator (Wittenburg et al., 2006)**   is an annotation tool for audio and video recordings. With ELAN, a user can add an unlimited number of textual annotations to audio and/or video recordings. An annotation can be a sentence, word, gloss, comment, translation, or description of any feature observed in the media. Annotations can be created on multiple layers, called tiers, which can be hierarchically interconnected. An annotation can either be time-aligned to the media or refer to other existing annotations. The content of annotations consists of Unicode text, and annotation documents are stored in an XML format (EAF). ELAN is open source (GPLv3), and installation is available for Windows, macOS, and Linux. PyMPI (Lubbers and Torreira, 2013) allows for simple python interaction with Elan files.

**iLex (Hanke, 2002)**   is a tool for sign language lexicography and corpus analysis, that combines features found in empirical sign language lexicography and sign language discourse transcription. It supports the user in integrated lexicon building while working on the transcription of a corpus and offers several unique features considered essential due to the specific nature of signed languages. iLex binaries are available for macOS.

**SignStream (Neidle et al., 2001)**   is a tool for linguistic annotations and computer vision research on visual-gestural language data SignStream installation is available for macOS and is distributed under an MIT license.

**Anvil - The Video Annotation Research Tool (Kipp, 2001)**   is a free video annotation tool, offering multi-layered annotation based on a user-defined coding scheme. In Anvil, the annotator can see color-coded elements on multiple tracks in time alignment. Some special features are cross-level links, non-temporal objects, timepoint tracks, coding agreement analysis, 3D viewing of motion capture data and a project tool for managing whole corpora of annotation files. Anvil installation is available for Windows, macOS, and Linux.

## 2.4 Resources

Signed language resources come in multiple forms, such as bilingual dictionaries, fingerspelling and isolated sign corpora, and continuous sign corpora. Each has its own limitations, but they are all essential for translation and production in signed languages.

**Bilingual dictionaries** for signed language (Mesch and Wallin, 2012; Fenlon et al., 2015; Crasborn et al., 2016; Gutierrez-Sigut et al., 2016) map a spoken language word or short phrase to a signed language video. One notable dictionary, SpreadTheSign[2] is a parallel dictionary containing around 25,000 words with 42 different spoken-signed language pairs and more than 600,000 videos in total. Unfortunately, while dictionaries lexically map between languages, they do not demonstrate the grammar or the usage of signs in context.

**Fingerspelling corpora** usually consist of videos of words borrowed from spoken languages that are signed letter-by-letter. They can be synthetically created (Dreuw et al., 2006) or mined from online resources (Shi et al., 2018, Shi et al. (2019)). However, they only capture one aspect of signed languages.

**Isolated sign corpora** are collections of a limited vocabulary (20-1000 signs) of annotated single signs. They are synthesized (Ebling et al., 2018; Huang et al., 2018; Sincan and Keles, 2020; Hassan et al., 2020) or mined from online resources (Vaezi Joze and Koller, 2019; Li et al., 2020), and can be used for isolated sign language recognition or contrastive analysis of minimal signing pairs (Imashev et al., 2020). However, like dictionaries, they do not describe relations between signs, nor do they capture coarticulation during the signing.

**Continuous sign corpora** contain parallel sequences of signs and spoken language. Available continuous sign corpora are extremely limited, containing 4-6 orders of magnitude fewer sentence pairs than similar corpora for spoken language machine translation (Arivazhagan et al., 2019). Moreover, while automatic speech recognition (ASR) datasets contain up to 50,000 hours of recordings (Pratap et al., 2020), the most extensive continuous sign language corpus contains only 1,150 hours, and only 50 of them are publicly available (Hanke et al., 2020). These datasets are usually synthesized (Databases, 2007; Crasborn and Zwitserlood, 2008; Ko et al., 2019; Hanke et al., 2020) or recorded in studio conditions (Forster et al., 2014, Camgöz et al. (2018)), which does not account for noise in real-life conditions. Moreover, some contain signed interpretations of spoken language rather than naturally-produced signs, which may not accurately represent native signing since translation is now a part of the discourse.

---

[2] https://www.spreadthesign.com/

**Availability**   Unlike the vast amount and diversity of available spoken language resources that allow various applications, sign language resources are scarce and, currently only support translation and production. Unfortunately, most of the sign language corpora discussed in the literature are either not available for use or available under heavy restrictions and licensing terms. Furthermore, sign language data is especially challenging to anonymize due to the importance of facial and other physical features in signing videos, limiting its open distribution. Developing anonymization with minimal information loss or accurate anonymous representations is a promising research direction.

## 2.4.1   Real-World Data Collection

Data is essential to develop any data-driven technology, and current efforts in SLP are often limited by the lack of adequate data. We discuss the considerations to keep in mind when building datasets, the challenges of collecting such data, and directions to facilitate data collection.

**What is Good Sign Language Data?**   For SLP models to be deployable, they must be developed using data representing the real world accurately. What constitutes an ideal sign language dataset is an open question; we propose the following requirements: (1) a broad domain; (2) sufficient data and vocabulary size; (3) real-world conditions; (4) naturally produced signs; (5) a diverse signer demographic; (6) native signers; and when applicable, (7) dense annotations.

To illustrate the importance of data quality during modeling, Yin et al. (2021) first take as an example a current benchmark for SLP, the RWTH-PHOENIX-Weather 2014T dataset (Camgöz et al., 2018) of German Sign Language, that does not meet most of the above criteria: it is restricted to the weather domain (1); contains only around 8K segments with 1K unique signs (2); filmed in studio conditions (3); interpreted from German utterances (4); and signed by nine Caucasian interpreters (5,6). Although this dataset successfully addressed data scarcity issues at the time and successfully rendered results and fueled competitive research, it does not accurately represent signed languages in the real world. On the other hand, the Public DGS Corpus (Hanke et al., 2020) is an open-domain (1) dataset consisting of 50 hours of natural signing (4) by 330 native signers from various regions in Germany (5,6), annotated with glosses, and German translations (7), meeting all but two requirements we suggest.

They train a gloss-to-text sign language translation transformer (Yin and Read, 2020a) on both datasets. On RWTH-PHOENIX-Weather 2014T, they obtain **22.17** BLEU on testing; on the Public DGS Corpus, they obtain a mere **3.2** BLEU. Although Transformers achieve encouraging results on RWTH-PHOENIX-Weather 2014T (Saunders et al., 2020c; Camgöz et al., 2020a), they fail on more realistic, open-domain data. These results reveal that, for real-world applications, we need more data to train such models. At the same time, available data is severely limited in size; less data-hungry and more linguistically-informed approaches may be more suitable. This experiment reveals how it is crucial to use data that accurately represent the complexity and diversity of signed languages to precisely assess what types of methods are suitable and how well our models would deploy to the real world.

**Challenges of Data Collection**   Collecting and annotating signed data in line with the ideal requires more resources than speech or text data, taking up to 600 minutes per minute of an annotated signed language video (Hanke et al., 2020). Moreover, annotation usually requires specific knowledge and skills, which makes recruiting or training qualified annotators challenging. Additionally, there is little existing sign language data in the wild openly licensed for use, especially from native signers that are not interpretations of speech. Therefore, data collection often requires significant efforts and costs of on-site recording.

**Automating Annotation**    One helpful research direction for collecting more data that enables the development of deployable SLP models is creating tools that can simplify or automate parts of the collection and annotation process. One of the most significant bottlenecks in obtaining more adequate sign language data is the time and scarcity of experts required to perform annotation. Therefore, tools that perform automatic parsing, detection of sign boundaries, extraction of articulatory features, suggestions for lexical annotations, and allow parts of the annotation process to be crowdsourced to non-experts, to name a few, have a high potential to accelerate the availability of good data.

### 2.4.2   Practicing Deaf Collaboration

Finally, when working with signed languages, it is vital to keep in mind *who* this technology should benefit and *what* they need. Researchers in SLP should acknowledge that signed languages belong to the Deaf community and avoid exploiting their language as a commodity (Bird, 2020).

**Solving Real Needs**    Many efforts in SLP have developed intrusive methods (e.g., requiring signers to wear special gloves), which are often rejected by signing communities and therefore have limited real-world value. Such efforts are often marketed to perform "sign language translation" when they, in fact, only identify fingerspelling or recognize a minimal set of isolated signs at best. These approaches oversimplify the rich grammar of signed languages, promote the misconception that signs are solely expressed through the hands, and are considered by the Deaf community as a manifestation of audism, where it is the signers who must make the extra effort to wear additional sensors to be un-derstood by non-signers (Erard, 2017). To avoid such mistakes, we encourage close Deaf involvement throughout the research process to ensure that we direct our efforts toward applications that will be adopted by signers and do not make false assumptions about signed languages or the needs of signing communities.

**Building Collaboration**   Deaf collaborations and leadership are essential for developing sign language technologies to ensure they address the community's needs and will be adopted, not relying on misconceptions or inaccuracies about signed language (Harris et al., 2009; Kusters et al., 2017). Hearing researchers cannot relate to the deaf experience or fully understand the context in which the tools being developed would be used, nor can they speak for the deaf. Therefore, we encourage creating a long-term collaborative environment between sign language researchers and users so that deaf users can identify meaningful challenges and provide insights on the considerations to take while researchers cater to the signers' needs as the field evolves. We also recommend reaching out to signing communities for reviewing papers on signed languages to ensure an adequate evaluation of this type of research results published at academic venues. There are several ways to connect with Deaf communities for collaboration: one can seek deaf students in their local community, reach out to schools for the deaf, contact deaf linguists, join a network of researchers of sign-related technologies[3], or participate in deaf-led projects.

## 2.5   Tasks

So far, the computer vision community has primarily led the SLP research to focus on processing the visual features in sign language videos. As a result, current SLP methods do not fully address the linguistic complexity of signed languages. We survey common SLP tasks and current methods' limitations, drawing on signed languages' linguistic theories.

Sign Language Translation (SLT) commonly refers to the translation of signed language to spoken language (De Coster et al., 2022; Müller et al., 2022). Sign Language Production is the reverse process of producing a sign language video from spoken language text. Sign Language Recognition (SLR) (Adaloglou et al., 2020) detects and labels signs from isolated (Imashev et al., 2020; Sincan and Keles, 2020) or continuous (Cui et al., 2017; Camgöz et al., 2018, 2020b) sign language videos.

---

[3]https://www.crest-network.com/

**Language Agnostic Tasks**                              **Language Specific Tasks**

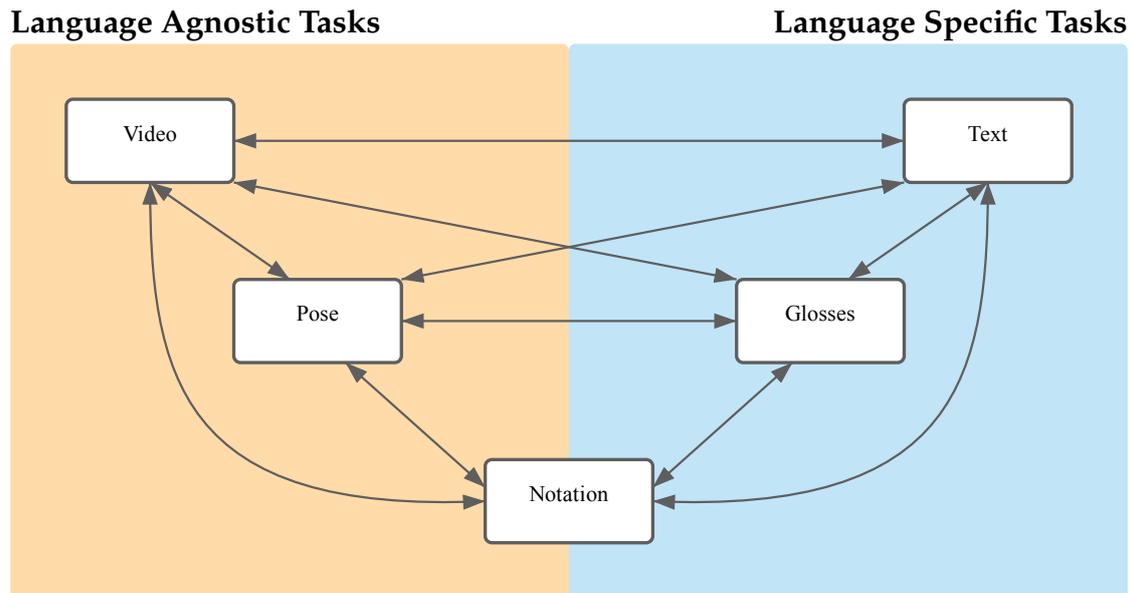

Figure 2.2: The various sign language processing tasks, visualized as a fully connected graph.

Figure 2.2 presents a fully connected graph where each node is a single data representation, and each directed edge represents the task of converting one data representation to another. We split the graph into two parts - language agnostic, and language specific.

- Every edge to the left, on the orange background, represents a task in computer vision. These tasks are inherently language-agnostic; thus, they generalize between signed languages.

- Every edge to the right, on the blue background, represents a task in natural language processing. These tasks are sign language-specific, requiring a specific sign language lexicon or spoken language tokens.

- Every edge crossing between backgrounds represents a task requiring a combination of computer vision and natural language processing.

This graph conceptually defines 20 different tasks, with varying amounts of previous research. Every path between two nodes that goes from left-to-right or right-to-left can be a valid pipeline of tasks. However, in this thesis, we make the case that the most valid paths are paths with tasks that do not cross between the modalities, and instead go through the 'Notation' representation.

The necessary background information for various tasks is disseminated throughout this thesis, specifically in Part II. Following is a guide to direct you to where you can find an overview of each task's background.

For insights into Sign Language Activity Detection, refer to Section 5.1. An overview of Pose Estimation (*Video-to-Pose*), as well as Isolated Sign Recognition (*Video-to-Gloss* and *Pose-to-Gloss*), is available in Section 5.2. Section 5.3 delves into previous work related to *Gloss-to-Text* translation. To understand Sign Language Sign and Phrase Segmentation, please consult Section 6.1. Section 6.2 provides the background for *Pose-to-Notation* sign language transcription. For a comprehensive background on both *Text-to-Notation* and *Notation-to-Text* sign language translation, visit Section 6.3. To explore the *Text-to-Gloss* and *Gloss-to-Pose* tasks, refer to Section 7.1. Section 7.3 offers insights into *Text-to-Pose* and *Notation-to-Pose* tasks. Lastly, background information on the *Pose-to-Video* animation task can be found in Section 7.4.

### 2.5.1 Fingerspelling

Fingerspelling is spelling a word letter-by-letter, borrowing from the spoken language alphabet (Battison, 1978; Wilcox, 1992; Brentari and Padden, 2001; Patrie and Johnson, 2011). This phenomenon, found in most signed languages, often occurs when there is no previously agreed-upon sign for a concept, like in technical language, colloquial conversations involving names, conversations involving current events, emphatic forms, and the context of code-switching between the signed language and the corresponding spoken language (Padden, 1998; Montemurro and Brentari, 2018). The relative amount of fingerspelling varies between signed languages, and for American Sign Language (ASL), accounts for 12-35% of the signed content (Padden and Gunsauls, 2003).

Patrie and Johnson (2011) described the following terminology to describe three different forms of fingerspelling:

- **Careful**—slower spelling where each letter pose is clearly formed.

- **Rapid**—quick spelling where letters are often not completed and contain remnants of other letters in the word.

- **Lexicalized**—a sign produced by often using no more than two letter-hand-shapes (Battison, 1978). For example, lexicalized `ALL` uses `A` and `L`, lexicalized `BUZZ` uses `B` and `Z`, etc...

**Recognition**

Fingerspelling recognition, a sub-task of sign language recognition, is the task of recognizing fingerspelled words from a sign language video.

Shi et al. (2018) introduced a large dataset available for American Sign Language fingerspelling recognition. This dataset includes both the "careful" and "rapid" forms of fingerspelling collected from naturally occurring videos "in the wild", which are more challenging than studio conditions. They trained a baseline model to take a sequence of images cropped around the signing hand. They found that using CTC outperformed autoregressive decoding, but that both achieved poor recognition rates (35-41% character level accuracy) compared to human performance (around 82%).

In follow-up work, Shi et al. (2019) collected nearly an order-of-magnitude larger dataset and designed a new recognition model. Instead of detecting the signing hand, they detected the face and cropped a large area around it. Then, they performed an iterative process of zooming in to the hand using visual attention to retain sufficient information in high resolution of the hand. Finally, like their previous work, they encoded the image hand crops sequence and used a CTC to obtain the frame labels. They showed that this method outperformed their original "hand crop" method by 4% and that they could achieve up to 62.3% character-level accuracy using the additional data collected. Looking through this dataset, we note that the videos in the dataset were taken from longer videos, and as they were cut, they did not retain the signing before the fingerspelling. This context relates to language modeling, where at first, one fingerspells a word carefully, and when repeating it, might fingerspell it rapidly, but the interlocutors can infer they are fingerspelling the same word.

**Production**

Fingerspelling production, a sub-task of sign language production, is the task of producing a fingerspelling video for words.

In its basic form, "careful" fingerspelling production can be trivially solved using pre-defined letter handshapes interpolation. Adeline (2013) demonstrated this approach for American Sign Language and English fingerspelling. They rigged a hand armature for each letter in the English alphabet ($N = 26$) and generated all ($N^2 = 676$) transitions between every two letters using interpolation or manual animation. Then, to fingerspell entire words, they chain pairs of letter transitions. For example, for the word "CHLOE", they would chain the following transitions sequentially: `#C CH HL LO OE E#`.

However, to produce life-like animations, one must also consider the rhythm and speed of holding letters, and transitioning between letters, as those can affect how intelligible fingerspelling motions are to an interlocutor (Wilcox (1992)). Wheatland et al. (2016) analyzed both "careful" and "rapid" fingerspelling videos for these features. They found that for both forms of fingerspelling, on average, the longer the word, the shorter the transition and hold time. Furthermore, they found that less time is spent on middle letters on average, and the last letter is held on average for longer than the other letters in the word. Finally, they used this information to construct an animation system using letter pose interpolation and controlled the timing using a data-driven statistical model.

# Chapter 3

# Preliminary Work (Libraries)

In the emerging field of sign language processing, standardized approaches for dataset distribution, loading, visualization, and augmentation remain undeveloped. Addressing this gap, we have prioritized two pivotal open-source projects. First, *pose-format* (§3.1) serves as a comprehensive library tailored to sign language processing, enabling users to easily read, write, visualize, and augment pose sequences. The second, *sign-language-datasets* (§3.2), facilitates seamless integration of datasets regardless of their distribution or format, by providing swift disk-mapped storing and loading of datasets. In tandem, *pose-format* and *sign-language-datasets* have become the foundational tools for contemporary sign language processing research. We further introduce *3d-hands-benchmark* (§3.3), a tool to evaluate the consistency and usefulness of hand pose estimation models in the context of sign language hand shapes.

## 3.1   pose-format: Library for Viewing, Augmenting, and Handling *.pose* Files (Moryossef et al., 2021a)

Managing and analyzing pose data is a complex task, with challenges ranging from handling diverse file structures and data types to facilitating effective data manipulations such as normalization and augmentation. This paper presents `pose-format`, a comprehensive toolkit designed to address these challenges by providing a unified, flexible, and easy-to-use interface. The library includes a specialized file format that encapsulates various types of pose data, accommodating multiple individuals and an indefinite number of time frames, thus proving its utility for both image and video data. Furthermore, it offers seamless integration with popular numerical libraries such as NumPy, PyTorch, and TensorFlow, thereby enabling robust machine-learning applications. Through benchmarking, we demonstrate that our `.pose` file format offers vastly superior performance against prevalent formats like OpenPose, with added advantages like self-contained pose specification. Additionally, the library includes features for data normalization, augmentation, and easy-to-use visualization capabilities, both in Python and Browser environments. `pose-format` emerges as a one-stop solution, streamlining the complexities of pose data management and analysis.

### 3.1.1   Introduction

Working with pose data introduces many complexities, from the diversity in file structures to the variety of data types that need to be accommodated. Developers and researchers often find themselves juggling numerous data manipulation tasks such as normalization, augmentation, and visualization. In addition to these challenges, pose data itself can be inherently multidimensional, frequently encompassing multiple individuals and varying time frames. This creates an intricate ecosystem of variables that can be challenging to manage and analyze effectively, which is important in fields like Sign Language Processing.

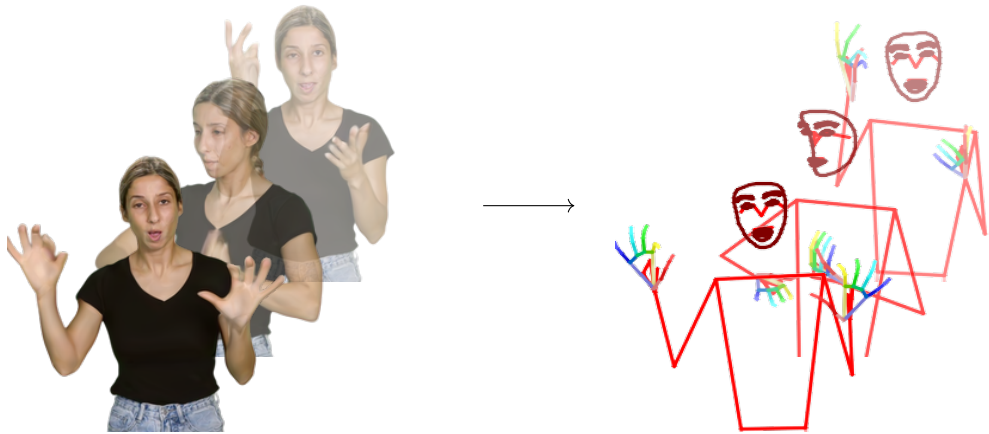

Figure 3.1: Examples of human skeletal poses extracted from a sign language video sequence.

To overcome these complexities, we designed `pose-format`, a comprehensive toolkit to alleviate these challenges by offering a unified, flexible, and easy-to-use interface for managing and analyzing pose data. Designed with versatility in mind, the library includes a specialized file format that accommodates an array of pose types, multiple people, and an indefinite number of time frames, making it highly adaptable for both video and single-frame data. Users can effortlessly import `.pose` files and perform a range of manipulations such as data normalization and augmentation. The library also integrates seamlessly with popular numerical libraries like NumPy (Harris et al., 2020), PyTorch (Paszke et al., 2019b), and TensorFlow (Abadi et al., 2015), allowing for additional computational flexibility for machine learning. With features for easy visualization and compatibility with other popular pose data formats like OpenPose (Cao et al., 2019) and MediaPipe Holistic (Grishchenko and Bazarevsky, 2020), `pose-format` emerges as a one-stop solution for working with poses.

### 3.1.2 Background

In the context of our library, a *pose* consists of *keypoints*, which are 2D or 3D coordinates marking points of interest usually on a human body in image or video frames (Figure 3.1). Systems like OpenPose and MediaPipe Holistic are prominent for pose estimation but have differing methodologies and keypoint configurations. OpenPose, for instance, uses a classification objective and outputs 135 or 137 keypoints with 2D coordinates. MediaPipe Holistic employs a regression objective, estimating 543 keypoints with 3D coordinates.

Keypoints are hierarchically organized, often attached to larger body components like `LEFT HAND` or `FACE`. Moreover, models implicitly define which keypoints are connected, forming an underlying graph structure. Confidence metrics vary across systems. OpenPose assigns a confidence score to each classification, while MediaPipe Holistic only predicts the likelihood of each `BODY` keypoint's presence in the original image.

The utility of human pose estimation (Zheng et al., 2023) spans various fields such as human-computer interaction, motion capture, motion analysis, and mixed reality, with specialized applications like automatic sign language processing (Moryossef et al., 2021b; Müller et al., 2022).

### 3.1.3 Justification

`pose-format` addresses a void in the ecosystem by delivering a uniform layer of abstraction over disparate pose estimation system outputs, such as OpenPose and MediaPipe Holistic. The necessity for this unified interface arises from three primary factors: inconsistent standards, inadequacy of existing libraries, and performance bottlenecks.

**Inconsistent Standards** As delineated in §3.1.2, there are competing pose estimation systems, each adhering to its own data storage and representation scheme. This inconsistency impedes interoperability between systems and makes the data hard to share or transition across different platforms. We remedy this by standardizing how pose data is managed, making it simpler to operate with multiple systems, switch between them, or even disseminate pose data.

**Limitations of Existing Libraries**  Current libraries focus extensively on low-level operations, lacking the higher-level abstractions that can expedite routine tasks. For instance, in the absence of our toolkit, users have to micromanage array values, discerning between coordinates and confidence scores or handling missing keypoints. Such intricacies detract from productivity and introduce unnecessary complexity. Our library fills this gap by offering user-friendly operations, many of which are indispensable for machine learning research, such as frame rate interpolation, rotation, scaling, frame dropout, or converting the underlying data into tensors of a specific machine learning library.

**Efficiency**  As demonstrated in §3.1.7, prevailing methods for pose data management suffer from performance limitations in both speed and storage. These inefficiencies create bottlenecks for data-intensive tasks, especially those prevalent in machine learning pipelines. `pose-format` offers optimized data storage and retrieval, mitigating these inefficiencies.

### 3.1.4   Format Specification

The core of the `pose-format` library is its specialized file format that accommodates a wide range of scenarios. This unique format enables the storing of multidimensional data capturing various pose types, multiple individuals, and an indefinite number of time frames. Currently, at version 0.1, the file format is bifurcated into two components: the `Header` and the `Body`.

**Header (`PoseHeader`)**

The header contains meta information that defines the overall structure of the pose data. This information is useful for visualization and code readability. Specifically, it includes:

- **(`float32`) Version** - The version of the file format.

- **(`uint16[3]`) Dimensions** - Width, height, and depth specifications.

- **(uint16) Number of Components** - The number of pose components.

**Component Details**    Each component includes its (`string`) name, (`string`) format, and the (`uint16`) number of points, (`uint16`) limbs, and (`uint16`) colors it contains.

- (`string[]`) Names of points.

- (`uint16[2][]`) Start and end indices of limbs.

- (`uint16[3][]`) Points color RGB values.

**Body (`PoseBody`)**

The body of the file comprises the actual pose data and includes the following:

- **(uint16) FPS** - The frame rate of the pose.

- **(uint16) Number of frames** - ~~deprecated due to challenges for longer pose sequences.~~

- **(uint16) Number of People** - The number of people included in every frame.

- **(float[][][][]) Data** - The coordinate of every point for every person in every frame.

- **(float[][][]) Confidence** - The confidence for every point for every person in every frame.

This format's granularity and modularity make it aptly suited for a wide range of applications, from simple image-based pose representation to more complex video analysis tasks. By leveraging this detailed yet flexible format, the `pose-format` library ensures ease of use without sacrificing the intricacies that pose data often necessitates.

**v0.1 Limitations**

While the `pose-format` library has been designed to cater to a wide array of needs, there are some limitations and criticisms in the current file format that users should be aware of:

- **FPS Representation**: The FPS is stored as `uint16`, which does not allow for floating-point values.

- **Number of Frames**: The number of frames is also restricted to `uint16`, which limits the frame count to 65,535. The current workaround calculates the number of frames based on the file size, which introduces computational overhead.

- **Pose Data Precision**: The pose data utilizes 32-bit floating-point values for storage. However, 16-bit numbers could be sufficient for many applications. Support for both types would improve memory efficiency.

- **Confidence Precision**: Similar to the pose data, the confidence metrics are stored as 32-bit floating-point numbers. A 16-bit representation would be more than sufficient for most practical purposes.

### 3.1.5   Data Manipulations

One of the key advantages of this toolkit is its robust support for various data manipulation tasks, which are crucial for the preprocessing and augmentation of pose data. This section elaborates on how the library facilitates operations such as normalization and augmentation.

**Normalization**   Normalization is a crucial step to make pose data scale and translation invariant, thereby improving the effectiveness of downstream tasks like training machine learning models. Our toolkit offers a simple yet powerful interface to normalize pose data. For example, when dealing with human body poses, we can specify the names of the left and right shoulders, and the skeleton will be scaled such that the mean distance between the shoulders is equal to 1, and the center point lies on $(0, 0)$. If we deal with 3D poses, we can also specify a plane by naming three points, to make sure they always fall on the same plane. These normalizations try to remove the effect of camera angles and distance from the subject on the resulting video.

**Augmentation**  Data augmentation is a technique to artificially increase the size and diversity of your training dataset by applying various transformations. In the context of pose data, these can include affine transformations such as translation, scaling, reflection, rotation, and shear, interpolation of frames at variable speeds, noise, and dropout, to name a few. The `pose-format` toolkit provides built-in functions to perform these augmentations effortlessly. You can either apply these transformations individually or chain them together to create a complex augmentation pipeline, thereby enhancing the library's adaptability to various project needs.

**Integration with Numerical Libraries**  Data manipulations are seamlessly integrated with popular numerical libraries like NumPy, PyTorch, and TensorFlow. This facilitates easy data flow between data manipulation and machine learning models, reducing the friction in the data science pipeline. It allows loading and augmenting the data in a framework of your choosing, minimizing the number of memory copy operations.

### 3.1.6 Visualization

The ability to visualize pose data is crucial for understanding its characteristics, debugging algorithms, and even for presentation purposes.

**Python**  In Python, users can use the `PoseVisualizer` for different visualization tasks, such as visualizing the pose by itself as a sequence of still images, a video, a GIF, with the background being either a fixed color or overlayed on another video. An example of visualizing the pose as a video would be:

```python
from pose_format import Pose
from pose_format.pose_visualizer import PoseVisualizer

with open("example.pose", "rb") as f:
    pose = Pose.read(f.read())

v = PoseVisualizer(pose)

v.save_video("example.mp4", v.draw())
```

**Browser**   Additionally, for web-based applications or quick interactive viewing, poses can be visualized in the browser. Unlike the rasterized Python visualization, the web-based visualization is vectorized and thus more suitable for client-facing applications.

```html
<script type="module"
src="https://unpkg.com/pose-viewer@latest
/dist/pose-viewer/pose-viewer.esm.js" />

<pose-format src="example.pose" />
```

### 3.1.7   Benchmarking

To evaluate our custom file format, we benchmarked it against OpenPose, a prevalent standard. Metrics of interest were read speed and file size. We obtained OpenPose data from a single video in the Public DGS Corpus (Hanke et al., 2020, DOI: /10.25592/dgs.corpus-3.0-text-1413451-11105600-11163240). Their format employs a monolithic JSON file to store frames, diverging from the common one-file-per-frame approach.

To gauge reading performance, we measured OpenPose's JSON load time in isolation, sidestepping tensor conversion. For our format, we include both full-file reads and body-only tensor reads where we skip loading the pose header, and only load the tensor of coordinates and confidences.

| # Frames | OpenPose | | pose-format | | |
|---|---|---|---|---|---|
| | Size | Speed | Size | Speed | Speed (Body) |
| 1 | 3.9 KB | 37.4 µs ± 600 ns | 3.6 KB | 535 µs ± 66.1 µs | 61.7 µs ± 6.94 µs |
| 10 | 38 KB | 364 µs ± 6.9 µs | 18 KB | 490 µs ± 63.8 µs | 57.9 µs ± 2.56 µs |
| 100 | 388 KB | 3.75 ms ± 113 µs | 163 KB | 415 µs ± 49.7 µs | 72.4 µs ± 4.87 µs |
| 1,000 | 3.9 MB | 43.1 ms ± 704 µs | 1.6 MB | 658 µs ± 110 µs | 228 µs ± 9.09 µs |
| 10,000 | 39 MB | 439 ms ± 29.5 ms | 16 MB | 2.72 ms ± 110 µs | 2.71 ms ± 245 µs |

Table 3.1: Benchmarking `pose-format` against `OpenPose` from the Public DGS Corpus. We compare both the resulting file size, and file read speed. *Speed (Body)* measures loading the *.pose* files data only, without metadata.

**Quantitative Edge**   Table 3.1 reveals we achieve up to a 60% file size reduction and outpace OpenPose in read speed by a staggering $162\times$, thereby obliterating any machine learning bottlenecks.

**Qualitative Edge**   Our `pose-format` packs all pose data into a singular, robust file, avoiding the file fragmentation issues seen in OpenPose. Moreover, our header encodes pose structure, obviating the need for hard-coded interpretation logic and boosting both portability and usability.

In summation, `pose-format` offers superior performance across key metrics, making it a compelling alternative for pose data management.

### 3.1.8   Community Contributions

Our library is fully open-source, and released under an MIT License. We welcome contributions from the community of any kind, and we encourage collaboration. Source code and bug reporting are available at **https://github.com/sign-language-processing/pose**.

## 3.2 Sign Language Datasets (Moryossef and Müller, 2021)

Following our discussion on the variety of resources essential for sign language processing in Section 2, and underscoring the importance of collecting real-world data in collaboration with the deaf community, we now shift our focus to the acquisition and processing of these datasets.

First, we compile an inventory of currently available datasets, detailing the types of annotations each dataset contains when that information is available. This list is presented in Table 3.2. In addition to its inclusion in this thesis, the inventory is maintained as a living document available at https://research.sign.mt/#resources. This document is updated periodically to incorporate newly released datasets, serving as a continually evolving resource for researchers in the field.

At present, there exists no standardized approach or consensus for downloading and loading sign language datasets, and as such, evaluation of these datasets is scarce. To address this gap, we have streamlined the process of dataset loading through the use of Tensorflow Datasets (authors, 2019). This utility enables researchers to effortlessly load datasets — regardless of their size — with a single command, thereby facilitating comparability across different studies. We offer access to these datasets through our custom-built library, Sign Language Datasets (Moryossef and Müller, 2021).

To demonstrate the use of our library, we show how a dataset can be easily loaded in one line of Python. In this example, we load the ASLG-PC12 dataset with its default configuration:

```python
import tensorflow_datasets as tfds
import sign_language_datasets.datasets

aslg_pc12 = tfds.load("aslg_pc12")
```

We also support loading datasets with custom configuration. For example,

we want to load the videos resized to 256*x*256 as tensors at 12 frames-per-second, and also load MediaPipe Holistic poses for each of the videoss.

```python
# Loading a dataset with custom configuration
from sign_language_datasets.datasets.config
    import SignDatasetConfig

config = SignDatasetConfig(
    name="videos_and_poses256x256:12",
    # Specific version
    version="3.0.0",
    # Download, and load dataset videos
    include_video=True,
    # Load videos at constant, 12 fps
    fps=12,
    # Convert videos to a constant resolution, 256x256
    resolution=(256, 256),
    # Download and load Holistic pose estimation
    include_pose="holistic")

rwth_phoenix2014_t = tfds.load(
    name='rwth_phoenix2014_t',
    builder_kwargs=dict(config=config))
```

We follow a unified interface when possible, making attributes the same and comparable between datasets:

```python
{
    "id": tfds.features.Text(),
    "signer": tfds.features.Text() | tf.int32,
    "video": tfds.features.Video(
        shape=(None, HEIGHT, WIDTH, 3)),
    "depth_video": tfds.features.Video(
```

```
                shape=(None, HEIGHT, WIDTH, 1)),
    "fps": tf.int32,
    "pose": {
        "data": tfds.features.Tensor(
            shape=(None, 1, POINTS, CHANNELS),
            dtype=tf.float32),
        "conf": tfds.features.Tensor(
            shape=(None, 1, POINTS),
            dtype=tf.float32)
    },
    "gloss": tfds.features.Text(),
    "text": tfds.features.Text()
}
```

🎥 Video | 👋 Pose | 👄 Mouthing | ✏️ Notation | 🗒️ Gloss | 📒 Text | 🔊 Speech | 💾 Save using *sign-language-datasets*

| Dataset | Publication | Language | Features | #Signs | #Samples | #Signers | License |
|---|---|---|---|---|---|---|---|
| ASL-100-RGBD | Hassan et al. (2020) | American | 🎥👋📒 | 100 | 4,150 Tokens | 22 | Authorized Academics |
| ASL-Homework-RGBD | Hassan et al. (2022) | American | 🎥👋📒 | | 935 | 45 | Authorized Academics |
| ASLG-PC12 💾 | Othman and Jemni (2012) | American (Synthetic) | 🗒️📒 | | > 100,000,000 Sentences | N/A | Sample Available (1, 2) |
| ASLLVD | Athitsos et al. (2008) | American | - | 3,000 | 12,000 Samples | 4 | Attribution |
| ATIS | Bungeroth et al. (2008) | Multilingual | - | 292 | 595 Sentences | | |
| AUSLAN | Johnston (2010) | Australian | - | | 1,100 Videos | 100 | |
| AUTSL 💾 | Sincan and Keles (2020) | Turkish | 🎥👋📒 | 226 | 36,302 Samples | 43 | Codalab |
| BosphorusSign | Camgöz et al. (2016) | Turkish | - | 636 | 24,161 Samples | 6 | Not Published |
| BSL Corpus 💾 | Schembri et al. (2013) | British | - | | 40,000 Lexical Items | 249 | Partially Restricted |
| CDPSL | Łacheta and Rutkowski (2014) | Polish | 🎥👋✏️📒 | | 300 hours | | |
| ChicagoFSWild 💾 | Shi et al. (2018) | American | 🎥🗒️ | 26 | 7,304 Sequences | 160 | Public |
| ChicagoFSWild+ 💾 | Shi et al. (2019) | American | 🎥🗒️ | 26 | 55,232 Sequences | 260 | Public |
| CopyCat | Zafrulla et al. (2010) | American | - | 22 | 420 Phrases | 5 | |
| Corpus NGT 💾 | Crasborn and Zwitserlood (2008) | Netherlands | - | | 15 Hours | 92 | CC BY-NC-SA 3.0 NL |
| DEVISIGN | Chai et al. (2014) | Chinese | - | 2,000 | 24,000 Samples | 8 | Research purpose |
| Dicta-Sign 💾 | Matthes et al. (2012) | Multilingual | - | | 6-8 Hours (/Participant) | 16-18 /Language | |
| How2Sign 💾 | Duarte et al. (2020) | American | 🎥👋🗒️📒🔊 | 16,000 | 79 hours (35,000 sentences) | 11 | CC BY-NC 4.0 |
| K-RSL | Imashev et al. (2020) | Kazakh-Russian | 🎥👋🗒️ | 600 | 28,250 Videos | 10 | Attribution |
| KETI | Ko et al. (2019) | Korean | 🎥👋🗒️📒 | 524 | 14,672 Videos | 14 | |

| Dataset | Publication | Language | Features | #Signs | #Samples | #Signers | License |
|---|---|---|---|---|---|---|---|
| KRSL-OnlineSchool | Mukushev et al. (2022) | Kazakh-Russian | 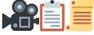 | | 890 Hours (1M sentences) | 7 | |
| LSE-SIGN | Gutierrez-Sigut et al. (2016) | Spanish | - | 2,400 | 2,400 Samples | 2 | Custom |
| MS-ASL | Vaezi Joze and Koller (2019) | American | - | 1,000 | 25,000 (25 hours) | 200 | Public |
| NCSLGR  | Databases (2007) | American | 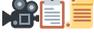 | | 1,875 sentences | 4 | - |
| Public DGS Corpus  | Prillwitz et al. (2008) | German | 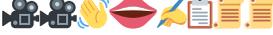 | | 50 Hours | 330 | Custom |
| RVL-SLLL ASL | Martínez et al. (2002) | American | - | 104 | 2,576 Videos | 14 | Research Attribution |
| RWTH Fingerspelling | Dreuw et al. (2006) | German | 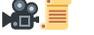 | 35 | 1,400 single-char videos | 20 | |
| RWTH-BOSTON-104 | Dreuw et al. (2008) | American | 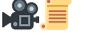 | 104 | 201 Sentences | 3 | |
| RWTH-PHOENIX-Weather T  | Forster et al. (2014);Camgöz et al. (2018) | German | 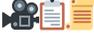 | 1,231 | 8,257 Sentences | 9 | CC BY-NC-SA 3.0 |
| S-pot | Viitaniemi et al. (2014) | Finnish | - | 1,211 | 5,539 Videos | 5 | Permission |
| Sign2MINT  | 2021 | German | 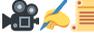 | 740 | 1135 | | CC BY-NC-SA 3.0 DE |
| SignBank  | | Multilingual | 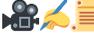 | | 222148 | | |
| SIGNOR | Vintar et al. (2012) | Slovene | 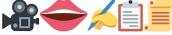 | | | 80 | |
| SIGNUM | Von Agris and Kraiss (2007) | German | - | 450 | 15,600 Sequences | 20 | |
| SMILE | Ebling et al. (2018) | Swiss-German | - | 100 | 9,000 Samples | 30 | Custom |
| SSL Corpus | Öqvist et al. (2020) | Swedish | 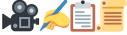 | | | | |
| SSL Lexicon | Mesch and Wallin (2012) | Swedish | 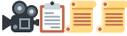 | 20,000 | | | CC BY-NC-SA 2.5 SE |
| Video-Based CSL | Huang et al. (2018) | Chinese | - | 500 | 125,000 Videos | 50 | Research Attribution |
| WLASL  | Li et al. (2020) | American | 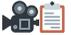 | 2,000 | | 100 | C-UDA 1.0 |

Table 3.2: Curated list of some of the existing datasets, showcasing the mostly restrictive licensing, and small sizes.

## 3.3   3D Hand Pose Benchmark (Moryossef, 2022)

### 3.3.1   Introduction to Hand Shapes in Sign Language

The most prominent feature of signed languages is their use of the hands. In fact, the hands play an integral role in the phonetics of signs, and a slight variation in hand shape can convey differences in meaning (Stokoe Jr, 1960). In sign languages such as American Sign Language (ASL) and British Sign Language (BSL), different hand shapes contribute to the vocabulary of the language, similar to how different sounds contribute to the vocabulary of spoken languages. ASL is estimated to use between 30 to 80 hand shapes[1], while BSL is limited to approximately 40 hand shapes[2]. SignWriting (Sutton, 1990), a system of notation used for sign languages, specifies a superset of 261 distinct hand shapes (Frost and Sutton, 2022). Each sign language uses a subset of these hand shapes.

Despite the fundamental role of hand shapes in sign languages, accurately recognizing and classifying them is a challenging task. In this section, we explore rule-based hand shape analysis in sign languages using 3D hand normalization. By performing 3D hand normalization, we can transform any given hand shape to a fixed orientation, making it easier for a model to extract the hand shape, and hence improving the recognition and classification of hand shapes in sign languages.

### 3.3.2   Characteristics of the Human Hand

The human hand consists of 27 bones and can be divided into three main sections: the wrist (carpals), the palm (metacarpals), and the fingers (phalanges). Each finger consists of three bones, except for the thumb, which has two. The bones are connected by joints, which allow for the complex movements and shapes that the hand can form.

Understanding the different characteristics of hands and their implications in signed languages is crucial for the extraction and classification of hand shapes. These characteristics are based on the SignWriting definitions of the five major axes of hand variation: handedness, plane, rotation, view, and shape.

---

[1] https://aslfont.github.io/Symbol-Font-For-ASL/asl/handshapes.html
[2] https://bsl.surrey.ac.uk/principles/i-hand-shapes

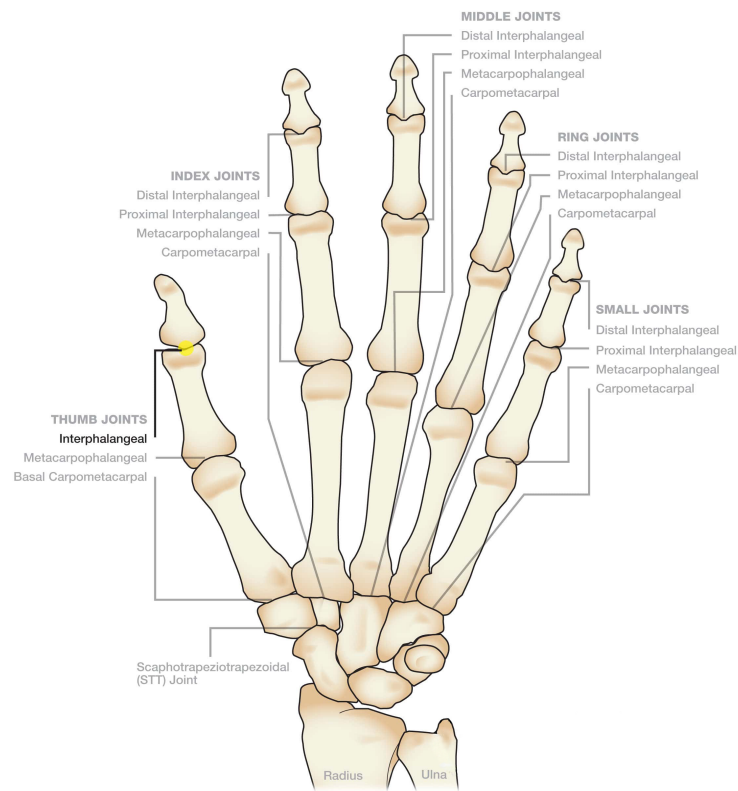

Figure 3.2: Anatomy of a human hand.
©American Society for Surgery of the Hand

**Handedness**   is the distinction between the right and left hands. Signed languages make a distinction between the dominant hand and the non-dominant hand. For right-handed individuals, the right hand is considered dominant, and vice-versa. The dominant hand is used for fingerspelling and all one-handed signs, while the non-dominant hand is used for support and two-handed signs. Using 3D pose estimation, the handedness analysis is trivial, as the pose estimation platform predicts which hand is which.

**Plane** refers to whether the hand is parallel to the wall or the floor. The variation in the plane can, but does not have to, create a distinction between two signs. For example, in ASL the signs for "date" and "dessert" exhibit the same hand shape, view, rotation, and movement, but differ by plane. The plane of a hand can be estimated by comparing the positions of the wrist and middle finger metacarpal bone ($M\_MCP$).

---

**Algorithm 1** Hand Plane Estimation

---

$y \leftarrow |M\_MCP.y - WRIST.y| \times 1.5$ // add bias to y
$z \leftarrow |M\_MCP.z - WRIST.z|$
**return** $y > z$ ? 'wall' : 'floor'

---

**Rotation** refers to the angle of the hand in relation to the body. SignWriting groups the hand rotation into eight equal categories, each spanning 45 degrees. The rotation of a hand can be calculated by finding the angle of the line created by the wrist and the middle finger metacarpal bone.

**View** refers to the side of the hand as observed by the signer, and is grouped into four categories: front, back, sideways, and other-sideways. The view of a hand can be estimated by analyzing the normal of the plane created by the palm of the hand (between the wrist, index finger metacarpal bone, and pinky metacarpal bone).

---

**Algorithm 2** Hand View Estimation

---

$normal \leftarrow \text{math.normal}(WRIST, I\_MCP, P\_MCP)$
$plane \leftarrow \text{get\_plane}(WRIST, M\_MCP)$
**if** plane = 'wall' **then**
    $angle \leftarrow \angle(normal.z, normal.x)$
    **return** $angle > 210$ ? 'front' : ($angle > 150$ ? 'sideways' : 'back')
**else**
    $angle \leftarrow \angle(normal.y, normal.x)$
    **return** $angle > 0$ ? 'front' : ($angle > -60$ ? 'sideways' : 'back')
**end if**

---

**Shape** refers to the configuration of the fingers and thumb. This characteristic of the hand is the most complex to analyze due to the vast array of possible shapes the human hand can form. The shape of a hand is determined by the state of each finger and thumb, specifically whether they are straight, curved, or bent, and their position relative to each other. Shape analysis can be accomplished by examining the bend and rotation of each finger joint. More advanced models may also take into consideration the spread between the fingers and other nuanced characteristics. 3D pose estimation can be used to extract these features for a machine learning model, which can then classify the hand shape.

### 3.3.3   3D Hand Normalization

3D hand normalization is an attempt at standardizing the orientation and position of the hand, thereby enabling models to effectively classify various hand shapes. The normalization process involves several steps, as illustrated below:

1. **Pose Estimation** Initially, the 3D pose of the hand is estimated from the hand image crop (Figure 3.3).

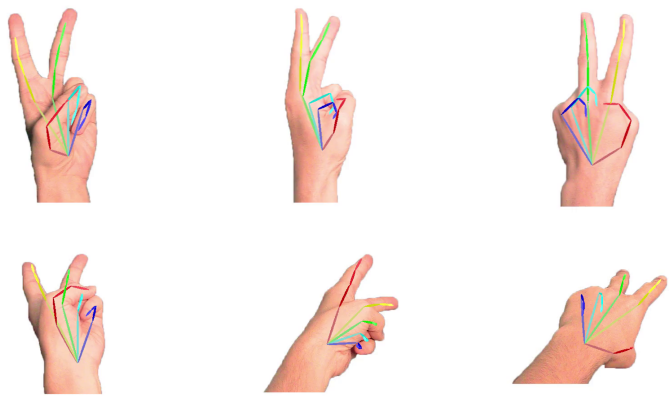

Figure 3.3: Pictures of six hands all performing the same hand shape (v-shape) taken from six different orientations. MediaPipe fails at estimating the pose of the bottom-middle image.

2. **3D Rotation** The pose is then rotated in 3D space such that the normal of the back of the hand aligns with the $Z$-axis. As a result, the palm plane now resides within the $XY$ plane (Figure 3.4).

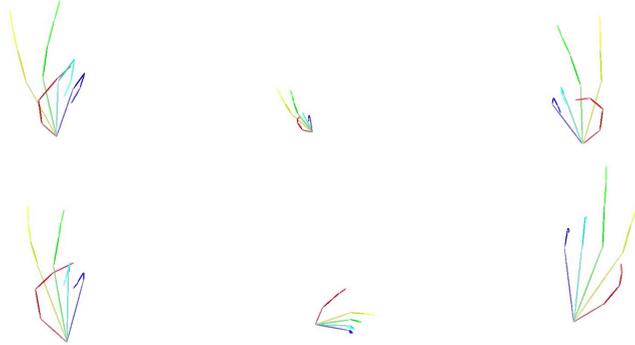

Figure 3.4: Hand poses after 3D rotation. The scale difference between the hands demonstrates a limitation of the 3D pose estimation system used.

3. **2D Orientation** Subsequently, the pose is rotated in 2D such that the metacarpal bone of the middle finger aligns with the $Y$-axis (Figure 3.5).

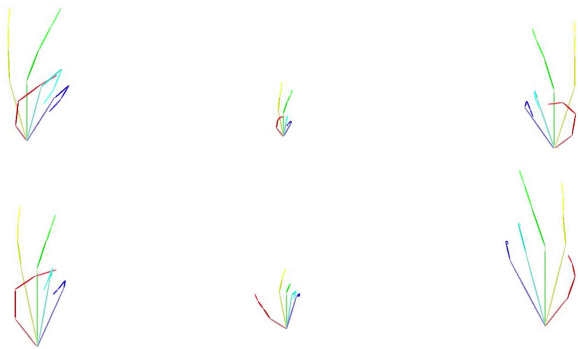

Figure 3.5: Hand poses after being rotated.

4. **Scale** The hand is scaled such that the metacarpal bone of the middle finger attains a constant length (which we typically set to 200, Figure 3.6).

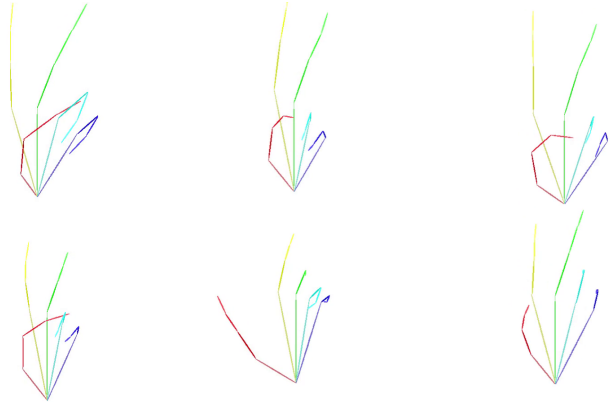

Figure 3.6: Hand poses after being scaled.

5. **Translation** Lastly, the wrist joint is translated to the origin of the coordinate system $(0,0,0)$. Figure 3.7 demonstrates how when overlayed, we can see all hands producing the same shape, except for one outlier.

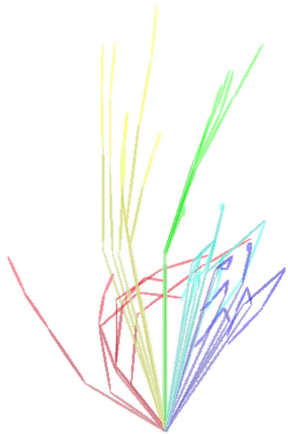

Figure 3.7: Normalized hand poses overlayed after being translated to the same position. The positions of the wrist and the metacarpal bone of the middle finger are fixed.

By conducting these normalization steps, a hand model can be standardized, reducing the complexity of subsequent steps such as feature extraction and hand shape classification. This standardization simplifies the recognition process and can contribute to improving the overall accuracy of the system.

### 3.3.4 3D Hand Pose Evaluation

In order to assess the performance of our 3D hand pose estimation and normalization, we introduce two metrics that gauge the consistency of the pose estimation across orientations and crops.

Our dataset is extracted from the SignWriting Hand Symbols Manual Frost and Sutton (2022), and includes images of 261 different hand shapes, from 6 different angles. All images are of the same hand, of an adult white man.

**Multi Angle Consistency Error (MACE)**   evaluates the consistency of the pose estimation system across the different orientations. We perform 3D hand normalization, and overlay the hands. The MACE score is the average standard deviation of all pose landmarks, between all views. A high MACE score indicates a problem in the pose estimation system's ability to maintain consistency across different orientations. This could adversely affect the model's performance when analyzing hand shapes in sign languages, as signs can significantly vary with hand rotation.

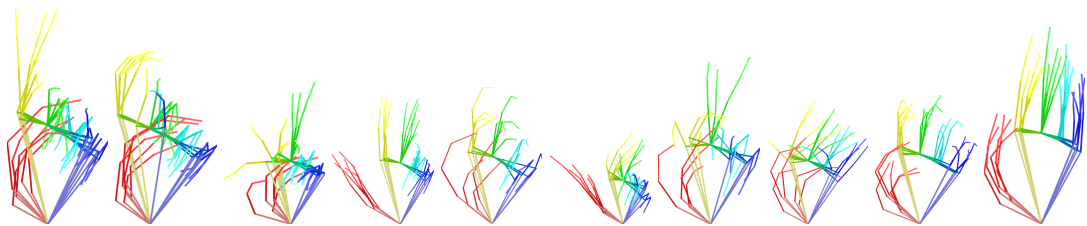

Figure 3.8: Visualizations of 10 hand shapes, each with 6 orientations 3D normalized and overlayed.

Figure 3.8 shows that our 3D normalization does work to some extent using MediaPipe. We can identify differences across hand shapes, but still note high variance within each hand shape.

**Crop Consistency Error (CCE)** gauges the pose estimation system's consistency across different crop sizes. We do not perform 3D normalization, but still overlay all the estimated hands, shifting the wrist point of each estimated hand to the origin $(0, 0, 0)$. The CCE score is the average standard deviation of all pose landmarks across crops. A high CCE score indicates that the pose estimation system is sensitive to the size of the input crop, which is a significant drawback as the system should be invariant to the size of the input image.

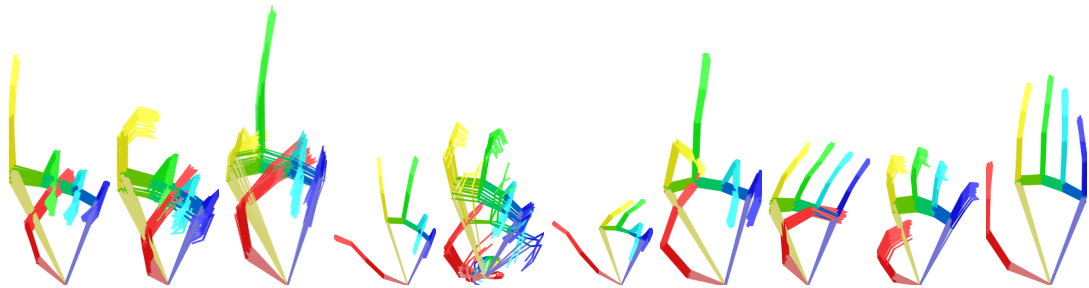

Figure 3.9: Visualizations of 10 hand shapes, each with 48 crops overlayed.

Figure 3.9 shows that for some poses, MediaPipe is very resilient to crop size differences (e.g. the first and last hand shapes). However, it is concerning that for some hand shapes, it exhibits very high variance, and possibly even wrong predictions.

### 3.3.5 Conclusion

Our normalization process appears to work reasonably well when applied to different views within the same crop size. It succeeds in simplifying the hand shape, which in turn, can aid in improving the accuracy of hand shape classification systems.

However, it is crucial to note that while this method may seem to perform well on a static image, its consistency and reliability in a dynamic context, such as a video, may be quite different. In a video, the crop size can change between frames, introducing additional complexity and variance. This dynamic nature coupled with the inherently noisy nature of the estimation process can pose challenges for a model that aims to consistently estimate hand shapes.

In light of these findings, it is clear that there is a need for the developers of 3D pose estimation systems to consider these evaluation methods and strive to make their systems more robust to changes in hand crops. The Multi Angle Consistency Error (MACE) and the Crop Consistency Error (CCE) can be valuable tools in this regard.

MACE could potentially be incorporated as a loss function for 3D pose estimation, thereby driving the model to maintain consistency across different orientations. Alternatively, MACE could be used as an indicator to identify hand shapes that require more training data. It is apparent from our study that the performance varies greatly across hand shapes and orientations, and this approach could help in prioritizing the allocation of training resources.

Ultimately, the goal of improving 3D hand pose estimation is to enhance the ability to encode signed languages accurately. The insights gathered from this study can guide future research and development efforts in this direction, paving the way for more robust and reliable sign language technology.

The benchmark, metrics, and visualizations are available at [https://github.com/sign-language-processing/3d-hands-benchmark](https://github.com/sign-language-processing/3d-hands-benchmark).

# Part II

# Sign Language Transcription

"we learn from history that we do not learn from history."

— Georg Wilhelm Friedrich Hegel

# Chapter 4

# Introduction

This chapter introduces the methodology adopted in this thesis to advance the field of Sign Language Processing, recognizing the unique challenges posed by the different modalities of signed and spoken languages. We embarked on a journey to adapt and extend the fundamental NLP theories to signed languages. To achieve this, a critical part of the puzzle was the introduction of an intermediary transcription system, able to map continuous signed language videos to a discrete, accurate representation without linguistic information loss.

At the core of this research was the acknowledgment of the shortcomings of existing SLP systems and datasets, which often rely on glosses for discretization. We identified three key challenges with this approach: the inability of linear, single-dimensional glosses to capture the multidimensional spatial nature of signed languages; the language-specificity of glosses requiring impractical glossing models for each language; and the lack of standardization across corpora, which inhibits data sharing and complicates modeling.

Our proposed solution was to adopt a universal and standardized representation for the transcription and tokenization of signed languages, considering key linguistic factors such as the degree to which phonological units of signed languages can be mapped to lexical units, the cross-linguistic phonological differences, and the extent to which automatic transcription methods could be applied to signed languages.

For our purposes, we relied on SignWriting, an existing universal notation system for signed languages. Although its 2D construction poses challenges for token representation, the advantages of human readability and comprehension of the transcription were deemed critical. SignWriting provided us with a framework to pivot from language-specific models to universal ones.

With the foundation of our transcription-based approach in place, we then explored the dual aspects of sign language translation, namely the translation from a sign language video to spoken language text (translation), and from spoken language text to a sign language video (production). The use of lexical transcription as an intermediary step was examined in both directions.

In the following chapters, we will delve into the detailed procedures and techniques utilized in our research, including preliminary work in the field (Chapter 5), and the application and evaluation of SignWriting in sign language translation (Chapter 6) and production (Chapter 7).

# Chapter 5

# Preliminary Work

## 5.1 Activity Detection (Moryossef et al., 2020)

We propose a lightweight real-time sign language detection model, as we identify the need for such a case in videoconferencing. We extract optical flow features based on human pose estimation and, using a linear classifier, show these features are meaningful with an accuracy of 80%, evaluated on the Public DGS Corpus. Using a recurrent model directly on the input, we see improvements of up to 91% accuracy, while still working under 4ms. We describe a demo application to sign language detection in the browser in order to demonstrate its usage possibility in videoconferencing applications.

### 5.1.1 Introduction

Sign language detection (Borg and Camilleri, 2019) is defined as the binary-classification task for any given frame of a video if a person is using sign-language or not. Unlike sign language recognition (Camgöz et al., 2018; Cui et al., 2017), where the task is to recognize the form and meaning of signs in a video, or sign language identification, where the task is to identify *which* sign language is used, the task of sign language detection is to detect *when* something is being signed.

With the recent rise of videoconferencing platforms, we identify the problem of signers not "getting the floor" when communicating, which either leads to them being ignored or to a cognitive load on other participants, always checking to see if someone starts signing. Hence, we focus on the real-time sign language detection task with uni-directional context to allow for videoconferencing sign language prominence.

We propose a simple human optical-flow representation for videos based on pose estimation (§5.1.3), which is fed to a temporally sensitive neural network (§5.1.3) to perform a binary classification per frame — is the person signing or not. We compare various possible inputs, such as full-body pose estimation, partial pose estimation, and bounding boxes (§5.1.4), and contrast their acquisition time in light of our targeted real-time application.

We demonstrate our approach on the Public DGS Corpus (German Sign Language) (Hanke et al., 2020), using full-body pose estimation (Schulder and Hanke, 2019) collected through OpenPose (Cao et al., 2019; Simon et al., 2017). We show results of 87%-91% prediction accuracy depending on the input, with per-frame inference time of $350 - 3500\mu s$ (§5.1.5), and release our training code and models[1].

### 5.1.2 Background

**Sign Language Detection**

Sign language detection (Borg and Camilleri, 2019; Pal et al., 2023) is the binary classification task of determining whether signing activity is present in a given video frame. A similar task in spoken languages is voice activity detection (VAD) (Sohn et al., 1999; Ramırez et al., 2004), the detection of when a human voice is used in an audio signal. As VAD methods often rely on speech-specific representations such as spectrograms, they are not necessarily applicable to videos.

---

[1] https://github.com/google-research/google-research/tree/master/sign_language_detection

Borg and Camilleri (2019) introduced the classification of frames taken from YouTube videos as either signing or not signing. They took a spatial and temporal approach based on VGG-16 (Simonyan and Zisserman, 2015) CNN to encode each frame and used a Gated Recurrent Unit (GRU) (Cho et al., 2014) to encode the sequence of frames in a window of 20 frames at 5fps. In addition to the raw frame, they either encoded optical-flow history, aggregated motion history, or frame difference. We improve upon their method by performing sign language detection in real time. They identified that sign language use involves movement of the body and, as such, designed a model that works on top of estimated human poses rather than directly on the video signal. They calculated the optical flow norm of every joint detected on the body and applied a shallow yet effective contextualized model to predict for every frame whether the person is signing or not.

While these recent detection models achieve high performance, we need well-annotated data that include interference and distractions with non-signing instances for proper real-world evaluation. Pal et al. (2023) conducted a detailed analysis of the impact of signer overlap between the training and test sets on two sign detection benchmark datasets (Signing in the Wild (Borg and Camilleri, 2019) and the DGS Corpus (Hanke et al., 2020)) used by Borg and Camilleri (2019) and by us. By comparing the accuracy with and without overlap, they noticed a relative decrease in performance for signers not present during training. As a result, they suggested new dataset partitions that eliminate overlap between train and test sets and facilitate a more accurate evaluation.

**Sign Language Identification**

Sign language identification (Gebre et al., 2013; Monteiro et al., 2016) classifies which signed language is used in a given video.

Gebre et al. (2013) found that a simple random-forest classifier utilizing the distribution of phonemes can distinguish between British Sign Language (BSL) and Greek Sign Language (ENN) with a 95% F1 score. This finding is further supported by Monteiro et al. (2016), which, based on activity maps in signing space, manages to differentiate between British Sign Language and French Sign Language (Langue des Signes Française, LSF) with a 98% F1 score in videos with static backgrounds, and between American Sign Language and British Sign Language, with a 70% F1 score for videos mined from popular video-sharing sites. The authors attribute their success mainly to the different fingerspelling systems, which are two-handed in the case of BSL and one-handed in the case of ASL and LSF.

Although these pairwise classification results seem promising, better models would be needed for classifying from a large set of signed languages. These methods only rely on low-level visual features, while signed languages have several distinctive features on a linguistic level, such as lexical or structural differences (McKee and Kennedy, 2000; Kimmelman, 2014; Ferreira-Brito, 1984; Shroyer and Shroyer, 1984) which have not been explored for this task.

### 5.1.3   Model

For a video, for every frame given, we would like to predict whether the person in the video is signing or not.

**Input Representation**

As evident by previous work (Borg and Camilleri, 2019), using the raw frames as input is computationally expensive, and noisy. Alternatively, in computer vision, optical flow is one way to calculate the movement of every object in a scene. However, because signing is inherently a human action, we do not care about the flow of every object, but rather only the flow of the human. Optimally, we would like to track the movement of every pixel on the human body from one frame to another, to gauge its movement vector. As a proxy to such data, we opt for full-body human pose estimation, defining a set of points detected in every video frame that marks informative landmarks, like joints and other moving parts (mouth, eyes, eyebrows, and others).

Getting the optical flow $F$ for these predefined points $P$ at time $t$ is then well defined as the L2 norm of the vector resulting from subtracting every two consecutive frames. We normalize the flow by the frame-rate in which the video was captured for the representation to be frame-rate invariant (Equation 5.1).

$$F(P)_t = ||P_t - P_{t-1}||_2 * fps \qquad (5.1)$$

We note that if a point $p$ was not identified in a given frame $t$, the value of $F(p)_t$ and $F(p)_{t+1}$ automatically equals to 0. This is done to avoid introducing fake movements from a poor pose estimation system or unknown movement from landmarks going out-of-frame.

An additional benefit of using full-body pose estimation is that we can normalize the size of all people, regardless of whether they use a high-/low-resolution camera and the distance at which they are from the camera.

**Temporal Model**

Figure 5.1 demonstrates our input representation for an example video. It shows, to the naked eye, that this representation is meaningful. The movement, indicated by the bright colors, is well aligned with the gold spans annotation. Thus, we opt to use a shallow sequence tagging model on top of it.

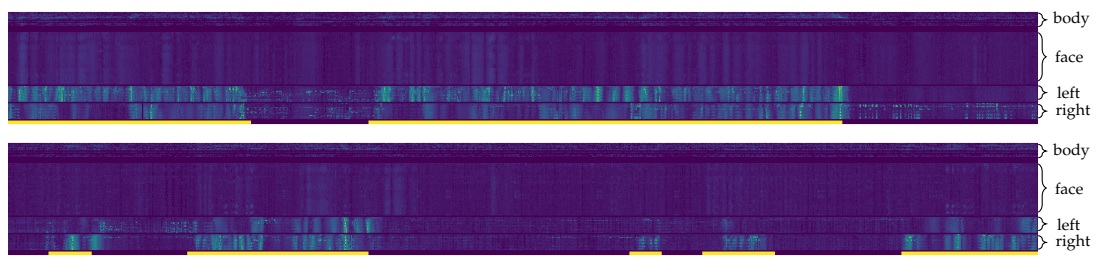

Figure 5.1: Optical-flow norm representation of a conversation between two signers. The x-axis is the progression of time, 1,500 frames over 30 seconds in total. The yellow marks are the gold labels for spans when a signer is signing.

We use a uni-directional LSTM (Hochreiter and Schmidhuber, 1997) with one layer and 64 hidden units directly on this input, normalized for frame rate, and project the output to a 2-dimensional vector. For training, we use the negative-log-likelihood loss on the predicted classes for every frame. For inference, we take the arg-max of the output vector (Equation 5.2).

$$signing(P) = \arg\max LSTM(F(P)) * W \qquad (5.2)$$

Note that this model allows us to process each frame as we get it, in real-time, by performing a single step of the LSTM and project its output. Unlike autoregressive models, we do not feed the last-frame classification as input for the next frame, as just classifying the new frame with the same tag would almost get 100% accuracy on this task, depending on gold labels to be available. Instead, we rely on the hidden state of the LSTM to hold such information.

### 5.1.4 Experiments

The Public DGS Corpus (Hanke et al., 2020) includes 301 videos with an average duration of 9 minutes, of two signers in conversation[2], at 50fps. Each video includes gloss annotations and spoken language translations (German and English). Using this information, we mark each frame as either "signing" (50.9% of the data) or "not-signing" (49.1% of the data) depending on whether it belongs to a gloss segment. Furthermore, this corpus is enriched with OpenPose (Cao et al., 2019) full-body pose estimations (Schulder and Hanke, 2019) including 137 points per frame (70 for the face, 25 for the body, and 21 for each hand). In order to disregard video resolution and distance from the camera, we normalize each of these poses such that the mean distance between the shoulders of each person equals 1. We split this dataset into 50:25:25 for training, validation, and test, respectively. For every "part" (face, body, left and right hands), we also calculate its bounding box based on the minimum and maximum value of all of the landmarks.

We experiment with three linear baselines with a fixed context (Linear-1, Linear-25, Linear-50) and four experimental recurrent models with different counts of input features:

---

[2]There are also monologue story-telling, but both signers are always shown.

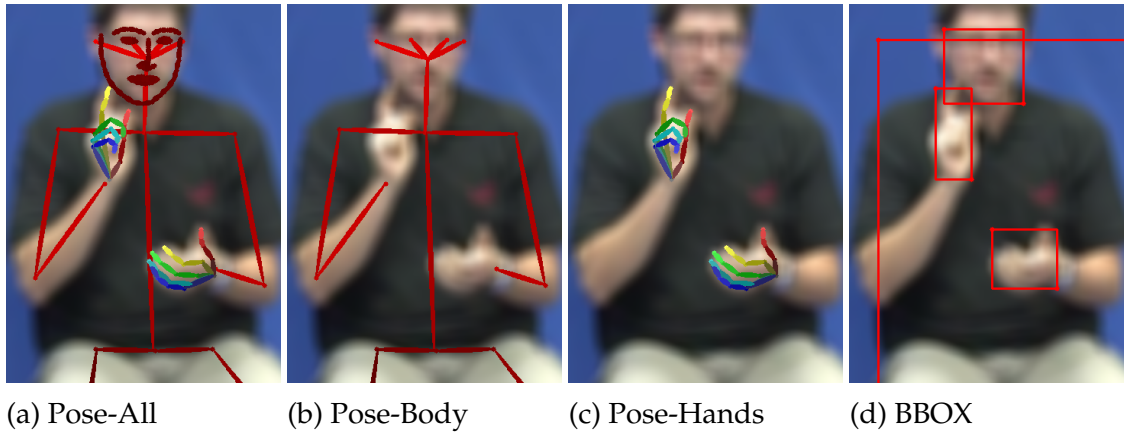

(a) Pose-All     (b) Pose-Body     (c) Pose-Hands     (d) BBOX

Figure 5.2: Visualization of our different experiments inputs.

1. **Pose-All**—all of the landmarks from the poses. (f. 5.2a)

2. **Pose-Body**—only the body landmarks. (f. 5.2b)

3. **Pose-Hands**—only the left- and right-hand landmarks. (f. 5.2c)

4. **BBOX**—the bounding boxes of the face, body, and hands. (f. 5.2d)

Finally, we measure the execution time of each model on CPU, using an Intel(R) Xeon(R) CPU E5-2650 v4 @ 2.20GHz. We measure the execution time per frame given a single frame at a time, using multiple frameworks: Scikit-Learn (sk) (Pedregosa et al., 2011), TensorFlow (tf) (Abadi et al., 2015) and PyTorch (pt) (Paszke et al., 2019b).

## 5.1.5 Results

Table 5.1 includes the accuracy and inference times for each of our scenarios. Our baseline systems show that using a linear classifier with a fixed number of context frames achieves between 79.9% to 84.3% accuracy on the test set. However, all of the baselines perform worse than our recurrent models, for which we achieve between 87.7% to 91.5% accuracy on the test set. Generally, we see that using more diverse sets of landmarks performs better. Although the hand landmarks are very indicative, using just the hand BBOX almost matches in accuracy, and using the entire body pose, with a single point per hand, performs much better. Furthermore, we see that regardless of the number of landmarks used, our models generally perform faster the fewer landmarks are used. We note that the prediction time varies between the different frameworks, but does not vary much within a particular framework. It is clear, however, that the speed of these models' is sufficient, as even the slowest model, using the slowest framework, runs at 285 frames-per-second on CPU.

We note from manually observing the gold data that sometimes a gloss segment starts before the person actually begins signing, or moving at all. This means that our accuracy ceiling is not 100%. We did not perform a rigorous re-annotation of the dataset to quantify how extensive this problem is.

| Model | Points | Params | Dev Acc | Test Acc | $\partial t$ (sk) | $\partial t$ (tf) | $\partial t$ (pt) |
|-------|--------|--------|---------|----------|---------|---------|---------|
| Linear-1 | 25 | 25 | 79.99% | 79.93% | $6.49\mu s$ | $823\mu s$ | $2.75\mu s$ |
| Linear-25 | 25 | 625 | 84.13% | 83.79% | $6.78\mu s$ | $824\mu s$ | $5.10\mu s$ |
| Linear-50 | 25 | $1,250$ | 85.06% | 83.39% | $6.90\mu s$ | $821\mu s$ | $7.41\mu s$ |
| BBOX | 8 | $18,818$ | 87.49% | 87.78% | — | $3519\mu s$ | $367\mu s$ |
| Pose-Hands | 42 | $27,522$ | 87.65% | 88.05% | — | $3427\mu s$ | $486\mu s$ |
| Pose-Body | 25 | $23,170$ | 92.17% | 90.35% | — | $3437\mu s$ | $443\mu s$ |
| Pose-All | 137 | $51,842$ | 92.31% | 91.53% | — | $3537\mu s$ | $588\mu s$ |

Table 5.1: Accuracy and inference-time ($\partial t$) results for the various experiments.

### 5.1.6 Analysis

As we know that different pose landmarks have varying importance to the classification, we use the *Linear-1* model's coefficients magnitude to visualize how the different landmarks contribute. Figure 5.3 visualizes the average human pose in the dataset, with the opacity of every landmark being the absolute value of the coefficient.

First, we note that the model attributes no importance to any landmark below the waist. This makes sense as they both do not appear in all videos, and bare no meaning in sign language. The eyes and nose seem to carry little weight, while the ears carry more. We do not attribute this to any signing phenomenon.

Additionally, we note hands asymmetry. While both wrists have a high weight, the elbow and shoulder for the right hand carry more weights than their corresponding left counterparts. This could be attributed to the fact that most people are right handed, and that in some sign languages the signer must decide which hand is dominant in a consistent manner. We see this asymmetry as a feature of our model, and note that apps using our models could also include a "dominant hand" selection.

To further understand what situations our models capture, we check multiple properties of them on the test set. We start by generally noting that our data is conversational. 84.87% of the time, only one participant is signing, while 8.5% of the time both participants are signing, and in the remaining 6.63% of the time no one is signing, primarily when the participants are being instructed on the task.

Our test set includes 4,138 *signing* sequences with an average length of 11.35 seconds, and a standard deviation of 29.82 seconds. It also includes 4,091 *not-signing* sequences with an average length of 9.95 seconds, and a standard deviation of 24.18 seconds.

For each of our models, we compare the following error types (Figure 5.4):

- **Bridged**—Cases where the model bridged between two signing sections,

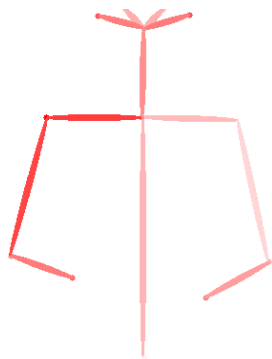

Figure 5.3: The average pose in the dataset. The opacity of every landmark is determined by its coeffient in the *Linear-1* model.

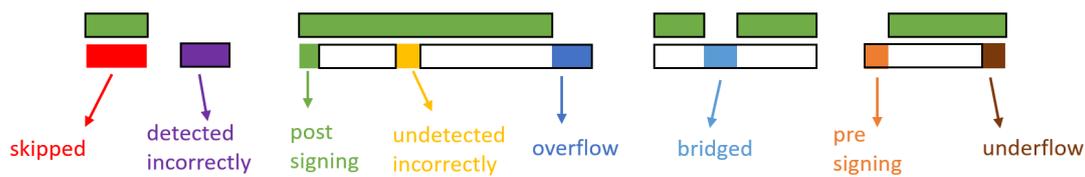

Figure 5.4: Visualization of the different types of errors. The first row contains the gold annotations, and the second row contains a model's prediction.

still predicting the person to be *signing* while the annotation says they are not.

- **Signing Detected Incorrectly**—Cases where the model predicted a *signing* span fully contained within a *not-signing* annotation.

- **Signing Overflow**—Cases where signing was still predicted after a *signing* section ended.

- **Started Pre-Signing**—Cases where *signing* was predicted before a *signing* section started.

- **Skipped**—Cases where the model did not detect entire *signing* sections.

- **Signing Undetected Incorrectly**—Cases where the model predicted a *not-signing* span fully contained within a *signing* annotation.

- **Started Post-Signing**—Cases where the *signing* section started before it was predicted to start.

- **Signing Underflow**—Cases where the *signing* section was predicted to end prematurely.

Table 5.2 includes the number of sequences, including average length and standard deviation in seconds, for each of the error types. Most notably, we see that the less context the model has, the more sporadic its predictions and thus it will generally completely bridge or skip less sequences. The same locality however introduces many signing detected / undetected incorrectly errors, albeit of short lengths.

|  | linear-1 | linear-25 | linear-50 | |
|---|---|---|---|---|
| Bridged | 107 (0.10 ± 0.15) | 308 (0.34 ± 0.40) | 426 (0.45 ± 0.46) | |
| Signing Detected Incorrectly | 132151 (0.04 ± 0.07) | 8773 (0.30 ± 0.81) | 6594 (0.34 ± 1.06) | |
| Signing Overflow | 4094 (0.09 ± 0.15) | 3893 (0.32 ± 0.43) | 3775 (0.46 ± 1.17) | |
| Started Pre-Signing | 873 (0.09 ± 0.13) | 345 (0.45 ± 0.68) | 257 (0.88 ± 4.27) | |
| Skipped | 50 (1.41 ± 1.95) | 298 (1.38 ± 1.43) | 446 (1.49 ± 1.60) | |
| Signing undetected incorrectly | 219531 (0.05 ± 0.10) | 26185 (0.27 ± 0.50) | 18037 (0.32 ± 0.66) | |
| Started Post-Signing | 4199 (0.17 ± 0.23) | 3951 (0.48 ± 0.57) | 3803 (0.60 ± 0.77) | |
| Signing Underflow | 1677 (0.15 ± 0.26) | 1092 (0.58 ± 0.91) | 827 (0.71 ± 0.96) | |
|  | **BBOX** | **Pose-Hands** | **Pose-Body** | **Pose-All** |
| Bridged | 754 (0.97 ± 1.94) | 861 (1.26 ± 2.63) | 747 (1.12 ± 2.35) | 573 (0.75 ± 1.08) |
| Signing Detected Incorrectly | 5697 (0.64 ± 1.93) | 12919 (0.33 ± 1.33) | 6286 (0.38 ± 1.29) | 11384 (0.25 ± 1.14) |
| Signing Overflow | 3337 (0.95 ± 2.10) | 3230 (1.01 ± 2.46) | 3344 (0.67 ± 1.29) | 3518 (0.48 ± 0.87) |
| Started Pre-Signing | 402 (1.33 ± 2.73) | 558 (1.59 ± 5.15) | 298 (1.48 ± 3.87) | 408 (0.70 ± 1.97) |
| Skipped | 199 (1.31 ± 1.40) | 115 (1.45 ± 1.54) | 243 (1.31 ± 1.30) | 146 (1.41 ± 1.42) |
| Signing undetected incorrectly | 4089 (0.48 ± 0.76) | 3526 (0.26 ± 0.51) | 4786 (0.32 ± 0.60) | 5526 (0.23 ± 0.44) |
| Started Post-Signing | 3939 (0.34 ± 0.44) | 4023 (0.24 ± 0.34) | 3895 (0.37 ± 0.49) | 3992 (0.29 ± 0.36) |
| Signing Underflow | 370 (0.82 ± 1.08) | 297 (0.55 ± 0.68) | 506 (0.63 ± 0.97) | 666 (0.44 ± 0.66) |

Table 5.2: We evaluate every model on the different error types, and show number of sequences with that error, including average sequence length in seconds and standard deviation.

In the sequential models, we generally see a lower number of sequences as they can incorporate global features in the classification. As indicated by the accuracy scores, we see fewer errors of most types the more diverse the input points are, with one notable exception for the *Pose-All* model which underperforms *Pose-Body* on all errors except for *Bridged* and *Skipped*.

### 5.1.7 Demo Application

With this publication, we release a demo application working in the browser for computers and mobile devices. Pragmatically, we choose to use the "Pose-Body" model variant, as it performs almost on par with our best model, "Pose-All", and we find it is feasible to acquire the human body poses in real-time with currently available tools.

We use PoseNet (Papandreou et al., 2017, 2018) running in the browser using TensorFlow.js (Smilkov et al., 2019). PoseNet includes two main image encoding variants: MobileNet (Howard et al., 2017), which is a lightweight model aimed at mobile devices, and ResNet (He et al., 2016), which is a larger model that requires a dedicated GPU. Each model includes many sub-variants with different image resolution and convolutional strides, to further allow for tailoring the network to the user's needs. In our demo, we first tailor a network to the current device to run at least at 25fps. While using a more lightweight network might be faster, it might also introduce pose estimation errors.

The pose estimation we use only returns 17 points compared to the 25 of OpenPose; hence, we map the 17 points to the corresponding indexes for Open-Pose. We then normalize the body pose vector by the mean shoulder width the person had in the past 50 frames in order to disregard camera resolution and distance of the signer from the camera.

Onward, there are two options: either send the pose vector to the video-conferencing server where inference could be done or perform the inference locally. As our method is faster than real-time, we chose the latter and perform inference on the device using TensorFlow.js. For every frame, we get a signing probability, which we then show on the screen.

In a production videoconferencing application, this signing probability should be streamed to the call server, where further processing could be done to show the correct people on screen. We suggest using the signing probability as a normalized "volume", such that further processing is comparable to videoconferencing users using speech.

While this is the recommended way to add sign language detection to a videoconferencing app, as the goal of this work is to empower signers, our demo application can trigger the speaker detection by transmitting audio when the user is signing. Transmitting ultrasonic audio at 20KHz, which is inaudible for humans, manages to fool Google Meet, Zoom and Slack into thinking the user is speaking, while still being inaudible. One limitation of this method is that videoconferencing app developers can crop the audio to be in the audible human range and thus render this application useless. Another limitation is that using high-frequency audio can sound crackly when compressed, depending on the signer's internet connection strength.

Our model and demo, in their current forms, only allow for the detection of a single signer per video stream. However, if we can detect more than a single person, and track which poses belong to which person in every frame, there is no limitation to run our model independently on each signer.

### 5.1.8 Discussion

**Limitations**

We note several limitations to our approach. The first is that it relies on the pose estimation system to run in real-time on any user's device. This proves to be challenging, as even performing state-of-the-art pose estimation on a single frame on a GPU with OpenPose (Cao et al., 2019; Cao et al., 2017) can take upwards of 300ms, which introduces two issues: (1) If in order to get the optical-flow, we need to pose two frames, we create a delay from when a person starts signing to when they could be accurately detected as signing, equal to at least two times the pose processing time. (2) Running this on mobile devices or devices without hardware acceleration like a GPU may be too slow.

As we only look at the input's optical flow norm, our model might not be able to pick up on times when a person is just gesturing rather than signing. However, as this approach is targeted directly at sign language users rather than the general non-signing public, erring on the side of caution and detecting any meaningful movements is preferred.

**Demographic Biases**

The data we use for training was collected from various regions of Germany, with equal number of males and females, as well as an equal number of participants from different age groups (Schulder et al., 2020). Although most of the people in the dataset are European white, we do not attribute any significance between the color of their skin to the performance of the system, as long as the pose estimation system is not biased.

Regardless of age, gender, and race, we do not address general ethnic biases such as different communities of signers outside of Germany signing differently - whether it is the size, volume, speed, or other properties.

**Signer Independence**

Following the publication of this work, Pal et al. (2023) conducted a detailed analysis of the impact of signer overlap between our training and test sets. By comparing the accuracy with and without overlap, they noticed a relative decrease in performance for signers not present during training. As a result, they suggested new dataset partitions that eliminate overlap between train and test sets and facilitate a more accurate evaluation of performance.

### 5.1.9   Conclusions

We propose a simple human optical-flow representation for videos based on pose estimation to perform a binary classification per frame — is the person signing or not. We compare various possible inputs, such as full-body pose estimation, partial pose estimation, and bounding boxes and contrast their acquisition time in light of our targeted real-time videoconferencing sign language detection application.

We demonstrate our approach on the Public DGS Corpus (German Sign Language), and show results of 87%-91% prediction accuracy depending on the input, with per-frame inference time of $350 - 3500\mu$s.

## 5.2 Isolated Recognition (Moryossef et al., 2021b)

In this section, we explore whether or not estimated skeletal poses are viable for use in sign language recognition. A large part of this section was independently published as "Evaluating the Immediate Applicability of Pose Estimation for Sign Language Recognition".

Sign languages are visual languages produced by the movement of the hands, face, and body. In this paper, we evaluate representations based on skeleton poses, as these are explainable, person-independent, privacy-preserving, low-dimensional representations. Basically, skeletal representations generalize over an individual's appearance and background, allowing us to focus on the recognition of motion. But how much information is lost by the skeletal representation? We perform two independent studies using two state-of-the-art pose estimation systems. We analyze the applicability of the pose estimation systems to sign language recognition by evaluating the failure cases of the recognition models. Importantly, this allows us to characterize the current limitations of skeletal pose estimation approaches in sign language recognition.

### 5.2.1 Introduction

Sign languages are visual languages produced by the movement of the hands, face, and body. As languages that rely on visual communication, recordings are in video form. Current state-of-the-art sign language processing systems rely on the video to model tasks such as sign language recognition (SLR) and sign language translation (SLT). However, using the raw video signal is computationally expensive and can lead to overfitting and person dependence.

In an attempt to abstract over the video information, skeleton poses have been suggested as an explainable, person-independent, privacy-preserving, and low-dimensional representation that provides the signer body pose and information on how it changes over time. Theoretically, skeletal poses contain all the relevant information required to understand signs produced in videos, except for interactions with elements in space (for example, a mug or a table).

The recording of accurate human skeleton poses is difficult and often intrusive, requiring signers to wear specialized and expensive motion capture hardware. Fortunately, advances in computer vision now allow the estimation of human skeleton poses directly from videos. However, as these estimation systems were not specifically designed with sign language in mind, we currently do not understand their suitability for use in processing sign languages both in recognition or production.

In this study, we evaluate two pose estimation systems and demonstrate their suitability (and limitations) for SLR by conducting two independent studies on the CVPR21 ChaLearn challenge Sincan et al. (2021). Because we perform no pretraining of the skeletal model, the final results are considerably lower than potential end-to-end approaches (§5.2.3). The results demonstrate that the skeletal representation loses considerable information. To better understand why, we evaluate our approaches (§5.2.4), categorize their failure cases (§5.2.5), and conclude by characterizing the attributes a pose estimation system should have to be applicable for SLR (§5.2.6).

## 5.2.2 Background

### Pose Estimation

Video-to-Pose—commonly known as pose estimation—is the task of detecting human figures in images and videos, so that one could determine, for example, where someone's elbow shows up in an image. It was shown that the face pose correlates with facial non-manual features like head direction (Vogler and Goldenstein, 2005).

This area has been thoroughly researched (Pishchulin et al., 2012; Chen et al., 2017; Cao et al., 2019; Güler et al., 2018) with objectives varying from predicting 2D / 3D poses to a selection of a small specific set of landmarks or a dense mesh of a person.

OpenPose (Cao et al., 2019; Simon et al., 2017; Cao et al., 2017; Wei et al., 2016) is the first multi-person system to jointly detect human body, hand, facial, and foot keypoints (in total 135 keypoints) in 2D on single images. While their model can estimate the full pose directly from an image in a single inference, they also suggest a pipeline approach where they first estimate the body pose and then independently estimate the hands and face pose by acquiring higher resolution crops around those areas. Building on the slow pipeline approach, a single network whole body OpenPose model has been proposed (Hidalgo et al., 2019), which is faster and more accurate for the case of obtaining all keypoints. With multiple recording angles, OpenPose also offers keypoint triangulation to reconstruct the pose in 3D.

DensePose (Güler et al., 2018) takes a different approach. Instead of classifying for every keypoint which pixel is most likely, they suggest a method similar to semantic segmentation, for each pixel to classify which body part it belongs to. Then, for each pixel, knowing the body part, they predict where that pixel is on the body part relative to a 2D projection of a representative body model. This approach results in the reconstruction of the full-body mesh and allows sampling to find specific keypoints similar to OpenPose.

However, 2D human poses might not be sufficient to fully understand the position and orientation of landmarks in space, and applying pose estimation per frame disregards video temporal movement information into account, especially in cases of rapid movement, which contain motion blur.

Pavllo et al. (2019) developed two methods to convert between 2D poses to 3D poses. The first, a supervised method, was trained to use the temporal information between frames to predict the missing Z-axis. The second is an unsupervised method, leveraging the fact that the 2D poses are merely a projection of an unknown 3D pose and training a model to estimate the 3D pose and back-project to the input 2D poses. This back projection is a deterministic process, applying constraints on the 3D pose encoder. Zelinka and Kanis (2020) followed a similar process and added a constraint for bones to stay of a fixed length between frames.

Panteleris et al. (2018) suggest converting the 2D poses to 3D using inverse kinematics (IK), a process taken from computer animation and robotics to calculate the variable joint parameters needed to place the end of a kinematic chain, such as a robot manipulator or animation character's skeleton, in a given position and orientation relative to the start of the chain. Demonstrating their approach to hand pose estimation, they manually explicitly encode the constraints and limits of each joint, resulting in 26 degrees of freedom. Then, nonlinear least-squares minimization fits a 3D model of the hand to the estimated 2D joint positions, recovering the 3D hand pose. This process is similar to the back-projection used by Pavllo et al. (2019), except here, no temporal information is being used.

MediaPipe Holistic (Grishchenko and Bazarevsky, 2020) attempts to solve 3D pose estimation by taking a similar approach to OpenPose, having a pipeline system to estimate the body, then the face and hands. Unlike OpenPose, the estimated poses are in 3D, and the pose estimator runs in real-time on CPU, allowing for pose-based sign language models on low-powered mobile devices. This pose estimation tool is widely available and built for Android, iOS, C++, Python, and the Web using JavaScript.

**Sign Language Recognition**

Sign language recognition (SLR) (Adaloglou et al., 2020) detects and labels signs from a video, either on isolated (Imashev et al., 2020; Sincan and Keles, 2020) or continuous Cui et al. (2017); Camgöz et al. (2018, 2020b) signs.

This task has been attempted both with computer vision models, assuming the input is the raw video, and with poses, assuming the video has been processed with a pose estimation tool.

**Video to Sign**    Video-to-Gloss, also known as sign language recognition, is the task of recognizing a sequence of signs from a video.

For this recognition, Cui et al. (2017) constructs a three-step optimization model. First, they train a video-to-gloss end-to-end model, where they encode the video using a spatio-temporal CNN encoder and predict the gloss using a Connectionist Temporal Classification (CTC) (Graves et al., 2006). Then, from the CTC alignment and category proposal, they encode each gloss-level seg-ment independently, trained to predict the gloss category, and use this gloss video segments encoding to optimize the sequence learning model.

Camgöz et al. (2018) fundamentally differ from that approach and formulate this problem as if it is a natural-language translation problem. They encode each video frame using AlexNet (Krizhevsky et al., 2012), initialized using weights trained on ImageNet (Deng et al., 2009). Then they apply a GRU encoder-decoder architecture with Luong Attention (Luong et al., 2015) to generate the gloss. In follow-up work, Camgöz et al. (2020b) use a transformer encoder (Vaswani et al., 2017) to replace the GRU and use a CTC to decode the gloss. They show a slight improvement with this approach on the video-to-gloss task.

Adaloglou et al. (2020) perform a comparative experimental assessment of computer vision-based methods for the video-to-gloss task. They implement various approaches from previous research (Camgöz et al., 2017; Cui et al., 2019; Vaezi Joze and Koller, 2019) and test them on multiple datasets (Huang et al., 2018; Camgöz et al., 2018; Von Agris and Kraiss, 2007; Vaezi Joze and Koller, 2019) either for isolated sign recognition or continuous sign recognition. They conclude that 3D convolutional models outperform models using only recurrent networks to capture the temporal information, and that these models are more scalable given the restricted receptive field, which results from the CNN "sliding window" technique.

Momeni et al. (2022) developed a comprehensive pipeline that combines various models to densely annotate sign language videos. By leveraging the use of synonyms and subtitle-signing alignment, their approach demonstrates the value of pseudo-labeling from a sign recognition model for sign spotting. They propose a novel method to increase annotations for both known and unknown classes, relying on in-domain exemplars. As a result, their framework significantly expands the number of confident automatic annotations on the BOBSL BSL sign language corpus (Albanie et al., 2021) from 670K to 5M, and they generously make these annotations publicly available.

**Pose to Sign** Pose-to-Gloss, also known as sign language recognition, is the task of recognizing a sequence of signs from a sequence of poses. Though some previous works have referred to this as "sign language translation," recognition merely determines the associated label of each sign, without handling the syntax and morphology of the signed language (Padden, 1988) to create a spoken language output. Instead, SLR has often been used as an intermediate step during translation to produce glosses from signed language videos.

Jiang et al. (2021) proposed a novel Skeleton Aware Multi-modal Framework with a Global Ensemble Model (GEM) for isolated SLR (SAM-SLR-v2) to learn and fuse multimodal feature representations. Specifically, they use a Sign Language Graph Convolution Network (SL-GCN) to model the embedded dynamics of skeleton keypoints and a Separable Spatial-Temporal Convolution Network (SSTCN) to exploit skeleton features. The proposed late-fusion GEM fuses the skeleton-based predictions with other RGB and depth-based modalities to provide global information and make an accurate SLR prediction.

Dafnis et al. (2022) work on the same modified WLASL dataset as Jiang et al. (2021), but do not require multimodal data input. Instead, they propose a bidirectional skeleton-based graph convolutional network framework with linguistically motivated parameters and attention to the start and end frames of signs. They cooperatively use forward and backward data streams, including various sub-streams, as input. They also use pre-training to leverage transfer learning.

Selvaraj et al. (2022) introduced an open-source OpenHands library, which consists of standardized pose datasets for different existing sign language datasets and trained checkpoints of four pose-based isolated sign language recognition models across six languages (American, Argentinian, Chinese, Greek, Indian, and Turkish). To address the lack of labeled data, they propose self-supervised pretraining on unlabeled data and curate the largest pose-based pretraining dataset on Indian Sign Language (Indian-SL). They established that pretraining is effective for sign language recognition by demonstrating improved fine-tuning performance especially in low-resource settings and high crosslingual transfer from Indian-SL to a few other sign languages.

The work of Kezar et al. (2023), based on the OpenHands library, explicitly recognizes the role of phonology to achieve more accurate isolated sign language recognition (ISLR). To allow additional predictions on phonological characteristics (such as handshape), they combine the phonological annotations in ASL-LEX 2.0 (Sehyr et al., 2021) with signs in the WLASL 2000 ISLR benchmark (Li et al., 2020). Interestingly, Tavella et al. (2022) construct a similar dataset aiming just for phonological property recognition in American Sign Language.

### 5.2.3   Experiments

To evaluate whether pose estimation models are applicable for SLR, we participated in the CVPR21 ChaLearn challenge for person-independent isolated SLR on the Ankara University Turkish Sign Language (AUTSL) Sincan and Keles (2020) dataset. Even though the dataset includes Kinect pose estimations, Kinect poses have not been made available for the challenge. We processed the dataset using two pose estimation tools: 1. OpenPose Single-Network Whole-Body Pose Estimation Hidalgo et al. (2019); and 2. MediaPipe Holistic Grishchenko and Bazarevsky (2020); and made the data available via an open-source sign language datasets repository Moryossef and Müller (2021).

We approach the recognition task with two independent experiments performed by different teams unaware of the other team's work throughout the validation stage. In the validation stage, each team focussed on one pose estimation approach, and in the test stage, both teams got access to both pose estimation outputs. We eventually submitted three systems: 1. based on *Open-Pose* poses; 2. based on *Holistic* poses; 3. based on both *OpenPose* and *Holistic* poses combined (concatenated).

#### Team 1

*Team 1* worked with OpenPose Hidalgo et al. (2019) pose estimation output and used the SLR transformer architecture from Camgöz et al. Camgöz et al. (2020b). The model takes as input a series of feature vectors, in this case, human upper body skeletal coordinates extracted from the video frames. These are each projected to a lower dimension hidden state vector. The size of the hidden state remains constant throughout the subsequent operations. A sinusoidal positional encoding is added to provide temporal information. This is then passed to a subnetwork consisting of a multiheaded self-attention layer, followed by a feedforward layer. After each of these layers, the output is added to the input and normalized. This subnetwork can be repeated any number of times. Finally, the output is fed to a linear layer and softmax to give probabilities for each class (Figure 5.5).

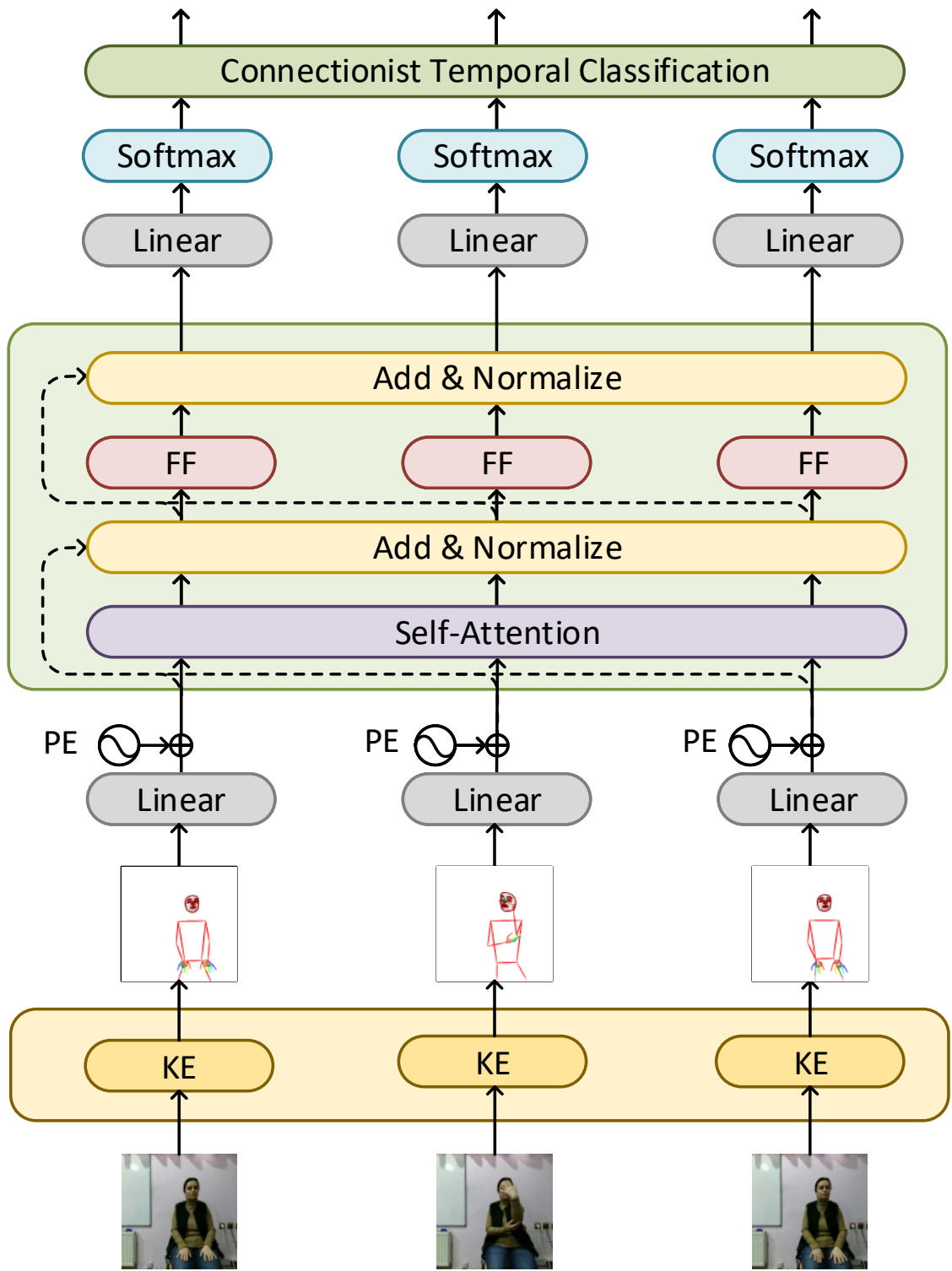

Figure 5.5: Diagram of *Team 1*'s model with one subnetwork (in green). (KE: Keypoint extraction, PE: Positional encoding, FF: feed forward)

The model is trained using CTC loss. This is designed to allow the output to be invariant to alignment; however, this is not a significant concern when there should only be one output symbol. The final prediction is obtained via CTC beam search decoding, collapsing multiple same class outputs into one. As the model is trained to predict a single class per video, it does not predict different classes within a sequence.

The number of layers, heads, hidden size, and dropout rate affect the model complexity. There is, therefore, a tradeoff between sufficient complexity to model the data and overfitting.

Additionally, as a baseline, the pose estimation keypoints were replaced with the output of three off-the-shelf image-based frame feature extractors, giving us small dense representations for each frame. Three extractors were used: 1. EfficientNet-B7 Tan and Le (2019); 2. I3D trained on Kinetics Carreira and Zisserman (2017); and 3. I3D trained on BSL1K Albanie et al. (2020).

**Team 2**

*Team 2* worked with the MediaPipe Holistic Grishchenko and Bazarevsky (2020) pose estimation system output. From the 543 landmarks, the face mesh was removed which consists of 468 landmarks and the remaining 75 landmarks were used for the body and hands.

A standard sequence classification architecture was used. The model takes as input a series of feature vectors, constructed from a flat vector representation of the pose concatenated with the 2D angle and length of every limb, using the *pose-format*[3] library. These representations are subjected to a 20% dropout, normalized using 1D batch normalization, and are projected to a lower dimension hidden state vector (512 dimensions). This is then passed to a two-layer BiLSTM with hidden dimension 256, followed by a max-pooling operation to obtain a single representation vector per video. Finally, the output is fed to a linear layer and softmax to give probabilities for each class (Figure 5.6).

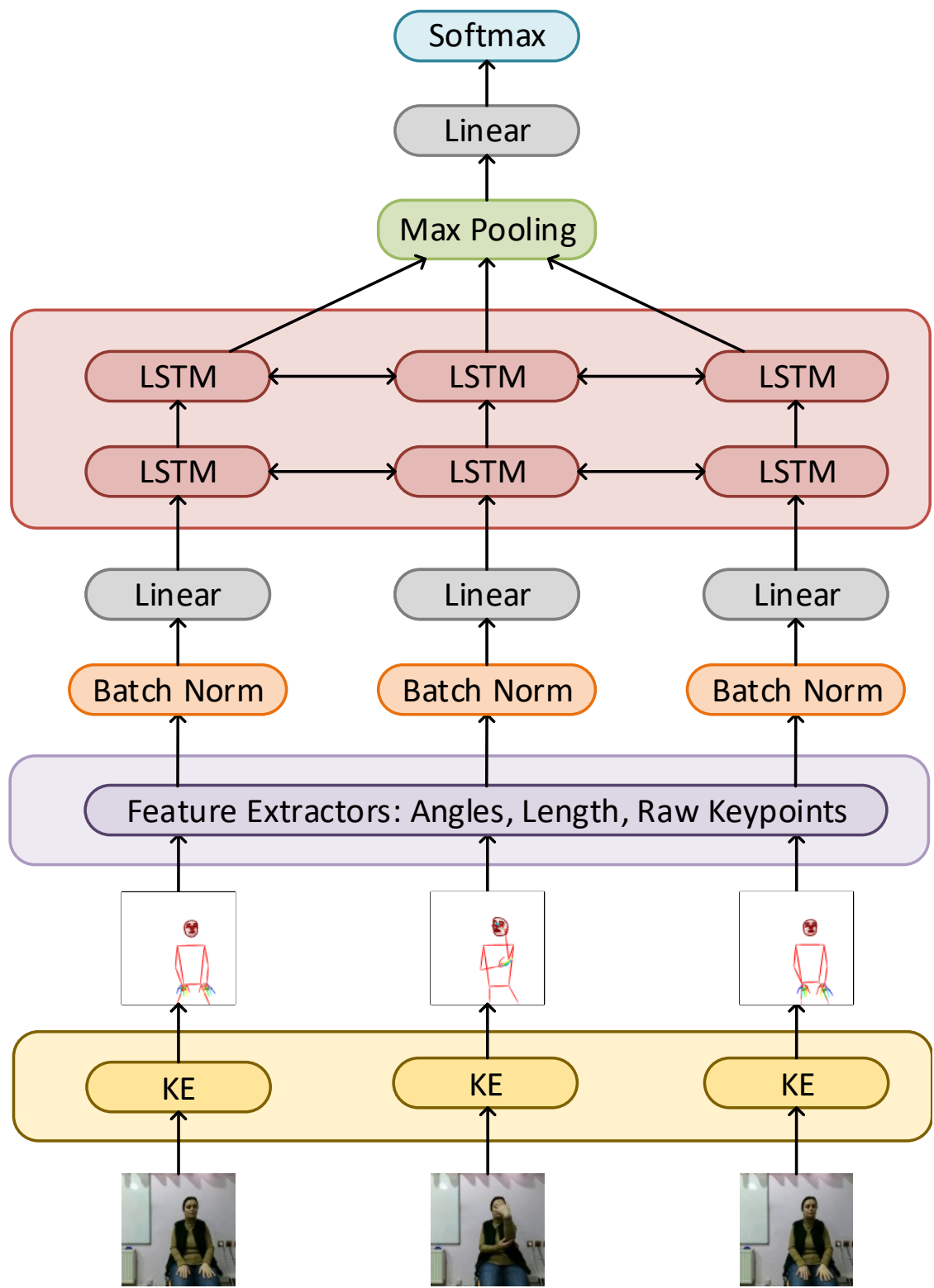

Figure 5.6: Diagram of *Team 2*'s model. (KE: Keypoint extraction)

The model is trained using cross-entropy loss with the Adam optimizer (with default parameters) and a batch size of 512 on a single GPU. *No* data augmentation or frame dropout is applied at training time, except for horizontal frame flip to account for left-handed signers in the dataset.

### 5.2.4   Results

Table 5.3 shows our teams' results on the validation set. We note that both teams' approaches using pose estimation performed similarly, with validation accuracy ranging between 80% and 85%. It rules out trivial errors and implementation issues that, despite working independently, and with two separate pose estimation tools, both teams achieve similar evaluation scores. Furthermore, from a comparison between the pose estimation based systems (80-85%) and the pretrained image feature extractors (38-68%), we can see that pose estimation features do indeed generalize better to the nature of the challenge, including unseen signers and backgrounds.

We submitted *Team 2*'s test set predictions to the official challenge evaluation. On the test set, both *OpenPose* and *Holistic* performed **equally well** despite making different predictions, each with 78.35% test set accuracy. However, our combined system, which was trained using both pose estimations, achieves 81.93% test set accuracy.

---

[3]https://github.com/AmitMY/pose-format

|  | Team 1 | Team 2 |
|---|---|---|
| EfficientNet-B7 | 38.80% | — |
| I3D (Kinetics) | 47.46% | — |
| I3D (BSL1K) | 68.65% | — |
| OpenPose | 83.25% | 79.99% |
| Holistic | 85.63% | 82.14% |
| OpenPose+Holistic | 84.16% | 82.89% |

Table 5.3: Results evaluated on the validation set with various frame-level features.

### 5.2.5 Analysis

The interpretability of skeletal poses allows us to assess them qualitatively using visualisation. We manually review our model's failure cases and categorize them into two main categories: hands interaction and hand-face interaction.

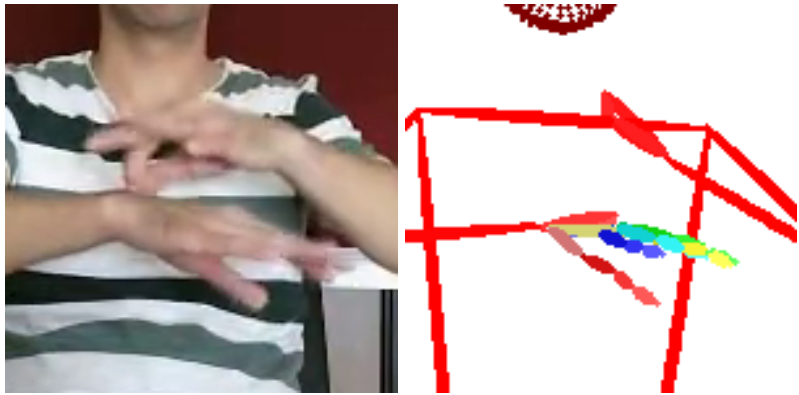

Figure 5.7: Example of hands interaction, where the pose estimation fails for one of the hands (Holistic).

**Hands Interaction**   When there exists an interaction between both hands, or one hand occludes the other from the camera's view, we often fail to estimate the pose of one of the hands (Figure 5.7) or estimate it incorrectly such that the interaction is not clearly shown (Figure 5.8).

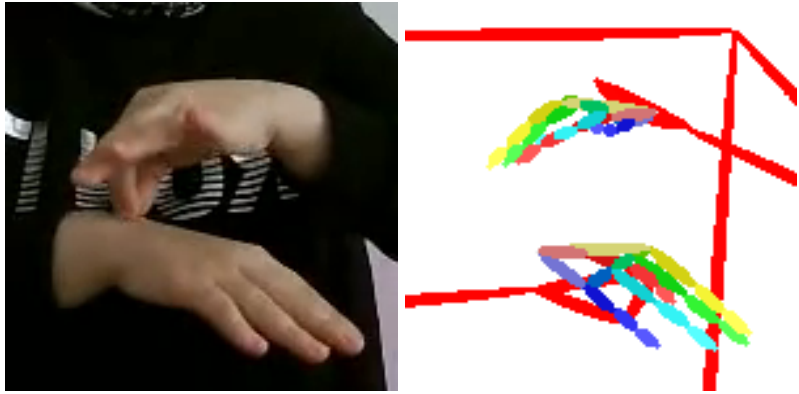

Figure 5.8: Example of hands interaction, where the pose estimation does not reflect the existing interaction (Holistic).

**Hand-Face Interaction**   When there exists an interaction between a hand and the face, or one hand overlaps with the face from the camera's angle, we often fail to estimate the pose of the interacting hand (Figure 5.9).

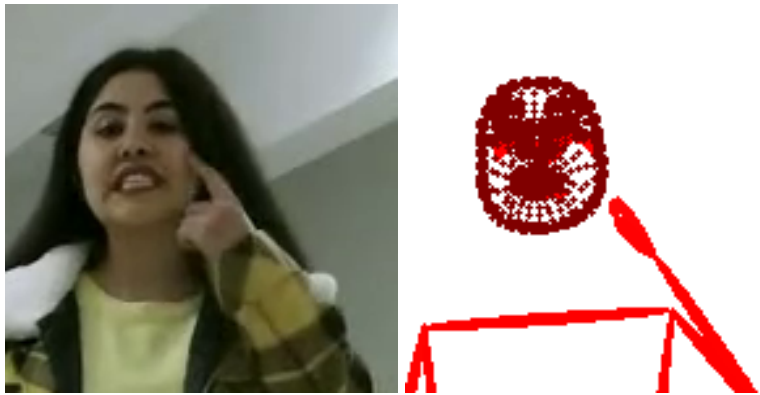

Figure 5.9: Example of hand-face interaction, where the pose estimation fails for the interacting hand (Holistic).

These cases of missed interactions between the different body parts often lose the essence of the sign, where the interaction and the hand shape are the main distinguishing features for those signs, and thus hinder the model's ability to extract meaningful information from the pose that is relevant to the sign.

**Presence or absence of hand pose**   We describe a number of failure cases of Holistic pose estimation above.  Many of them mean that keypoints for the hands are not available at all, since Holistic can omit them if it fails to detect the hand. As a complementary quantitative analysis, we correlate prediction outcomes with the average number of frames where hand pose was present (Figure 5.10).

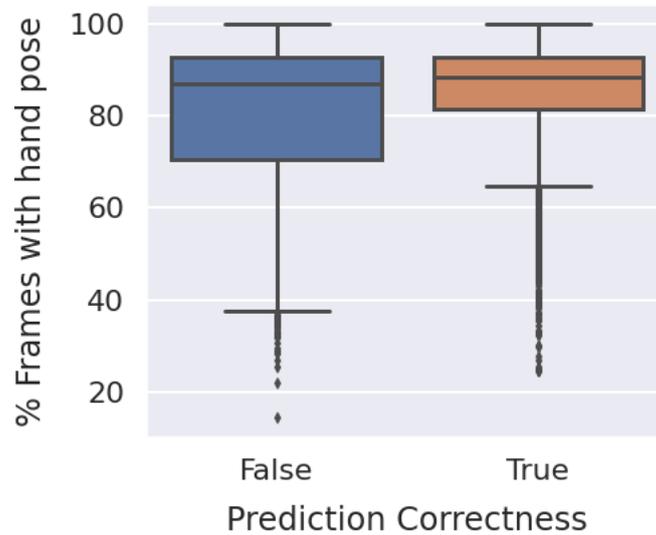

Figure 5.10: Distribution of percent of frames containing the Holistic pose estimation of the dominant hand in each validation sample, grouped by whether the final prediction of our model was correct.

We find that on average, for all correct predictions the percentage of frames that do contain hand keypoints (85.13%) is significantly higher[4] than for all incorrect predictions (79.78%). This is in line with our qualitative analysis.

### 5.2.6 Conclusions

Although many teams outperformed our models that use only off-the-shelf skeletal representations, with the best submission reaching 98.4% test set accuracy, it is unclear how well such approaches will generalise to other datasets. Our initial questions related to how good skeletal representations are for recognition, given their natural ability to generalise. However, performance in the ChaLearn challenge suggests that despite their benefits, considerable information is lost in the skeletal representation that must be represented in the image domain. A qualitative analysis of our models' failure cases shows that pose estimation tools suffer from shortcomings when body parts interact. We conclude that pose estimation tools are not immediately applicable for the use in sign language recognition – the current representations are not sufficiently expressive, and that further improvements with regard to interacting body parts is crucial for their applicability.

---

[4]We tested for a significant difference of the mean values with a Wilcoxon rank-sum test Wilcoxon (1992), $p < 0.0001$.

### 5.2.7 Discussion

As shown in Figure 5.10, improved pose estimation strongly correlates with better prediction accuracy. Since this work's publication in 2021, more advanced models have emerged, surpassing MediaPipe Holistic and OpenPose, such as Meta's Sapiens model (Khirodkar et al., 2024). These new models significantly enhance pose estimation and may offer substantial improvements over the results presented here. By 2024, these advancements could make pose estimation more applicable to sign language recognition.

## 5.3 Gloss Translation (Moryossef et al., 2021c)

In this section, we explore the maximum potential of translating from signed to spoken language using glosses, based on the assumption of flawless sign language recognition. Among other things, the findings reveal that glosses alone cannot fully encapsulate the nuances of sign language.

Sign language translation (SLT) is often decomposed into *video-to-gloss* recognition and *gloss-to-text* translation, where a gloss is a sequence of transcribed spoken-language words in the order in which they are signed. We focus here on gloss-to-text translation, which we treat as a low-resource neural machine translation (NMT) problem. However, unlike traditional low-resource NMT, gloss-to-text translation differs because gloss-text pairs often have a higher lexical overlap and lower syntactic overlap than pairs of spoken languages. We exploit this lexical overlap and handle syntactic divergence by proposing two rule-based heuristics that generate pseudo-parallel gloss-text pairs from monolingual spoken language text. By pre-training on this synthetic data, we improve translation from American Sign Language (ASL) to English and German Sign Language (DGS) to German by up to 3.14 and 2.20 BLEU, respectively.

### 5.3.1 Introduction

Sign language is the most natural mode of communication for the Deaf. However, in a predominantly hearing society, they often resort to lip-reading, text-based communication, or closed-captioning to interact with others. Sign language translation (SLT) is an important research area that aims to improve communication between signers and non-signers while allowing each party to use their preferred language. SLT consists of translating a sign language (SL) video into a spoken language (SpL) text, and current approaches often decompose this task into two steps: (1) *video-to-gloss*, or continuous sign language recognition (CSLR) (Cui et al., 2017; Camgöz et al., 2018); (2) *gloss-to-text*, which is a text-to-text machine translation (MT) task (Camgöz et al., 2018; Yin and Read, 2020a).

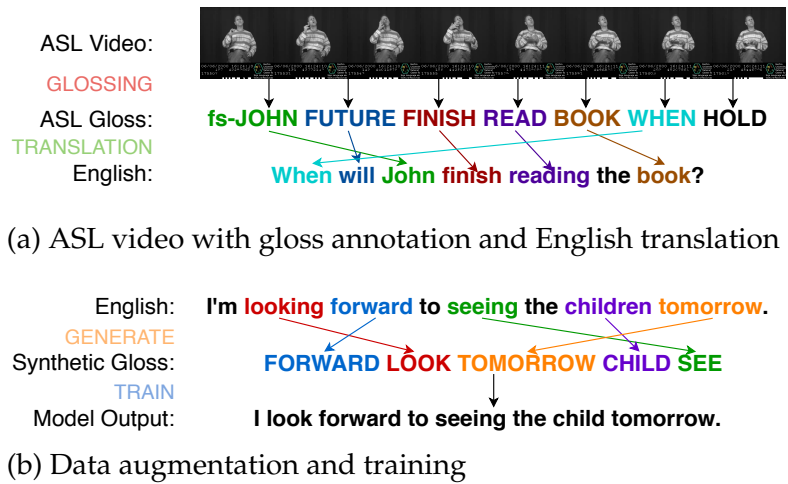

(a) ASL video with gloss annotation and English translation

(b) Data augmentation and training

Figure 5.11: Real and synthetic gloss-spoken pairs.

In this paper, we focus on gloss-to-text translation. SL data and resources are often scarce, or nonexistent (§5.3.2; Bragg et al. (2019)). Gloss-to-text translation is, therefore, an example of an extremely low-resource MT task. However, while there is extensive literature on low-resource MT between spoken languages (Sennrich et al., 2016a; Zoph et al., 2016; Xia et al., 2019; Zhou et al., 2019), the dissimilarity between sign and spoken languages calls for novel methods. Specifically, as SL glosses borrow the lexical elements from their ambient spoken language, handling syntax and morphology poses greater challenges than lexeme translation (§5.3.4).

In this work, we (1) discuss the scarcity of SL data and quantify how the relationship between a sign and spoken language pair is different from a pair of two spoken languages; (2) show that the *de facto* method for data augmentation using back-translation is not viable in extremely low-resource SLT; (3) propose two rule-based heuristics that exploit the lexical overlap and handles the syntactic divergence between sign and spoken language, to synthesize pseudo-parallel gloss/text examples (Figure 5.11b); (4) demonstrate the effectiveness of our methods on two sign-to-spoken language pairs.

### 5.3.2 Background

**Sign Language Glossing**  SLs are often transcribed word-for-word using a spoken language through *glossing* to aid in language learning, or automatic sign language processing (Ormel et al., 2010). While many SL glosses are words from the ambient spoken language, glossing preserves SL's original syntactic structure and therefore differs from translation (Figure 5.11a).

**Data Scarcity**  While standard machine translation architectures such as the Transformer (Vaswani et al., 2017) achieve reasonable performance on gloss-to-text datasets (Yin and Read, 2020b; Camgöz et al., 2020b), parallel SL and spoken language corpora, especially those with gloss annotations, are usually far more scarce when compared with parallel corpora that exist between many spoken languages (Table 5.4).

| | Language Pair | # Parallel Gloss-Text Pairs | Vocabulary Size (Gloss / Spoken) |
|---|---|---|---|
| Signum (Von Agris and Kraiss, 2007) | DGS-German | 780 | 565 / 1,051 |
| NCSLGR (SignStream, 2007) | ASL-English | 1,875 | 2,484 / 3,104 |
| RWTH-PHOENIX-Weather 2014T (Camgöz et al., 2018) | DGS-German | 7,096 + 519 + 642 | 1,066 / 2,887 + 393 / 951 + 411 / 1,001 |
| Dicta-Sign-LSF-v2 (Limsi, 2019) | French SL-French | 2,904 | 2,266 / 5,028 |
| The Public DGS Corpus (Hanke et al., 2020) | DGS-German | 63,912 | 4,694 / 23,404 |

Table 5.4: Some publicly available SL corpora with gloss annotations and spoken language translations.

### 5.3.3 Previous Work

Gloss-to-Text, also known as sign language translation, is the natural language processing task of translating between gloss text representing sign language signs and spoken language text. These texts commonly differ in terminology, capitalization, and sentence structure.

Camgöz et al. (2018) experimented with various machine-translation architectures and compared using an LSTM (Hochreiter and Schmidhuber, 1997) vs. GRU for the recurrent model, as well as Luong attention (Luong et al., 2015) vs. Bahdanau attention (Bahdanau et al., 2015) and various batch sizes. They concluded that on the RWTH-PHOENIX-Weather-2014T dataset, which was also presented in this work, using GRUs, Luong attention, and a batch size of 1 outperforms all other configurations.

In parallel with the advancements in spoken language machine translation, Yin and Read (2020a) proposed replacing the RNN with a Transformer (Vaswani et al., 2017) encoder-decoder model, showing improvements on both RWTH-PHOENIX-Weather-2014T (DGS) and ASLG-PC12 (ASL) datasets both using a single model and ensemble of models. Interestingly, in gloss-to-text, they show that using the sign language recognition (video-to-gloss) system output outperforms using the gold annotated glosses.

### 5.3.4   Signed vs. Spoken Language

Due to the paucity of parallel data for gloss-to-text translation, we can treat it as a low-resource translation problem and apply existing techniques for improving accuracy in such settings. However, we argue that the relationship between glossed SLs and their spoken counterparts is different from the usual relationship between two spoken languages. Specifically, glossed SLs are *lexically similar but syntactically different* from their spoken counterparts. This contrasts heavily with the relationship among spoken language pairs where lexically similar languages tend also to be syntactically similar the great majority of the time.

To demonstrate this empirically, we adopt measures from (Lin et al., 2019) to measure the lexical and syntactic similarity between languages, two features also shown to be positively correlated with the effectiveness of performing cross-lingual transfer in MT.

**Lexical similarity**   between two languages is measured using word overlap:

$$o_w = \frac{|T_1 \cap T_2|}{|T_1| + |T_2|}$$

where $T_1$ and $T_2$ are the sets of types in a corpus for each language. The word overlap between spoken language pairs is calculated using the TED talks dataset (Qi et al., 2018). The overlap between sign-spoken language pairs is calculated

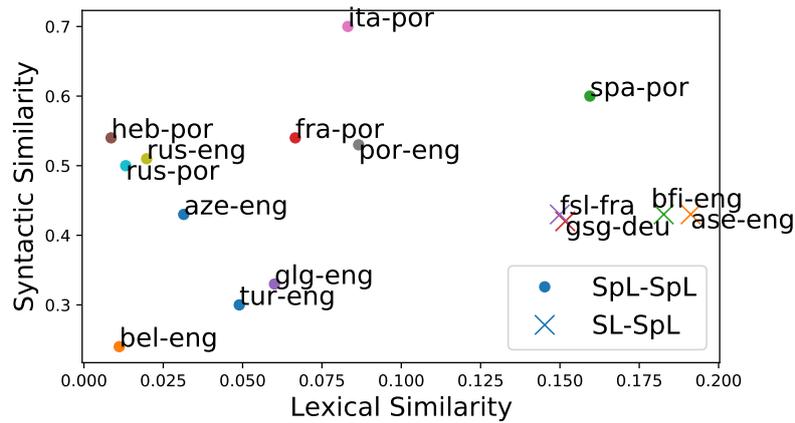

Figure 5.12: Lexical and syntactic similarity between different language pairs denoted by their ISO639-2 codes.

from the corresponding corpora in Table 5.4.

**Syntactic similarity** between two languages is measured by $1 - d_{syn}$ where $d_{syn}$ is the syntactic distance from (Littell et al., 2017) calculated by taking the cosine distance between syntactic features adapted from the World Atlas of Language Structures (Dryer and Haspelmath, 2013).

Figure 5.12 shows that sign-spoken language pairs are indeed outliers with lower syntactic similarity and higher lexical similarity. We seek to leverage this fact and the high availability of monolingual spoken language data to compensate for the scarcity of SL resources. In the following section, we propose data augmentation techniques using word order modifications to create synthetic sign gloss data from spoken language corpora.

### 5.3.5 Data Augmentation

This section discusses methods to improve gloss-to-text translation through data augmentation, specifically those that take monolingual corpora of standard spoken languages and generate pseudo-parallel "gloss" text. We first discuss back-translation, point out its potential failings in the SL setting, and then propose a novel rule-based data augmentation algorithm.

### Back-translation

Back-translation (Irvine and Callison-Burch, 2013; Sennrich et al., 2016a) automatically creates pseudo-parallel sentence pairs from monolingual text to improve MT in low-resource settings. However, back-translation is only effective with sufficient parallel data to train a functional MT model, which is not always the case in extremely low-resource settings (Currey et al., 2017), and particularly when the domain of the parallel training data and monolingual data to be translated are mismatched (Dou et al., 2020).

### Proposed Rule-based Augmentation Strategies

Given the limitations of standard back-translation techniques, we next move to the proposed method of using rule-based heuristics to generate SL glosses from spoken language text.

**General rules**  The differences in SL glosses from spoken language can be summarized by (1) A lack of word inflection, (2) An omission of punctuation and individual words, and (3) Syntactic diversity.

We, therefore, propose the corresponding three heuristics to generate pseudo-glosses from spoken language: (1) Lemmatization of spoken words; (2) POS-dependent and random word deletion; (3) Random word permutation.

We use spaCy (Honnibal and Montani, 2017) for (1) lemmatization and (2) POS tagging to only keep nouns, verbs, adjectives, adverbs, and numerals. We also drop the remaining tokens with probability $p = 0.2$, and (3) randomly reorder tokens with maximum distance $d = 4$.

For a given sentence $\mathcal{S}$:

1. Discard all tokens $t \in \mathcal{S}$ if **POS**$(t) \notin \{$noun, verb, adjective, adverb, numeral$\}$

2. Discard remaining tokens $t \in \mathcal{S}$ with probability $p = 0.2$

3. Lemmatize all tokens $t \in \mathcal{S}$

4. Apply a random permutation $\sigma$ to $\mathcal{S}$ verifying $\forall i \in \{1, n\}, |\sigma(i) - i| \leq 4$

where $n$ is the number of tokens in $\mathcal{S}$ at step 4 and **POS** is a part-of-speech tagger.

**Language-specific rules**   While random permutation allows some degree of robustness to word order, it cannot capture all aspects of syntactic divergence between signed and spoken language. Therefore, inspired by previous work on rule-based syntactic transformations for reordering in MT (Collins et al., 2005; Isozaki et al., 2010; Zhou et al., 2019), we manually devise a shortlist of syntax transformation rules based on the grammar of DGS and German.

We perform lemmatization and POS filtering as before. In addition, we apply compound splitting (Tuggener, 2016) on nouns and only keep the first noun, reorder German SVO sentences to SOV, move adverbs and location words to the start of the sentence, and move negation words to the end.

For a given sentence $\mathcal{S}$:

1. For each subject-verb-object triplet $(s, v, o) \in \mathcal{S}$, swap the positions of $v$ and $o$ in $\mathcal{S}$

2. Discard all tokens $t \in \mathcal{S}$ if **POS**$(t) \notin \{$noun, verb, adjective, adverb, numeral$\}$

3. For $t \in \mathcal{S}$, if **POS**$(t) =$ adverb, then move $t$ to the start of $s$

4. For $t \in \mathcal{S}$, if **NER**$(t) =$ location, then move $t$ to the start of $s$

5. For $t \in \mathcal{S}$, if **DEP**$(t) =$ negation, then move $t$ to the end of $s$

6. For $t \in \mathcal{S}$, if $t$ is a compound noun $c_1 c_2 ... c_n$, replace $t$ by $c_1$

7. Lemmatize all tokens $t \in \mathcal{S}$

where **POS** is a part-of-speech tagger, **NER** is a named entity recognizer and **DEP** is a dependency parser.

## 5.3.6 Experimental Setting

**Datasets**

**DGS & German**   RWTH-PHOENIX-Weather 2014T (Camgöz et al., 2018) is a parallel corpus of 8,257 DGS interpreted videos from the Phoenix[5] weather news channel, with corresponding SL glosses and German translations.

To obtain monolingual German data, we crawled tagesschau[6] and extracted news caption files containing the word "wetter" (German for "weather"). We split the 1,506 caption files into 341,023 German sentences using the spaCy sentence splitter and generate synthetic glosses using our methods described in §5.3.5.

**ASL & English**   The NCSLGR dataset (SignStream, 2007) is a small, general domain dataset containing 889 ASL videos with 1,875 SL glosses and English translations.

We use ASLG-PC12 (Othman and Jemni, 2012), a large synthetic ASL gloss dataset created from English text using rule-based methods with 87,710 publicly available examples, for our experiments on ASL-English with language-specific rules. We also create another synthetic variation of this dataset using our proposed general rule-based augmentation.

**Baseline Setup**

We first train a **Baseline** system on the small manually annotated SL dataset we have available in each language pair. The model architecture and training method are based on Yin and Read (2020a)'s Transformer gloss-to-text trans-

---

lation model. While previous work (**Yin and Read** **Reimpl.**) used word-level tokenization, for Baseline and all other models described below, we instead use BPE tokenization (Sennrich et al. (2016b); with 2,000 BPE codes) for efficiency and simple handling of unknown words. For all tested methods, we repeat every experiment 3 times to account for variance in training.

**Model Reproduction**

For reproduction purposes, here we lay the exact commands for training a single model using OpenNMT 1.2.0 (Klein et al., 2017). These commands are taken from Yin and Read (2020a).

Given 6 files—*train.gloss / train.txt*, *dev.gloss / dev.txt*, *test.gloss / test.txt*—we start by preprocessing the data using the following command:

```
onmt_preprocess -dynamic_dict -save_data processed_data \
-train_src train.gloss -train_tgt train.txt \
-valid_src dev.gloss -valid_tgt dev.txt
```

Then, we train a translation system using the train command:

```
onmt_train -data processed_data -save_model model -layers 2 \
-rnn_size 512 -word_vec_size 512 -heads 8 -encoder_type transformer \
-decoder_type transformer -position_encoding -transformer_ff 2048 \
-dropout 0.1 -early_stopping 3 -early_stopping_criteria accuracy ppl \
-batch_size 2048 -accum_count 3 -batch_type tokens \
-max_generator_batches 2 -normalization tokens \
-optim adam -adam_beta2 0.998 -decay_method noam -warmup_steps 3000 \
-learning_rate 0.5 -max_grad_norm 0 -param_init 0 -param_init_glorot \
-label_smoothing 0.1 -valid_steps 100 -save_checkpoint_steps 100 \
-world_size 1 -gpu_ranks 0
```

At the end of the training procedure, it prints to console "Best model found at step X". Locate it, and use it for translating the data:

```
onmt_translate -model model_step_X.pt -src test.gloss \
-output hyp.txt -gpu 0 -replace_unk -beam_size 4
```

Finally, evaluate the output using SacreBLEU:

```
cat hyp.txt | sacrebleu test.txt
```

**Pre-training on Augmented Data**

For **General-*pre*** and **Specific-*pre***, we pre-train a tokenizer and translation model on pseudo-parallel data obtained using general and language-specific rules respectively, until the accuracy on the synthetic validation set drops. We test both models on the parallel SL dataset in a zero-shot setting.

For **BT-*tuned***, **General-*tuned*** and **Specific-*tuned***, we take models pre-trained on pseudo-parallel data obtained with either back-translation, general rules, or language-specific rules, and continue training with half of the training data taken from the synthetic pseudo-parallel data and the other half taken from the real SL data. Then, we fine-tune these models on the real SL data and evaluate them on the test set.

### 5.3.7 Results

We evaluate our models across all datasets and sizes using SacreBLEU (v1.4.14) ([Post, 2018](#)) and COMET (*wmt-large-da-estimator-1719*) ([Rei et al., 2020](#)). We also compare our results to previous work on PHOENIX in Table [5.5](#).

First, we note results on General-*pre* and Specific-*pre*. Interestingly, the scores are non-negligible, demonstrating that the model can learn with *only* augmented data.[7] Moreover, on PHOENIX Specific-*pre* achieves significantly better performance than General-*pre*, which suggests our hand-crafted syntax transformations effectively expose the model to the divergence between DGS and German during pre-training.

Next, turning to the *tuned* models, we see that Specific and General outperform both the baseline and BT by large margins, demonstrating the effective-

---

[7]In contrast, merely outputting the source sentence results in 1.36 BLEU, -90.28 COMET on PHOENIX and 1.5 BLEU, -119.45 COMET on NCSLGR.

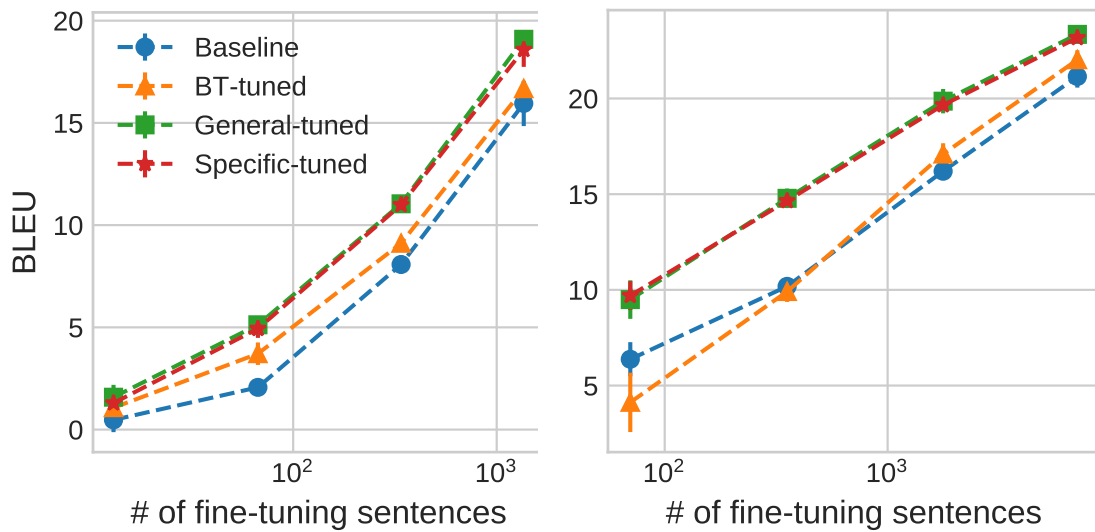

<div align="center">(a) NCSLGR (ASL)          (b) PHOENIX (DGS)</div>

Figure 5.13: Translation results using various amounts of annotated parallel data.

ness of our proposed methods. Interestingly, General-*tuned* performs slightly better, in contrast to the previous result. We posit that, similarly to previously reported results on sampling-based back translation (Edunov et al., 2018), General is benefiting from the diversity provided by sampling multiple reordering candidates, even if each candidate is of lower quality.

Looking at Figure 5.13, we see that the superior performance of our methods holds for all data sizes, but it is particularly pronounced when the parallel-data-only baseline achieves moderate BLEU scores in the range of 5-20. This confirms that BT is not a viable data augmentation method when parallel data is not plentiful enough to train a robust back-translation system.

Table 5.6 includes the evaluation scores for all of our experiments, ran three times.

---

[8]The original work achieves 23.32 BLEU; correspondence with the authors has led us to believe that the discrepancy is due to different versions of the underlying software.

|  | PHOENIX | | NCSLGR | |
|---|---|---|---|---|
|  | BLEU↑ | COMET↑ | BLEU↑ | COMET↑ |
| Yin and Read Reimpl.[8] | 22.17 | -2.93 | - | - |
| Baseline | 21.15 | -5.74 | 15.95 | -61.00 |
| General-*pre* (0-shot) | 3.95 | -69.09 | 0.97 | -135.99 |
| Specific-*pre* (0-shot) | 7.26 | -53.14 | 0.95 | -134.13 |
| BT-*tuned* | **22.02** | **6.84** | 16.67 | **-51.86** |
| General-*tuned* | **23.35** | **13.65** | **19.09** | **-34.50** |
| Specific-*tuned* | **23.17** | **11.70** | **18.5**8 | **-39.96** |

Table 5.5: Results of our models on PHOENIX and NCSLGR. We **bold** scores statistically significantly higher than baseline at the 95% confidence level.

| % of available annotated data used | | 1% | | 5% | | 25% | | 100% | |
|---|---|---|---|---|---|---|---|---|---|
|  | | BLEU | COMET | BLEU | COMET | BLEU | COMET | BLEU | COMET |
| PHOENIX | Baseline | 6.37 ± 0.89 | -89.21 ± 12.82 | 10.18 ± 0.40 | -71.37 ± 2.86 | 16.20 ± 0.27 | -33.88 ± 4.35 | 21.15 ± 0.58 | -5.74 ± 2.35 |
|  | BT-*tuned* | 4.12 ± 1.55 | -91.87 ± 16.35 | 9.91 ± 0.54 | **-53.38 ± 4.04** | **17.10 ± 0.56** | **-16.46 ± 2.52** | **22.02 ± 0.50** | **6.84 ± 0.34** |
|  | General-*tuned* | **9.49 ± 1.01** | **-52.23 ± 6.31** | **14.78 ± 0.51** | **-27.13 ± 2.29** | **19.86 ± 0.64** | **-0.72 ± 2.44** | **23.35 ± 0.22** | **13.65 ± 1.68** |
|  | Specific-*tuned* | **9.70 ± 0.75** | **-55.94 ± 2.08** | **14.65 ± 0.29** | **-30.85 ± 1.45** | **19.66 ± 0.08** | **-5.62 ± 0.51** | **23.17 ± 0.30** | **11.70 ± 1.20** |
| NCSLGR | Baseline | 0.47 ± 0.60 | -153.90 ± 11.89 | 2.07 ± 0.32 | -145.14 ± 1.15 | 8.07 ± 0.43 | -101.24 ± 5.14 | 15.95 ± 1.11 | -61.00 ± 6.86 |
|  | BT-*tuned* | 1.07 ± 0.47 | **-139.80 ± 3.78** | **3.71 ± 0.55** | **-117.33 ± 3.03** | **9.11 ± 0.05** | **-82.41 ± 2.29** | 16.67 ± 0.32 | **-51.86 ± 0.66** |
|  | General-*tuned* | 1.58 ± 0.60 | **-134.22 ± 1.73** | **5.13 ± 0.15** | **-106.59 ± 1.56** | **11.04 ± 0.04** | **-66.35 ± 2.00** | **19.09 ± 0.20** | **-34.50 ± 1.19** |
|  | Specific-*tuned* | 1.30 ± 0.52 | **-128.14 ± 1.58** | **4.94 ± 0.45** | **-107.60 ± 4.01** | **10.99 ± 0.12** | **-71.37 ± 1.01** | **18.58 ± 0.84** | **-39.96 ± 1.91** |

Table 5.6: Mean and standard deviation of BLEU and COMET over different experimental settings. We **bold** scores statistically significantly higher than baseline at the 95% confidence level.

### 5.3.8 Implications and Future Work

Consistent improvements over the baseline across two language pairs by our proposed rule-based augmentation strategies demonstrate that data augmentation using monolingual spoken language data is a promising approach for sign language translation.

Given the efficiency of our general rules compared to language-specific rules, future work may also include a more focused approach on specifically pre-training the target-side decoder with spoken language sentences so that by learning the syntax of the target spoken language, it can generate fluent sentences from sign language glosses having little to no parallel examples during training (both few-shot and zero-shot settings).

# Chapter 6

# Sign Language Translation

## 6.1 Segmentation (Moryossef et al., 2023a)

Sign language segmentation is a crucial task in sign language processing systems. It enables downstream tasks such as sign recognition, transcription, and machine translation. In this work, we consider two kinds of segmentation: into individual signs and into *phrases*, larger units comprising several signs. We propose a novel approach to jointly model these two tasks.

Our method is motivated by linguistic cues observed in sign language corpora. We replace the predominant IO tagging scheme with BIO tagging to account for continuous signing. Given that prosody plays a significant role in phrase boundaries, we explore the use of optical flow features. We also provide an extensive analysis of hand shapes and 3D hand normalization.

We find that introducing BIO tagging is necessary to model sign boundaries. Explicitly encoding prosody by optical flow improves segmentation in shallow models, but its contribution is negligible in deeper models. Careful tuning of the decoding algorithm atop the models further improves the segmentation quality.

We demonstrate that our final models generalize to out-of-domain video content in a different signed language, even under a zero-shot setting. We observe that including optical flow and 3D hand normalization enhances the robustness of the model in this context.

### 6.1.1 Introduction

Signed languages are natural languages used by deaf and hard-of-hearing individuals to communicate through a combination of manual and non-manual elements (Sandler and Lillo-Martin, 2006). Like spoken languages, signed languages have their own distinctive grammar, and vocabulary, that have evolved through natural processes of language development (Sandler, 2010).

Sign language transcription and translation systems rely on the accurate temporal segmentation of sign language videos into meaningful units such as signs (Santemiz et al., 2009; Renz et al., 2021a) or signing sequences corresponding to subtitle units[1] (Bull et al., 2020b). However, sign language segmentation remains a challenging task due to the difficulties in defining meaningful units in signed languages (De Sisto et al., 2021). Our approach is the first to consider two kinds of units in one model. We simultaneously segment single signs and phrases (larger units) in a unified framework.

Previous work typically approached segmentation as a binary classification task (including segmentation tasks in audio signal processing and computer vision), where each frame/pixel is predicted to be either part of a segment or not. However, this approach neglects the intricate nuances of continuous signing, where segment boundaries are not strictly binary and often blend in reality. One sign or phrase can immediately follow another, transitioning smoothly, without a frame between them being distinctly *outside* (Figure 6.1 and §6.1.3).

We propose incorporating linguistically motivated cues to address these challenges and improve sign language segmentation. To cope with continuous signing, we adopt a BIO-tagging approach (Ramshaw and Marcus, 1995), where in addition to predicting a frame to be *in* or *out* of a segment, we also classify the *beginning* of the segment as shown in Figure 6.2. Since phrase segmentation is primarily marked with prosodic cues (i.e., pauses, extended sign duration, facial expressions) (Sandler, 2010; Ormel and Crasborn, 2012), we explore using optical flow to explicitly model them. Since signs employ a limited number of hand shapes, we additionally perform 3D hand normalization (§6.1.3).

---

[1]Subtitles may not always correspond directly to sentences. They frequently split within a sentence and could be temporally offset from the corresponding signing segments.

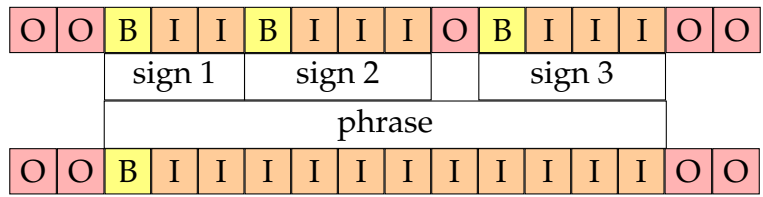

Figure 6.1: Per-frame classification of a sign language utterance following a BIO tagging scheme. Each box represents a single frame of a video. We propose a joint model to segment *signs* (top) and *phrases* (bottom) at the same time. B=beginning, I=inside, O=outside. The figure illustrates continuous signing where signs often follow each other without an O frame between them.

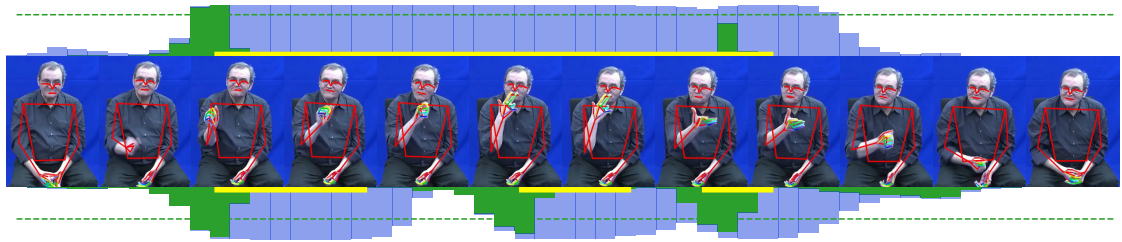

Figure 6.2: The annotation of the first phrase in a video from the test set (dgskorpus_goe_02), in yellow, signing: "Why do you smoke?" through the use of three signs: *WHY* (+mouthed), *TO-SMOKE*, and a gesture (+mouthed) towards the other signer. At the top, our phrase segmentation model predicts a single phrase that initiates with a B tag (in green) above the B-threshold (green dashed line), followed by an I (in light blue), and continues until falling below a certain threshold. At the bottom, our sign segmentation model accurately segments the three signs.

Evaluating on the Public DGS Corpus (Prillwitz et al., 2008; Hanke et al., 2020) (DGS stands for German Sign Language), we report enhancements in model performance following specific modifications. We compare our final models after hyperparameter optimization, including parameters for the decoding algorithm, and find that our best architecture using only the poses is comparable to the one that uses optical flow and hand normalization.

Reassuringly, we find that our model generalizes when evaluated on additional data from different signed languages in a zero-shot approach. We obtain segmentation scores that are competitive with previous work and observe that incorporating optical flow and hand normalization makes the model more robust for out-of-domain data. Our code and models are available at `https://github.com/sign-language-processing/transcription`.

### 6.1.2   Related Work

**Sign Language Detection**

Segmentation consists of detecting the frame boundaries for signs or phrases in videos to divide them into meaningful units. While the most canonical way of dividing a spoken language text is into a linear sequence of words, due to the simultaneity of sign language, the notion of a sign language "word" is ill-defined, and sign language cannot be fully linearly modeled.

Current methods resort to segmenting units loosely mapped to signed language units (Santemiz et al., 2009; Farag and Brock, 2019; Bull et al., 2020b; Renz et al., 2021a,b; Bull et al., 2021) and do not explicitly leverage reliable linguistic predictors of sentence boundaries such as prosody in signed languages (i.e., pauses, extended sign duration, facial expressions) (Sandler, 2010; Ormel and Crasborn, 2012). De Sisto et al. (2021) call for a better understanding of sign language structure, which they believe is the necessary ground for the design and development of sign language recognition and segmentation methodologies.

Santemiz et al. (2009) automatically extracted isolated signs from continuous signing by aligning the sequences obtained via speech recognition, modeled by Dynamic Time Warping (DTW) and Hidden Markov Models (HMMs).

Farag and Brock (2019) used a random forest classifier to distinguish frames containing signs in Japanese Sign Language based on the composition of spatio-temporal angular and distance features between specific pairs of joint segments.

Bull et al. (2020b) segmented French Sign Language into segments corresponding to subtitle units by relying on the alignment between subtitles and sign language videos, leveraging a spatio-temporal graph convolutional network (STGCN; Yu et al. (2017)) with a BiLSTM on 2D skeleton data.

Renz et al. (2021a) located temporal boundaries between signs in continuous sign language videos by employing 3D convolutional neural network representations with iterative temporal segment refinement to resolve ambiguities between sign boundary cues. Renz et al. (2021b) further proposed the Changepoint-Modulated Pseudo-Labelling (CMPL) algorithm to solve the problem of source-free domain adaptation.

Bull et al. (2021) presented a Transformer-based approach to segment sign language videos and align them with subtitles simultaneously, encoding subtitles by BERT (Devlin et al., 2019) and videos by CNN video representations.

### 6.1.3 Motivating Observations

To motivate our proposed approach, we make a series of observations regarding the intrinsic nature of sign language expressions. Specifically, we highlight the unique challenges posed by the continuous flow of sign language expressions, the role of prosody in determining phrase boundaries, and the influence of hand shape changes in indicating sign boundaries.

**Boundary Modeling**

When examining the nature of sign language expressions, we note that the utterances are typically signed in a continuous flow, with minimal to no pauses between individual signs. This continuity is particularly evident when dealing with lower frame rates. This continuous nature presents a significant difference from *text* where specific punctuation marks serve as indicators of phrase boundaries, and a semi-closed set of tokens represent the *words*.

Given these characteristics, directly applying conventional segmentation or sign language detection models—that is, utilizing IO tagging in a manner similar to image or audio segmentation models—may not yield the optimal solution, particularly at the sign level. Such models often fail to precisely identify the boundaries between signs.

A promising alternative is Beginning-Inside-Outside (BIO) tagging (Ramshaw and Marcus, 1995). BIO tagging was originally used for named entity recognition, but its application extends to any sequence chunking task beyond the text modality. In the context of sign language, BIO tagging provides a more refined model for discerning boundaries between signs and phrases, thus significantly improving segmentation performance (Figure 6.1).

To test the viability of the BIO tagging approach in comparison with the IO tagging, we conducted an experiment on the Public DGS Corpus. The entire corpus was transformed to various frame rates and the sign segments were converted to frames using either BIO or IO tagging, then decoded back into sign segments. Figure 6.3 illustrates the results of this comparison. Note that the IO tagging was unable to reproduce the same number of segments as the BIO tagging on the gold data. This underscores the importance of BIO tagging in identifying sign and phrase boundaries.

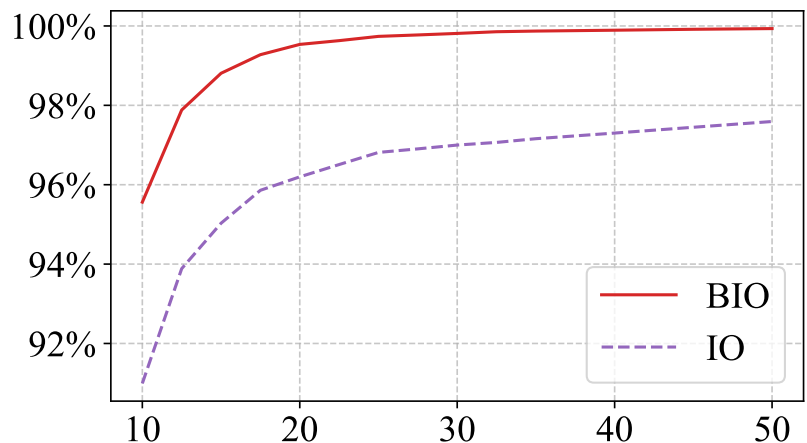

Figure 6.3: Reproduced sign segments in the Public DGS Corpus by BIO and IO tagging at various frame rates. 99.7% of segments reproduced at 25fps by BIO tagging.

**Phrase Boundaries**

Linguistic research has shown that prosody is a reliable predictor of phrase boundaries in signed languages (Sandler, 2010; Ormel and Crasborn, 2012). We observe that this is also the case in the Public DGS Corpus dataset used in our experiments. To illustrate this, we model pauses and movement using optical flow directly on the poses as proposed by Moryossef et al. (2020). Figure 5.1 demonstrates that a change in motion signifies the presence of a pause, which, in turn, indicates a phrase boundary.

**Sign Boundaries**

We observe that signs generally utilize a limited number of hand shapes, with the majority of signs utilizing a maximum of two hand shapes. For example, linguistically annotated datasets, such as ASL-LEX (Sehyr et al., 2021) and ASLLVD (Neidle et al., 2012), only record one initial hand shape per hand and one final hand shape. Mandel (1981, p. 87) argued that there can only be one set of selected fingers per sign, constraining the number of handshapes in signs. This limitation is referred to as the *Selected Fingers Constraint*. And indeed, Sandler et al. (2008) find that the optimal form of a sign is monosyllabic, and that handshape change is organized by the syllable unit.

To illustrate this constraint empirically, we show a histogram of hand shapes per sign in SignBank[2] for $705,151$ signs in Figure 6.4.

Additionally, we found that a change in the dominant hand shape often signals the presence of a sign boundary. Specifically, out of $27,658$ sentences, comprising $354,955$ pairs of consecutive signs, only $17.38\%$ of consecutive signs share the same base hand shape[3]. Based on these observations, we propose using 3D hand normalization as an indicative cue for hand shapes to assist in detecting sign boundaries. We hypothesize that performing 3D hand normalization makes it easier for the model to extract the hand shape.

---

[2] https://signbank.org/signpuddle2.0/

[3] It is important to note that this percentage is inflated, as it may encompass overlaps across the dominant and non-dominant hands, which were not separated for this analysis.

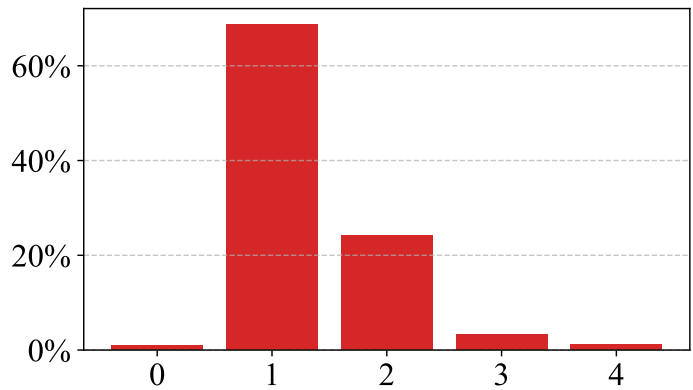

Figure 6.4: Number of hand shapes per sign in SignBank.

### 6.1.4 Experimental Setup

In this section, we describe the experimental setup used to evaluate our linguistically motivated approach for sign language segmentation. This includes a description of the Public DGS Corpus dataset used in the study, the methodology employed to perform sign and phrase segmentation, and the evaluation metrics used to measure the performance of the proposed approach.

**Dataset**

The Public DGS Corpus (Prillwitz et al., 2008; Hanke et al., 2020) is a sign language dataset that includes both accurate sign-level annotation from continuous signing, and well-aligned phrase-level translation in spoken language.

The corpus comprises 404 documents / 714 videos[4] with an average duration of 7.55 minutes, featuring either one signer or two signers, at 50 fps. Most of these videos feature gloss transcriptions and spoken language translations (German and English), except for the ones in the "Joke" category, which are not annotated and thus excluded from our model. We also exclude documents with missing annotations. $id \in \{1289910, 1245887, 1289868, 1246064, 1584617\}$. The translations are comprised of full spoken language paragraphs, sentences, or phrases (i.e., independent/main clauses).

---

[4]The number of videos is nearly double the number of documents because each document typically includes two signers, each of whom produces one video for segmentation.

Each gloss span is considered a gold sign segment, following a tight annotation scheme (Hanke et al., 2012). Phrase segments are identified by examining every translation, with the segment assumed to span from the start of its first sign to the end of its last sign, correcting imprecise annotation.

This corpus is enriched with full-body pose estimations from OpenPose (Cao et al., 2019; Schulder and Hanke, 2019) and Mediapipe Holistic (Grishchenko and Bazarevsky, 2020). We use the *3.0.0-uzh-document* split from Zhang et al. (2023). After filtering the data, we are left with 296 documents / 583 videos for training, 6 / 12 for validation, and 9 / 17 for testing. The mean number of signs and phrases in a video from the training set is 613 and 111, respectively.

### Methodology

Our proposed approach for sign language segmentation is based on the following steps:

1. **Pose Estimation** Given a video, we first adjust it to 25 fps and estimate body poses using the MediaPipe Holistic pose estimation system. We do not use OpenPose because it lacks a $Z$-axis, which prevents 3D rotation used for hand normalization. The shape of a pose is represented as (frames $\times$ keypoints $\times$ axes).

2. **Pose Normalization** To generalize over video resolution and distance from the camera, we normalize each of these poses such that the mean distance between the shoulders of each person equals 1, and the mid-point is at $(0, 0)$ (Celebi et al., 2013). We also remove the legs since they are less relevant to signing.

3. **Optical Flow** We follow Moryossef et al. (2020, Equation 1).

4. **3D Hand Normalization** Following §3.3, we rotate and scale each hand to ensure that the same hand shape is represented in a consistent manner across different frames. We rotate the 21 $XYZ$ keypoints of the hand so that the back of the hand lies on the *XY* plane, we then rotate the hand so that the metacarpal bone of the middle finger lies on the *Y*-axis, and finally, we scale the hand such that the bone is of constant length.

5. **Sequence Encoder** For every frame, the pose is first flattened and projected into a standard dimension (256), then fed through an LSTM encoder (Hochreiter and Schmidhuber, 1997).

6. **BIO Tagging** On top of the encoder, we place two BIO classification heads for sign and phrase independently. *B* denotes the beginning of a sign or phrase, *I* denotes the middle of a sign or phrase, and *O* denotes being outside a sign or phrase. Our cross-entropy loss is proportionally weighted in favor of *B* as it is a *rare* label[5] compared to *I* and *O*.

7. **Greedy Segment Decoding** To decode the frame-level BIO predictions into sign/phrase segments, we define a segment to start with the first frame possessing a *B* probability greater than a predetermined threshold (defaulted at 0.5). The segment concludes with the first frame among the subsequent frames, having either a *B* or *O* probability exceeding the threshold. We provide our exact decoding algorithm in Algorithm 3. We opt to employ adjustable thresholds rather than *argmax* prediction, as our empirical findings demonstrate superior performance (§6.1.5).

---

[5]B:I:O is about 1:5:18 for signs and 1:58:77 for phrases.

---

**Algorithm 3** Probabilities to Segments Conversion.

---

**Require:** *probs*, a list of probabilities from 0 to 100
    $threshold_b \leftarrow 50.0$
    $threshold_o \leftarrow 50.0$

    $start \leftarrow None$
    $did\_pass\_start \leftarrow False$

    **for** $i = 0$ **to** $len(probs)$ **do**
      $b, i, o \leftarrow probs[i]$

      **if** $start = None$ **then**
        **if** $b > threshold_b$ **then**
          $start \leftarrow i$
        **end if**
      **else**
        **if** $did\_pass\_start$ **then**
          **if** $b > threshold_b$ **or** $o > threshold_o$ **then**
            **yield** $(start, i - 1))$
            $start \leftarrow None$
            $did\_pass\_start \leftarrow False$
          **end if**
        **else**
          **if** $b < threshold_b$ **then**
            $did\_pass\_start \leftarrow True$
          **end if**
        **end if**
      **end if**
    **end for**

    **if** $start \neq None$ **then**
      **yield** $(start, len(probs)))$
    **end if**

---

**Experiments**

Our approach is evaluated through a series of six sets of experiments. Each set is repeated three times with varying random seeds. Preliminary experiments were conducted to inform the selection of hyperparameters and features, the details of which can be found in Table 6.2 in §6.1.5. Model selection is based on validation metrics.

1. **E0:  IO Tagger** We re-implemented and reproduced[6] the sign language detection model proposed by Moryossef et al. (2020), in PyTorch (Paszke et al., 2019a) as a naive baseline.  This model processes optical flow as input and outputs *I* (is signing) and *O* (not signing) tags.

2. **E1: Bidirectional BIO Tagger** We replace the IO tagging heads in *E0* with BIO heads to form our baseline.  Our preliminary experiments indicate that inputting only the 75 hand and body keypoints and making the LSTM layer bidirectional yields optimal results.

3. **E2:  Adding Reduced Face Keypoints** Although the 75 hand and body keypoints serve as an efficient minimal set for sign language detection/segmentation models, we investigate the impact of other nonmanual sign language articulators, namely, the face. We introduce a reduced set of 128 face keypoints that signify the signer's *face contour*[7].

4. **E3: Adding Optical Flow** At every time step *t* we append the optical flow between *t* and *t* − 1 to the current pose frame as an additional dimension after the *XYZ* axes.

5. **E4:  Adding 3D Hand Normalization** At every time step, we normalize the hand poses and concatenate them to the current pose.

6. **E5:  Autoregressive Encoder** We replace the encoder with the one proposed by Jiang et al. (2023b) for the detection and classification of great ape calls from raw audio signals. Specifically, we add autoregressive connections between time steps to encourage consistent output labels. The logits at time step $t$ are concatenated to the input of the next time step, $t + 1$. This modification is implemented bidirectionally by stacking two autoregressive encoders and adding their output up before the Softmax operation. This approach is slow, as we have to wait for the previous time step predictions before we can feed them to the next time step.

---

[6]The initial implementation uses OpenPose (Cao et al., 2019), at 50 fps. Preliminary experiments reveal that these differences do not significantly influence the results.

[7]We reduce the dense *FACE_LANDMARKS* in Mediapipe Holistic to the contour keypoints according to the variable *mediapipe.solutions.holistic.FACEMESH_CONTOURS*.

**Evaluation Metrics**

We evaluate the performance of our proposed approach for sign and phrase segmentation using the following metrics:

- **Frame-level F1 Score** For each frame, we apply the *argmax* operation to make a local prediction of the BIO class and calculate the macro-averaged per-class F1 score against the ground truth label. We use this frame-level metric during validation as the primary metric for model selection and early stopping, due to its independence from a potentially variable segment decoding algorithm.

- **Intersection over Union (IoU)** We compute the IoU between the ground truth segments and the predicted segments to measure the degree of overlap. Note that we do not perform a one-to-one mapping between the two using techniques like DTW. Instead, we calculate the total IoU based on all segments in a video.

- **Percentage of Segments (%)** To complement IoU, we introduce the percentage of segments to compare the number of segments predicted by the model with the ground truth annotations. It is computed as follows: $\frac{\#\text{predicted segments}}{\#\text{ground truth segments}}$. The optimal value is 1.

- **Efficiency** We measure the efficiency of each model by the number of parameters and the training time of the model on a Tesla V100-SXM2-32GB GPU for 100 epochs[8].

---

[8]Exceptionally the autoregressive models in *E5* were trained on an NVIDIA A100-SXM4-

### 6.1.5 Results and Discussion

We report the mean test evaluation metrics for our experiments in Table 6.1.

| | | Sign | | | Phrase | | | Efficiency | |
|---|---|---|---|---|---|---|---|---|---|
| **Experiment** | | **F1** | **IoU** | **%** | **F1** | **IoU** | **%** | **#Params** | **Time** |
| **E0** | **Moryossef et al. (2020)** | — | 0.46 | 1.09 | — | 0.70 | **1.00** | **102K** | **0:50:17** |
| **E1** | **Baseline** | 0.56 | 0.66 | 0.91 | 0.59 | 0.80 | 2.50 | 454K | 1:01:50 |
| **E2** | **E1 + Face** | 0.53 | 0.58 | 0.64 | 0.57 | 0.76 | 1.87 | 552K | 1:50:31 |
| **E3** | **E1 + Optical Flow** | 0.58 | 0.62 | 1.12 | 0.60 | 0.82 | 3.19 | 473K | 1:20:17 |
| **E4** | **E3 + Hand Norm** | 0.56 | 0.61 | 1.07 | 0.60 | 0.80 | 3.24 | 516K | 1:30:59 |
| **E1s** | **E1 + Depth=4** | **0.63** | **0.69** | 1.11 | **0.65** | 0.82 | 1.63 | 1.6M | 4:08:48 |
| **E2s** | **E2 + Depth=4** | 0.62 | **0.69** | 1.07 | 0.63 | 0.84 | 2.68 | 1.7M | 3:14:03 |
| **E3s** | **E3 + Depth=4** | 0.60 | 0.63 | 1.13 | 0.64 | 0.80 | 1.53 | 1.7M | 4:08:30 |
| **E4s** | **E4 + Depth=4** | 0.59 | 0.63 | 1.13 | 0.62 | 0.79 | 1.43 | 1.7M | 4:35:29 |
| **E1s*** | **E1s + Tuned Decoding** | — | **0.69** | **1.03** | — | **0.85** | 1.02 | — | — |
| **E4s*** | **E4s + Tuned Decoding** | — | 0.63 | 1.06 | — | 0.79 | 1.12 | — | — |
| **E5** | **E4s + Autoregressive** | 0.45 | 0.47 | 0.88 | 0.52 | 0.63 | 2.72 | 1.3M | ~3 days |

Table 6.1: Mean test evaluation metrics for our experiments. The best score of each column is in bold and a star (*) denotes further optimization of the decoding algorithm without changing the model (only affects *IoU* and %). Table 6.3 contains a complete report including validation metrics and standard deviation of all experiments.

We do not report F1 Score for *E0* since it has a different number of classes and is thus incomparable. Comparing *E1* to *E0*, we note that the model's bidirectionality, the use of poses, and BIO tagging indeed help outperform the model from previous work where only optical flow and IO tagging are used. While *E1* predicts an excessive number of phrase segments, the IoUs for signs and phrases are both higher.

Adding face keypoints (*E2*) makes the model worse, while including optical flow (*E3*) improves the F1 scores. For phrase segmentation, including optical flow increases IoU, but over-segments phrases by more than 300%, which further exaggerates the issue in *E1*. Including hand normalization (*E4*) on top of *E3* slightly deteriorates the quality of both sign and phrase segmentation.

---

80GB GPUA100 which doubles the training speed of V100, still the training is slow.

From the non-exhaustive hyperparameter search in the preliminary experiments (Table 6.2), we examined different hidden state sizes (64, 128, 256, 512, 1024) and a range of 1 to 8 LSTM layers, and conclude that a hidden size of 256 and 4 layers with $1e-3$ learning rate are optimal for *E1*, which lead to *E1s*. We repeat the setup of *E2*, *E3*, and *E4* with these refined hyper-parameters, and show that all of them outperform their counterparts, notably in that they ease the phrase over-segmentation problem.

In *E2s*, we reaffirmed that adding face keypoints does not yield beneficial results, so we exclude face in future experiments. Although the face is an essential component to understanding sign language expressions and does play some role in sign and phrase level segmentation, we believe that the 128 face contour points are too dense for the model to learn useful information compared to the 75 body points, and may instead confuse the model.

In addition, the benefits of explicitly including optical flow (*E3s*) fade away with the increased model depth and we speculate that now the model might be able to learn the optical flow features by itself. Surprisingly, while adding hand normalization (*E4s*) still slightly worsens the overall results, it has the best phrase percentage.

From *E4s* we proceeded with the training of *E5*, an autoregressive model. Unexpectedly, counter to our intuition and previous work, *E5* underachieves our baseline across all evaluation metrics[9].

**Extended Experimental Results**

We conducted some preliminary experiments (starting with *P0*) on training a sign language segmentation model to gain insights into hyperparameters and feature choices. The results are shown in Table 6.2[10]. We found in *P1.3.2* the optimal hyperparameters and repeated them with different feature choices.

---

[9]*E5* should have more parameters than *E4s*, but because of an implementation bug, each LSTM layer has half the parameters. Based on the current results, we assume that autoregressive connections (even with more parameters) will not improve our models.

[10]Note that due to an implementation issue on edge cases (which we fixed later), the *IoU* and % values in Table 6.2 are lower than the ones in Table 6.1 and Table 6.3 thus not comparable across tables. The comparison inside of Table 6.2 between different experiments remains meaningful. In addition, the results in Table 6.2 are based on only one run instead of three.

| Experiment | | | Sign | | | Phrase | | |
|---|---|---|---|---|---|---|---|---|
| | | | **F1** | **IoU** | **%** | **F1** | **IoU** | **%** |
| **P0** | **Moryossef et al. (2020)** | test | — | 0.4 | 1.45 | — | 0.65 | 0.82 |
| | | dev | — | 0.35 | 1.36 | — | 0.6 | 0.77 |
| **P0.1** | **P0 + Holistic 25fps** | test | — | 0.39 | 0.86 | — | 0.64 | 0.5 |
| | | dev | — | 0.32 | 0.81 | — | 0.58 | 0.52 |
| **P1** | **P1 baseline** | test | 0.55 | 0.49 | 0.83 | 0.6 | 0.67 | 2.63 |
| | | dev | 0.56 | 0.43 | 0.75 | 0.58 | 0.62 | 2.61 |
| **P1.1** | **P1 - encoder_bidirectional** | test | 0.48 | 0.45 | 0.68 | 0.5 | 0.64 | 2.68 |
| | | dev | 0.46 | 0.41 | 0.64 | 0.51 | 0.61 | 2.56 |
| **P1.2.1** | **P1 + hidden_dim=512** | test | 0.47 | 0.42 | 0.44 | 0.52 | 0.63 | 1.7 |
| | | dev | 0.46 | 0.4 | 0.43 | 0.52 | 0.61 | 1.69 |
| **P1.2.2** | **P1 + hidden_dim=1024** | test | 0.48 | 0.45 | 0.42 | 0.58 | 0.65 | 1.53 |
| | | dev | 0.46 | 0.41 | 0.36 | 0.53 | 0.61 | 1.49 |
| **P1.3.1** | **P1 + encoder_depth=2** | test | 0.55 | 0.48 | 0.76 | 0.58 | 0.67 | 2.56 |
| | | dev | 0.56 | 0.43 | 0.69 | 0.58 | 0.62 | 2.52 |
| **P1.3.2** | **P1 + encoder_depth=4** | test | 0.63 | 0.51 | 0.91 | 0.66 | 0.67 | 1.41 |
| | | dev | 0.61 | 0.47 | 0.84 | 0.64 | 0.6 | 1.39 |
| **P1.4.1** | **P1 + hidden_dim=128 + encoder_depth=2** | test | 0.58 | 0.48 | 0.8 | 0.6 | 0.67 | 2.0 |
| | | dev | 0.55 | 0.43 | 0.75 | 0.54 | 0.62 | 2.03 |
| **P1.4.2** | **P1 + hidden_dim=128 + encoder_depth=4** | test | 0.62 | 0.51 | 0.91 | 0.64 | 0.68 | 2.43 |
| | | dev | 0.6 | 0.47 | 0.83 | 0.6 | 0.62 | 2.57 |
| **P1.4.3** | **P1 + hidden_dim=128 + encoder_depth=8** | test | 0.59 | 0.52 | 0.91 | 0.63 | 0.68 | 3.04 |
| | | dev | 0.6 | 0.47 | 0.84 | 0.6 | 0.62 | 3.02 |
| **P1.5.1** | **P1 + hidden_dim=64 + encoder_depth=4** | test | 0.57 | 0.5 | 0.8 | 0.6 | 0.68 | 2.41 |
| | | dev | 0.58 | 0.45 | 0.75 | 0.59 | 0.62 | 2.39 |
| **P1.5.2** | **P1 + hidden_dim=64 + encoder_depth=8** | test | 0.62 | 0.51 | 0.85 | 0.64 | 0.68 | 2.53 |
| | | dev | 0.6 | 0.46 | 0.79 | 0.6 | 0.62 | 2.53 |
| **P2** | **P1 + optical_flow** | test | 0.58 | 0.5 | 0.95 | 0.63 | 0.68 | 3.17 |
| | | dev | 0.59 | 0.45 | 0.84 | 0.59 | 0.61 | 3.08 |
| **P2.1** | **P1.3.2 + optical_flow** | test | 0.63 | 0.51 | 0.92 | 0.66 | 0.67 | 1.51 |
| | | dev | 0.62 | 0.46 | 0.81 | 0.62 | 0.6 | 1.53 |
| **P3** | **P1 + hand_normalization** | test | 0.55 | 0.48 | 0.77 | 0.58 | 0.67 | 2.79 |
| | | dev | 0.55 | 0.42 | 0.71 | 0.57 | 0.62 | 2.73 |
| **P3.1** | **P1.3.2 + hand_normalization** | test | 0.63 | 0.51 | 0.91 | 0.66 | 0.66 | 1.43 |
| | | dev | 0.61 | 0.46 | 0.82 | 0.64 | 0.61 | 1.46 |
| **P4** | **P2.1 + P3.1** | test | 0.56 | 0.51 | 0.92 | 0.61 | 0.66 | 1.45 |
| | | dev | 0.61 | 0.46 | 0.81 | 0.63 | 0.6 | 1.41 |
| **P4.1** | **P4 + encoder_depth=8** | test | 0.6 | 0.51 | 0.95 | 0.62 | 0.67 | 1.08 |
| | | dev | 0.61 | 0.47 | 0.86 | 0.62 | 0.6 | 1.12 |
| **P5** | **P1.3.2 + reduced_face** | test | 0.63 | 0.51 | 0.94 | 0.64 | 0.66 | 1.16 |
| | | dev | 0.61 | 0.47 | 0.86 | 0.64 | 0.58 | 1.14 |
| **P5.1** | **P1.3.2 + full_face** | test | 0.54 | 0.49 | 0.8 | 0.6 | 0.68 | 2.29 |
| | | dev | 0.57 | 0.45 | 0.7 | 0.59 | 0.62 | 2.29 |

Table 6.2: Results of the preliminary experiments.

We selected some promising models from our preliminary experiments and reran them three times using different random seeds to make the final conclusion reliable and robust. Table 6.3 includes the standard deviation and the validation results for readers to scrutinize.

| Experiment | | | Sign | | | Phrase | | | Efficiency | |
|---|---|---|---|---|---|---|---|---|---|---|
| | | | F1 | IoU | % | F1 | IoU | % | #Params | Time |
| E0 | Moryossef et al. (2020) | test | — | 0.46 ± 0.03 | 1.09 ± 0.41 | — | 0.70 ± 0.01 | 1.00 ± 0.06 | 102K | 0:50:17 |
| | | dev | — | 0.42 ± 0.05 | 1.21 ± 0.59 | — | 0.61 ± 0.06 | 2.47 ± 0.85 | 102K | 0:50:17 |
| E1 | Baseline | test | 0.56 ± 0.03 | 0.66 ± 0.01 | 0.91 ± 0.05 | 0.59 ± 0.02 | 0.80 ± 0.03 | 2.50 ± 0.13 | 454K | 1:01:50 |
| | | dev | 0.55 ± 0.01 | 0.59 ± 0.00 | 1.12 ± 0.11 | 0.56 ± 0.02 | 0.75 ± 0.05 | 2.94 ± 0.08 | 454K | 1:01:50 |
| E2 | E1 + Face | test | 0.53 ± 0.05 | 0.58 ± 0.07 | 0.64 ± 0.30 | 0.57 ± 0.02 | 0.76 ± 0.03 | 1.87 ± 0.83 | 552K | 1:50:31 |
| | | dev | 0.50 ± 0.07 | 0.53 ± 0.11 | 0.90 ± 0.19 | 0.53 ± 0.05 | 0.71 ± 0.07 | 2.43 ± 1.02 | 552K | 1:50:31 |
| E3 | E1 + Optical Flow | test | 0.58 ± 0.01 | 0.62 ± 0.00 | 1.12 ± 0.05 | 0.60 ± 0.03 | 0.82 ± 0.03 | 3.19 ± 0.11 | 473K | 1:20:17 |
| | | dev | 0.58 ± 0.00 | 0.62 ± 0.00 | 1.50 ± 0.19 | 0.59 ± 0.01 | 0.79 ± 0.00 | 3.94 ± 0.14 | 473K | 1:20:17 |
| E4 | E3 + Hand Norm | test | 0.56 ± 0.02 | 0.61 ± 0.00 | 1.07 ± 0.05 | 0.60 ± 0.00 | 0.80 ± 0.00 | 3.24 ± 0.17 | 516K | 1:30:59 |
| | | dev | 0.57 ± 0.01 | 0.61 ± 0.01 | 1.50 ± 0.07 | 0.58 ± 0.00 | 0.79 ± 0.00 | 4.04 ± 0.31 | 516K | 1:30:59 |
| E1s | E1 + Depth=4 | test | 0.63 ± 0.01 | 0.69 ± 0.00 | 1.11 ± 0.01 | 0.65 ± 0.02 | 0.82 ± 0.04 | 1.63 ± 0.10 | 1.6M | 4:08:48 |
| | | dev | 0.61 ± 0.00 | 0.63 ± 0.00 | 1.27 ± 0.01 | 0.63 ± 0.01 | 0.77 ± 0.01 | 2.17 ± 0.18 | 1.6M | 4:08:48 |
| E2s | E2 + Depth=4 | test | 0.62 ± 0.02 | 0.69 ± 0.00 | 1.07 ± 0.03 | 0.63 ± 0.01 | 0.84 ± 0.03 | 2.68 ± 0.53 | 1.7M | 3:14:03 |
| | | dev | 0.60 ± 0.01 | 0.63 ± 0.01 | 1.20 ± 0.12 | 0.59 ± 0.02 | 0.76 ± 0.05 | 3.30 ± 0.62 | 1.7M | 3:14:03 |
| E3s | E3 + Depth=4 | test | 0.60 ± 0.01 | 0.63 ± 0.00 | 1.13 ± 0.01 | 0.64 ± 0.03 | 0.80 ± 0.03 | 1.53 ± 0.18 | 1.7M | 4:08:30 |
| | | dev | 0.62 ± 0.00 | 0.63 ± 0.00 | 1.63 ± 0.05 | 0.63 ± 0.00 | 0.76 ± 0.00 | 2.14 ± 0.09 | 1.7M | 4:08:30 |
| E4s | E4 + Depth=4 | test | 0.59 ± 0.00 | 0.63 ± 0.00 | 1.13 ± 0.03 | 0.62 ± 0.00 | 0.79 ± 0.00 | 1.43 ± 0.10 | 1.7M | 4:35:29 |
| | | dev | 0.61 ± 0.00 | 0.63 ± 0.00 | 1.56 ± 0.04 | 0.63 ± 0.00 | 0.77 ± 0.01 | 1.89 ± 0.07 | 1.7M | 4:35:29 |
| E4ba | E4s + Autoregressive | test | 0.45 ± 0.03 | 0.47 ± 0.05 | 0.88 ± 0.08 | 0.52 ± 0.02 | 0.63 ± 0.10 | 2.72 ± 1.33 | 1.3M | 2 days, 21:28:42 |
| | | dev | 0.40 ± 0.01 | 0.40 ± 0.01 | 2.02 ± 0.73 | 0.47 ± 0.00 | 0.57 ± 0.04 | 4.26 ± 1.26 | 1.3M | 2 days, 21:28:42 |

Table 6.3: Mean evaluation metrics for our main experiments. A complete version of Table 6.1.

### Challenges with 3D Hand Normalization

While the use of 3D hand normalization is well-justified in §6.1.3, we believe it does not help the model due to poor depth estimation quality, as further corroborated by recent research from De Coster et al. (2023). Therefore, we consider it a negative result, showing the deficiencies in the 3D pose estimation system. The evaluation metrics we propose in §3.3 could help identify better pose estimation models for this use case.

### Tuning the Segment Decoding Algorithm

We selected *E1s* and *E4s* to further explore the segment decoding algorithm. As detailed in §6.1.4, the decoding algorithm has two tunable parameters, $threshold_b$ and $threshold_o$. We conducted a grid search with these parameters, using values from 10 to 90 in increments of 10. We additionally experimented with a variation of the algorithm that conditions on the most likely class by *argmax* instead of fixed threshold values, which turned out similar to the default version.

We only measured the results using IoU and the percentage of segments at validation time since the F1 scores remain consistent in this case. For sign segmentation, we found using $threshold_b = 60$ and $threshold_o = 40/50/60$ yields slightly better results than the default setting (50 for both). For phrase segmentation, we identified that higher threshold values ($threshold_b = 90$, $threshold_o = 90$ for *E1s* and $threshold_b = 80$, $threshold_o = 80/90$ for *E4s*) improve on the default significantly, especially on the percentage metric. We report the test results under *E1s\** and *E4s\**, respectively.

Despite formulating a single model, we underline a separate sign/phrase model selection process to archive the best segmentation results. Figure 6.5 illustrates how higher threshold values reduce the number of predicted segments and skew the distribution of predicted phrase segments towards longer ones in *E1s/E1s\**. As Bull et al. (2020b) suggest, advanced priors could also be integrated into the decoding algorithm.

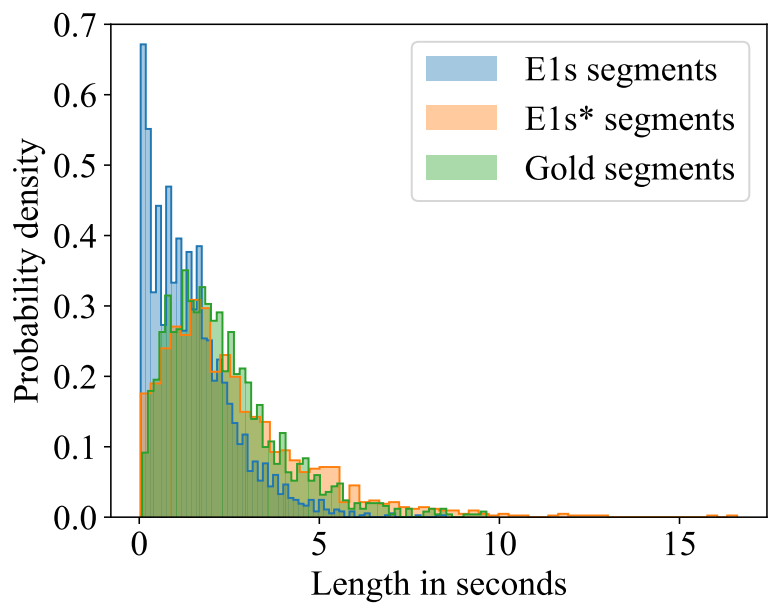

Figure 6.5: Probability density of phrase segment lengths.

**Comparison to Previous Work**

We re-implemented and re-purposed the sign language detection model introduced in Moryossef et al. (2020) for our segmentation task as a baseline since their work is the state-of-the-art and the only comparable model designed for the Public DGS Corpus dataset. As a result, we show the need of replacing IO tagging with BIO tagging to tackle the subtle differences between the two tasks.

For *phrase* segmentation, we compare to Bull et al. (2020b). We note that our definition of sign language phrases (spanning from the start of its first sign to the end of its last sign) is tighter than the subtitle units used in their paper and that we use different training datasets of different languages and domains. Nevertheless, we implemented some of their frame-level metrics and show the results in Table 6.4 on both the Public DGS Corpus and the MEDIAPI-SKEL dataset (Bull et al., 2020a) in French Sign Language (LSF). We report both zero-shot out-of-domain results[11] and the results of our models trained specifically on their dataset without the spatio-temporal graph convolutional network (ST-GCN; Yan et al. (2018)) used in their work for pose encoding.

For *sign* segmentation, we do not compare to Renz et al. (2021a,b) due to different datasets and the difficulty in reproducing their segment-level evaluation metrics. The latter depends on the decoding algorithm and a way to match the gold and predicted segments, both of which are variable.

## 6.1.6 Conclusions

This work focuses on the automatic segmentation of signed languages. We are the first to formulate the segmentation of individual signs and larger sign phrases as a joint problem.

We propose a series of improvements over previous work, linguistically motivated by careful analyses of sign language corpora. Recognizing that sign language utterances are typically continuous with minimal pauses, we opted for a BIO tagging scheme over IO tagging. Furthermore, leveraging the fact that phrase boundaries are marked by prosodic cues, we introduce optical flow features as a proxy for prosodic processes. Finally, since signs typically employ a limited number of hand shapes, to make it easier for the model to understand handshapes, we attempt 3D hand normalization.

---

[11]The zero-shot results are not directly comparable to theirs due to different datasets and labeling approaches.

| Data | Model | ROC-AUC | F1-M |
|------|-------|---------|------|
| LSF | **full (theirs)** | 0.87 | — |
| | **body (theirs)** | 0.87 | — |
| | **E1s (ours, zero-shot)** | 0.71 | 0.41 |
| | **E4s (ours, zero-shot)** | 0.76 | 0.44 |
| | **E1s (ours, trained)** | 0.87 | 0.49 |
| | **E4s (ours, trained)** | 0.87 | 0.51 |
| DGS | **E1s (ours)** | 0.91 | 0.65 |
| | **E4s (ours)** | 0.90 | 0.62 |

Table 6.4: Evaluation metrics used in Bull et al. (2020b). *ROC-AUC* is applied exclusively on the *O*-tag. For comparison *F1-M* denotes the macro-averaged per-class F1 used in this work across all tags. The first two rows are the best results taken from Table 1 in their paper. The next four rows represent how our models perform on their data in a zero-shot setting, and in a supervised setting, and the last two rows represent how our models perform on our data.

Our experiments conducted on the Public DGS Corpus confirmed the efficacy of these modifications for segmentation quality. By comparing to previous work in a zero-shot setting, we demonstrate that our models generalize across signed languages and domains and that including linguistically motivated cues leads to a more robust model in this context.

Finally, we envision that the proposed model has applications in real-world data collection for signed languages. Furthermore, a similar segmentation approach could be leveraged in various other fields such as co-speech gesture recognition (Moryossef, 2023a) and action segmentation (Tang et al., 2019).

# Limitations

### Pose Estimation

In this work, we employ the MediaPipe Holistic pose estimation system (Grishchenko and Bazarevsky, 2020). There is a possibility that this system exhibits bias towards certain protected classes (such as gender or race), underperforming in instances with specific skin tones or lower video quality. Thus, we cannot attest to how our system would perform under real-world conditions, given that the videos utilized in our research are generated in a controlled studio environment, primarily featuring white participants.

### Encoding of Long Sequences

In this study, we encode sequences of frames that are significantly longer than the typical 512 frames often seen in models employing Transformers (Vaswani et al., 2017). Numerous techniques, ranging from basic temporal pooling/downsampling to more advanced methods such as a video/pose encoder that aggregates local frames into higher-level 'tokens' (Renz et al., 2021a), graph convolutional networks (Bull et al., 2020b), and self-supervised representations (Baevski et al., 2020), can alleviate length constraints, facilitate the use of Transformers, and potentially improve the outcomes. Moreover, a hierarchical method like the Swin Transformer (Liu et al., 2021) could be applicable.

### Limitations of Autoregressive LSTMs

In this paper, we replicated the autoregressive LSTM implementation originally proposed by Jiang et al. (2023b). Our experiments revealed that this implementation exhibits significant slowness, which prevented us from performing further experimentation. In contrast, other LSTM implementations employed in this project have undergone extensive optimization (Appleyard, 2016), including techniques like combining general matrix multiplication operations (GEMMs), parallelizing independent operations, fusing kernels, rearranging matrices, and implementing various optimizations for models with multiple layers (which are not necessarily applicable here).

A comparison of CPU-based performance demonstrates that our implementation is x6.4 times slower.  Theoretically, the number of operations performed by the autoregressive LSTM is equivalent to that of a regular LSTM.  However, while the normal LSTM benefits from concurrency based on the number of layers, we do not have that luxury.  The optimization of recurrent neural networks (RNNs) (Que et al., 2020, 2021, 2022) remains an ongoing area of research.  If proven effective in other domains, we strongly advocate for efforts to optimize the performance of this type of network.

**Interference Between Sign and Phrase Models**

In our model, we share the encoder for both the sign and phrase segmentation models, with a shallow linear layer for the BIO tag prediction associated with each task.  It remains uncertain whether these two tasks interfere with or enhance each other.  An ablation study (not presented in this work) involving separate modeling is necessary to obtain greater insight into this matter.

**Noisy Training Objective**

Although the annotations utilized in this study are of expert level, the determination of precise sign (Hanke et al., 2012) and phrase boundaries remains a challenging task, even for experts.  Training the model on these annotated boundaries might introduce excessive noise.  A similar issue was observed in classification-based pose estimation (Cao et al., 2019).  The task of annotating the exact anatomical centers of joints proves to be nearly impossible, leading to a high degree of noise when predicting joint position as a 1-hot classification task. The solution proposed in this previous work was to distribute a Gaussian around the annotated location of each joint.  This approach allows the joint's center to overlap with some probability mass, thereby reducing the noise for the model.  A similar solution could be applied in our context.  Instead of predicting a strict 0 or 1 class probability, we could distribute a Gaussian around the boundary.

### Naive Segment Decoding

We recognize that the frame-level greedy decoding strategy implemented in our study may not be optimal. Previous research in audio segmentation (Venkatesh et al., 2022) employed a You Only Look Once (YOLO; Redmon et al. (2015)) decoding scheme to predict segment boundaries and classes. We propose using a similar prediction atop a given representation, such as the LSTM output or classification logits of an already trained network. Differing from traditional object detection tasks, this process is simplified due to the absence of a $Y$ axis and non-overlapping segments. In this scenario, the network predicts the segment boundaries using regression, thereby avoiding the class imbalance issue of the BIO tagging. We anticipate this to yield more accurate sign language segmentation.

### Lack of Transcription

Speech segmentation is a close task to our sign language segmentation task on videos. In addition to relying on prosodic cues from audio, the former could benefit from automatic speech transcription systems, either in terms of surrogating the task to text-level segmentation and punctuation (Cho et al., 2015), or gaining additional training data from automatic speech recognition / spoken language translation (Tsiamas et al., 2022).

However, for signed languages, there is neither a standardized and widely used written form nor a reliable transcription procedure into some potential writing systems like SignWriting (Sutton, 1990), HamNoSys (Prillwitz and Zienert, 1990), and glosses (Johnston, 2008). Transcription/recognition and segmentation tasks need to be solved simultaneously, so we envision that a multi-task setting helps. Sign spotting, the localization of a specific sign in continuous signing, is a simplification of the segmentation and recognition problem in a closed-vocabulary setting (Wong et al., 2022; Varol et al., 2022). It can be used to find candidate boundaries for some signs, but not all.

## 6.2 Transcription

Sign Language Transcription involves converting visual sign language, either segmented into single sign units or presented as continuous signing, into a corresponding phonetic form. It serves as an essential intermediary step, bridging the gap between skeletal poses and their translation into spoken languages.

### 6.2.1 Preface

Ideally, this thesis would include a model for automatic transcription, thus completing the envisioned translation pipeline. However, due to the already comprehensive nature of this research and the practical constraints, this component remains a topic for future exploration. This section aims to provide some background and conceptual understanding of the transcription task, setting the stage for future research endeavors in this domain.

### 6.2.2 Introduction

Unlike spoken languages, signed languages have traditionally lacked a standardized written form, presenting significant challenges in transcription and documentation. This has far-reaching implications for accessibility, education, research, and broader communication within and beyond the deaf community. While SignWriting (Sutton, 1990) provides a comprehensive transcription method by capturing the complex movements, facial expressions, and body positions that characterize signed languages, its manual transcription process can be time-consuming and requires expertise.

To address these issues, we urge an automatic transcription system for signed languages using the SignWriting notation. The development of sign-level transcription will offer numerous potential applications and opportunities for future research. Once established, this transcription method can be combined with sign language segmentation models to transcribe full videos, sign by sign. Moreover, a sign language transcription model based on SignWriting could po-tentially be employed in the context of our pipeline to translate these transcriptions into spoken language text (Jiang et al., 2023a; Moryossef and Jiang, 2023). This opens up exciting possibilities for further integration of signed languages into the digital realm, particularly in areas such as automatic subtitling and real-time translation. Furthermore, this approach can enrich existing parallel resources with aligned signed and spoken language data, potentially enhancing the performance of machine translation systems.

### 6.2.3   Background

**Writing Signed Languages**

Written notation systems represent signs as discrete visual features. Some systems are written linearly, and others use graphemes in two dimensions. While various universal (Sutton, 1990; Prillwitz and Zienert, 1990) and language-specific notation systems (Stokoe Jr, 1960; Kakumasu, 1968; Bergman, 1977) have been proposed, no writing system has been adopted widely by any sign language community, and the lack of standards hinders the exchange and unification of resources and applications between projects.

**Semantic Transcription**

This is the first automatic transcription model of signed languages into a phonetic transcription system. To put our work in context, we include related work on semantic transcription.

Pose-to-Gloss, also known as sign language recognition, is the task of recognizing a sequence of signs from a sequence of poses. Though some previous works have referred to this as "sign language translation," recognition merely determines the associated label of each sign, without handling the syntax and morphology of the signed language (Padden, 1988) to create a spoken language output. Instead, SLR has often been used as an intermediate step during translation to produce glosses from signed language videos.

Jiang et al. (2021) proposed a novel Skeleton Aware Multi-modal Framework with a Global Ensemble Model (GEM) for isolated SLR (SAM-SLR-v2) to learn and fuse multimodal feature representations. Specifically, they use a Sign Language Graph Convolution Network (SL-GCN) to model the embedded dynamics of skeleton keypoints and a Separable Spatial-Temporal Convolution Network (SSTCN) to exploit skeleton features. The proposed late-fusion GEM fuses the skeleton-based predictions with other RGB and depth-based modalities to provide global information and make an accurate SLR prediction.

Dafnis et al. (2022) work on the same modified WLASL dataset as Jiang et al. (2021), but do not require multimodal data input. Instead, they propose a bidirectional skeleton-based graph convolutional network framework with linguistically motivated parameters and attention to the start and end frames of signs. They cooperatively use forward and backward data streams, including various sub-streams, as input. They also use pre-training to leverage transfer learning.

Selvaraj et al. (2022) introduced an open-source OpenHands library, which consists of standardized pose datasets for different existing sign language datasets and trained checkpoints of four pose-based isolated sign language recognition models across six languages (American, Argentinian, Chinese, Greek, Indian, and Turkish). To address the lack of labeled data, they propose self-supervised pretraining on unlabeled data and curate the largest pose-based pretraining dataset on Indian Sign Language (Indian-SL). They established that pretraining is effective for sign language recognition by demonstrating improved finetuning performance especially in low-resource settings and high crosslingual transfer from Indian-SL to a few other sign languages.

The work of Kezar et al. (2023), based on the OpenHands library, explicitly recognizes the role of phonology to achieve more accurate isolated sign language recognition (ISLR). To allow additional predictions on phonological characteristics (such as handshape), they combine the phonological annotations in ASL-LEX 2.0 (Sehyr et al., 2021) with signs in the WLASL 2000 ISLR benchmark (Li et al., 2020). Interestingly, Tavella et al. (2022) construct a similar dataset aiming just for phonological property recognition in American Sign Language.

### 6.2.4   Datasets

For this study, there are two notable lexicons, containing isolated sign language videos with SignWriting transcriptions, which can be used to train automatic transcription systems.

**Sign2MINT (Barth et al., 2021)**   is a lexicon of German Signed Language (DGS) focusing on natural science subjects. It features $5,263$ videos with SignWriting transcriptions.

**SignSuisse (Schweizerischer Gehörlosenbund SGB-FSS, 2023)**   is a Swiss Signed Languages Lexicon that covers Swiss-German Sign Language (DSGS), French Sign Language (LSF), and Italian Sign Language (LIS). The lexicon includes approximately $4,500$ LSF videos with SignWriting transcriptions in SignBank[12].

### 6.2.5   Outlook

Automatic sign language transcription would not only contribute to more robust machine translation but also pave the way for more seamless integration of sign languages into various digital platforms and services, and allow for the anonymized distribution of sign language data.

Chapter 9 expands on how automatic sign language transcription can be used in spoken language processing pipelines, including the critical but often overlooked non-verbal cues such as co-speech gestures and facial expressions.

---

[12]https://www.signbank.org/signpuddle2.0/index.php?ui=4&sgn=49

## 6.3   Translation (Moryossef and Jiang, 2023)

During this thesis, I had the privilege of supervising the research undertaken by Jiang et al. (2023a). Our work established a foundational framework for translating between spoken languages rendered in textual format and signed languages depicted through SignWriting. The ensuing section delves deeper into the initiatives undertaken to improve upon our work, refine data quality, and develop a model tailored for client-side deployment.

In this thesis, we introduce SignBank+, a clean version of the SignBank dataset, optimized for machine translation. Contrary to previous work that employs complex factorization techniques for translation, we advocate for a traditional text-to-text translation approach. Our naive evaluation shows that models trained on SignBank+ surpass those on the original dataset, establishing a new benchmark and providing an open resource for future research.

### 6.3.1   Introduction

Sign Language serves as an indispensable mode of communication for the deaf. Unfortunately, the available methods for translating between signed and spoken languages, have been limited in scope and effectiveness. The main objective of this research is to explore technological advancements that can enhance the translation process, focusing on the cleaning and enrichment of an existing sign language dataset, *SignBank*[13], a multilingual collection of *puddles*, covering a range of domains.

The pioneering work of Jiang et al. (2023a) set the stage for this task. They presented an approach to translating SignWriting through specialized parsing and factorized machine translation techniques. Motivated by their efforts, this research aims to build upon their foundation by:

1. Undertaking a rigorous data cleaning and expansion process.

2. Reverting to a simple translation mechanism, omitting any factorization.

---

[13] https://www.signbank.org/signpuddle/

The hypothesis driving this study is twofold: First, a meticulously curated dataset will enhance the accuracy and reliability of translation models. Second, by simplifying the translation process, it becomes feasible to train a diverse array of models and streamline their deployment.

To validate our claims, we compare the translation quality of signed-to-spoken translation using the original and cleaned data. We show that with our new, cleaner data, we can train standard machine translation models with improved quality over the original data. We share our data openly under CC-BY-4.0 (available at https://github.com/sign-language-processing/signbank-plus) to be used in future machine translation research.

### 6.3.2 Background

This work only concerns machine translation between signed and spoken languages where both the input and the output are represented as text.

**Signed-to-Spoken**

Jiang et al. (2023a) explore text-to-text sign to spoken language translation, with SignWriting as the chosen sign language notation system. Despite SignWriting usually represented in 2D, they use the 1D Formal SignWriting specification and propose a neural factored machine translation approach to encode sequences of the SignWriting graphemes as well as their position in the 2D space. They verify the proposed approach on the SignBank dataset in both a bilingual setup (American Sign Language to English) and two multilingual setups (4 and 21 signed-to-spoken language pairs, respectively). They apply several low-resource machine translation techniques used to improve spoken language translation to similarly improve the performance of sign language translation. Their findings validate the use of an intermediate text representation for signed language translation, and pave the way for including sign language translation in natural language processing research.

**Spoken-to-Signed**

Jiang et al. (2023a) also explore the reverse translation direction, i.e., text to Sign-Writing translation. They conduct experiments under a same condition of their multilingual SignWriting to text (4 language pairs) experiment, and again propose a neural factored machine translation approach to decode the graphemes and their position separately. They borrow BLEU from spoken language translation to evaluate the predicted graphemes and mean absolute error to evaluate the positional numbers.

Walsh et al. (2022) explore Text to HamNoSys (T2H) translation, with Ham-NoSys as the target sign language notation system. They experiment with direct T2H and Text to Gloss to HamNoSys (T2G2H) on a subset of the data from the MEINE DGS dataset (Hanke et al., 2020), where all glosses are mapped to Ham-NoSys by a dictionary look up. They find that direct T2H translation results in higher BLEU (it still needs to be clarified how well BLEU represents the quality of HamNoSys translations, though). They encode HamNoSys with BPE (Sennrich et al., 2016b), and it outperforms character-level and word-level tokenization. They also leverage BERT to create better sentence-level embeddings and use HamNoSys to extract the hand shape of a sign as additional supervision during training.

**Machine Translation Frameworks**

Machine translation has witnessed substantial advancements in recent years, both in terms of model architectures and frameworks that facilitate their training and deployment. When it comes to text-to-text translation, several open-source platforms have emerged, leading to the democratization of machine translation technology.

**Prominent machine translation frameworks**   include *OpenNMT* (Klein et al., 2017), *Sockeye* (Hieber et al., 2017, 2020), Joey NMT (Kreutzer et al., 2019), and *Fairseq* (Ott et al., 2019). They are all widely renowned for simplicity, efficiency, and emphasis on performance, promoting rapid prototyping and thus becom-ing a popular choice among machine translation researchers.

**Bergamot (2022)**   aims to bring machine translation to local clients.  Leveraging advancements in *Marian NMT* (Junczys-Dowmunt et al., 2018), *Bergamot* provides recipes for fast, local, multilingual machine translation models.  It provides an opinionated pipeline and assumes both the source and the target come from spoken languages.  It only supports text-to-text translation, and expects a shared source-target vocabulary and a huge amount of data, uncommon in sign language resources.  Despite the project's disadvantages, it is the only one that includes a realistic training pipeline for machine translation deployment.

### 6.3.3   Data

In our efforts to improve sign language translation through a text-to-text approach, data quality and quantity are of paramount importance.  This section outlines our data curation strategy, encompassing both the data we generate ourselves and the data we clean and expand.

**Fingerspelling Data**

Fingerspelling is a significant component of signed languages, often used for spelling out names, places, or other words that might not have a designated sign.  Given its importance, we collected and annotated fingerspelling for letters and numbers across 22 different signed languages[14] to be used in future machine translation systems.  These annotations are largely derived manually from the fingerspelling keyboard[15].

---

[14]American, Brazilian, British, Chinese, Danish, Flemish, French, French Belgian, German, Honduran, Irish, Israeli, Italian, Japanese, Mexican, Nicaraguan, Norwegian, Portuguese, Spanish, Swedish, Swiss German, and Thai.

[15]https://www.signwriting.org/forums/software/fingkeys/fkey001.html

**SignBank Cleaning and Expansion**

The SignBank dataset, while invaluable, includes numerous inconsistencies and imperfections. Multiple non-parallel textual entries were associated with singular signing sequences. For instance, while some entries indicated chapter and page numbers from a book, the actual text was missing. In others, definitions were jumbled with the intended word. In light of these challenges, we initiated meticulous data-cleaning and expansion processes detailed below:

**Dataset Cleaning**   Initially, we manually corrected at least five entries for each puddle. Given the formulaic nature of certain puddles (e.g., the bible), rule-based corrections enabled immediate annotation of multiple entries. Comprehensive rules used in this phase are detailed in §6.3.4.

Using ChatGPT (OpenAI, 2022), we defined a pseudo function that gets the number of signs, language code, and existing terms to return a cleaned, parallel version of the terms:

```
clean(number of signs, language code, terms).
```
An illustration would be the function call:

```
clean(1, "sl", ["Koreja (mednarodno)", "Korea", "S125-P1"])
```
returning `["Koreja", "Korea"]`. More examples are available in §6.3.5.

To ascertain the efficacy of this cleaning method, we employed the *gpt-3.5-turbo-0613* model on the manually cleaned samples. By comparing these results to the cleaned dataset, we assessed the quality via the Intersection over Union (IoU)[16] metric between the predicted terms and the annotated terms. We compared multiple settings, with various approaches to cleaning the data:

1. **E0:** No changes.

2. **E1:** Rule-based cleaning (§6.3.4).

3. **E2:** E1 + ChatGPT with four fixed, manually selected few-shot examples.

---

[16]Note: The maximum IoU is not 1. We can not ignore possible annotation errors/variations, especially when dealing with non-English data.

4. **E3:** E1 + ChatGPT with five few-shot examples from the same puddle.

5. **E4:** E1 + ChatGPT with four fixed examples and five examples from the same puddle.

6. **E5:** E4 + using *gpt-4-0613*.

Doing nothing (*E0*) leads to a base IoU of **0.50**. The rule-based approach (*E1*), which conservatively eliminated undesired text entries, provided a slight boost, resulting in an IoU of **0.53**. Incorporating general few-shot examples into the cleaning process (*E2*) significantly increased the IoU to **0.63**. A more targeted approach using five few-shot examples from the same puddle (*E3*) further improved this to **0.71** IoU. When combining the general few-shot examples with puddle-specific examples (*E4*), we achieved an IoU of **0.74**. Our best results, however, came from GPT-4 (*E5*), which achieved an IoU of **0.80**.

For cost considerations, the following pricing was assumed: $0.0015/1*K* tokens for *gpt-3.5-turbo* and $0.03/1*K* tokens for *gpt-4*, indicating a 20× price disparity. Given the average of 714 tokens for *E4* and *E5* and around 200*K* annotations, the projected costs for *gpt-3.5-turbo* and *gpt-4* are approximately $200 and $4000, respectively. For financial reasons, we use *gpt-3.5-turbo*. The final cost ended up being $230.18, paid to OpenAI.

**Dataset Expansion** Our next objective is to further enrich the dataset by introducing variations for each cleaned term. Variability in language representation can significantly benefit the robustness of machine translation models by providing multiple ways of expressing the same idea. For this, we designed a function, `expand(language code, terms)`, producing expanded terms and proper capitalization. As some terms were in English, outputs for both the specific language and English were generated separately. Prompt in §6.3.5.

For an illustration, consider a term in Swedish such as 'tre'. When passed to our function like so: `expand("sv", ["tre"])`, the returned output could be `{"sv": ["Tre", "3"], "en": ["Three", "3"]}`. This means that for the Swedish language ('sv'), the term 'tre' can be represented as 'Tre' or the numeral '3'. The corresponding English translation for the term would be 'Three'. Another example would be the German term for 'father'. The function call `expand("de", ["Vater", "father"])` yields
`{"de": ["Vater", "Vati", "Papa", "Erzeuger"],`
`"en": ["Father", "Dad", "Daddy"]}`.
Here, the term expands to multiple terms in both German and English.

This expansion approach (using *gpt-3.5-turbo* with 9 fixed few-shot examples), although seemingly straightforward with a similar cost to the cleaning process, introduces vast richness to our dataset. Each term is now associated with multiple representations, thereby enhancing the potential of our model to understand the nuances and variability of language. However, this expansion can also introduce errors, either when expanding terms that were not properly cleaned, or when the expansion itself is wrong. The expansion cost ended up being $299.72, paid to OpenAI.

Evaluating the efficacy of this expansion step is non-trivial, due to the inherent subjectivity involved in determining which expansions are valid or more useful than others. Interested readers are referred to §6.3.6 for more outputs.

### 6.3.4   Cleaning Rules (Appendix)

**Automatic Annotation Rules**

**Question Marks**   It is rare, but sometimes, this movement symbol is used as a question mark, because of visual resemblance. We remove all entries that contain only a question mark (M510x517S29f0c491x484).

**Korean (puddle 78)**   This large puddle (25*k* entries) is quite standardized. Most entries include four terms, in a predictable fashion. For all 22*k* entries that match this fashion, we annotate them with the second term, excluding the number that follows (i.e., in English, `hello3` becomes `hello`).

**Slovene (puddle 52)**   Out of 6*k* entries, about 3*k* seem to fit a specific pattern. A single term, with possibly a single uppercase letter (variation) and the source in parenthesis. For example, {`zdarma B (UPOL)`} is annotated by removing the variation and source, to result in `zdarma`.

**The Bible (puddles 151 and 152)**   These puddles include translation of the Bible into SignWriting in Signed Exact English (SEE) and not American Sign Language (ASL). Almost every entry includes a book, chapter, and verse identifier, for example `1Corinthians01v03` means The First Epistle to the Corinthians, Chapter 1, Verse 3. We only address entries that we can extract the book, chapter, and verse from, and that are of a single verse, not split apart (some entries contain parts of verses, and others contain multiple verses). Based on the match, we extract the verse from the *bible-corpus*[17], and disregard any other text in the entry. In some entries, the SignWriting starts with indicating 'Verse' and a number. We attempt to recognize when this happens based on simple string matching, and when it does, we add `Verse {number}:` to the beginning of the verse.

**Data Filtering Rules**

- We remove all terms that include a URL in them. These usually link to an image, a video, or a source.

- For Slovene entries in puddle 52 that did not match our criteria for automatic annotation, we strip the suffix as mentioned above from all terms.

- For Swiss-French in puddle 49, we remove entries that indicate the source based on the following regex: (`lexique SGBFSS|lexique SGB-FSS|` `^liste:|jeu SignEcriture|JEU-COULEURS|CCSS|ApéroSignes`). We also remove entries that start with `FMS` or `EMM`, or include a `n°`

- For Honduran in puddle 16, we remove entries that include 'SWS-TAG'.

---

[17] https://github.com/christos-c/bible-corpus

- For German in puddle 53, we remove all entries that include `vgl`, `KK` or `delegs`, or that match `Variante \d`, `Geschichte ".*?"`, `[Ss][\d.]*` or `rwth\d*`.

- For American in puddle 4, we remove all entries equal to `English sign`.

- For Argentinian in puddle 41, we remove all entries that start with `.LSC`.

- For Quebecois in puddle 47, we remove all entries that start with either `Liste:` or `Alice`. We also filter out the last term if it matches a part of speech (nom, verbe, adjectif, adverbe, pronom, préposition, conjonction, interjection, déterminant, phrase, géographie).

## 6.3.5   Prompts (Appendix)

**Data Cleaning**

**System**

```
You are a proficient assistant, responsible for data sanitization \
for a machine translation model. Your main task involves \
operating the `clean` function:

The `clean` function is specifically designed to return a list of \
accurate translations that correspond to the provided SignWriting text.
The text output must be valid spoken language text, \
fixed in case of errors.
This aids in the training of machine translation models.
Since SignWriting can also write fingerspelling, \
a single character can also be a valid output.

When confronted with an unfamiliar SignWriting string, \
the function employs heuristic methods, including:

1. A similarity in the number of words to the number of signs. \
   For example, a string with 20 signs and 3 words is unlikely \
   to have a parallel translation.
2. Potential multiple entries are considered when several terms \
   in the text have the same meaning. For instance, "one (1)" \
   or "one-also-1" or "one/1" will result in "one" and "1".
3. In cases of uncertainty, the function returns an empty array.

    Parameters:
        SignWriting (int): Number of signs present.
        Language (string): language code (Optional)
        texts (list): A list of raw, unfiltered candidate texts.

    Returns:
        translations (list): A list of strings that accurately \
        match the SignWriting, intended for machine translation.
```

```
In the course of the session, users will invoke the 'clean' \
function and you will respond with the function's output.
```

**Manually Curated Few Shots**

| User | Assistant |
| --- | --- |
| `clean(1, "sl", ["Koreja (mednarodno)", "Korea"])` | `["Koreja", "Korea"]` |
| `clean(1, "sl", ["Bosna in Hercegovina 2", "Bosnia and Herzegovina"])` | `["Bosna in Hercegovina", "Bosnia and Herzegovina"]` |
| `clean(18, "en", ["Acts 04_27-31c", "James Orlow"])` | `[]` |
| `clean(8, "es", ["Juan el Bautista predica", "1:1 El principio de la buena noticia de Jesucristo, el Hijo de Dios."])` | `["El principio de la buena noticia de Jesucristo, el Hijo de Dios."]` |

Table 6.5: Manually curated few-shot `user` and `assistant` interactions.

## Data Expansion

## System

You are a proficient assistant, responsible for data augmentation \
for a machine translation model. \
Your main task involves executing the `expand` function:

The expand function is specifically crafted to return a list \
of equivalent expressions that match the provided text \
in a certain language. The text output must be accurately \
spelled and grammatically correct spoken language text. \
This is beneficial for improving the robustness of \
machine translation models.

The function uses a series of methods to ensure a variety \
of equivalent terms, including:

1. Paraphrases in the same language. For example, the text "hello" \
   can be paraphrased as "Hi" or "Hey"
2. Capitalization corrections: For instance, the text "one" may \
   result in ["One"]. "donald duck" may result in ["Donald Duck"] \
3. Numerical translation: When the text represents a number, \
   its numeric  equivalent is added. For example, \
   "one" would return ["One", "1"].
4. Language translation: If the text is in a non-English language, \
   and it's feasible to translate it to English, the English \
   translation is included. For instance, \
   "domingo" would return ["Domingo", "Sunday"].
5. If the text is not spoken language text, but instead a random \
identifier like "rom-ale-10-44r", ignore it completely.

  Parameters:
    language (string): language code for the terms in the list.
    texts (list): A list of terms requiring expansion.

  Returns:
    expansions (obj):
      language (list): A list of unique strings that are equivalent \
      to the provided terms, intended for machine translation training.

```
    en (list): A list of unique translations to English if \
    the language is not english and translation is feasible.
```

```
During the session, users will call the `expand` function and you \
will respond with the function's output.
```

**Manually Curated Few Shots**

| User | Assistant |
|------|-----------|
| `expand("sl", ["2"])` | `{"sl": ["2", "Dva"], "en": ["2", "Two"]}` |
| `expand("fr", ["Lac", "LEMAN"])` | `{"fr": ["Lac Leman"], "en": ["Lake Geneva"]}` |
| `expand("no", ["fire", "tall", "4"])` | `{"no": ["fire", "4"], "en": ["four", "4"]}` |
| `expand("pt", ["MAGIA", "MAGICO"])` | `{"pt": ["MAGIA", "MÁGICO"], "en": ["Magic", "Magical", "Magician"]}` |
| `expand("de", ["Vater", "father"])` | `{"de": ["Vater", "Papa", "Papi", "Vati", "Erzeuger"], "en": ["Father", "Dad", "Daddy", "Papa"]}` |
| `expand("en", ["Hello", "hi"])` | `{"en": ["Hello", "Hi", "Hey", "Greetings", "Howdy", "Hiya", "Aloha", "Bonjour", "Hola", "Salutations", "Hello there", "Hi there"]}` |
| `expand("pt", ["Eu-tentar"])` | `{"pt": ["Eu tentar"], "en": ["I try"]}` |
| `expand("de", ["zwaiund zwanzig", "S3-07163-V"])` | `{"de": ["Zwei und Zwanzig", "22", "Zweiundzwanzig"], "en": ["Twenty-two", "22"]}` |

Table 6.6: Manually curated few-shot `user` and `assistant` interactions.

## 6.3.6 Dataset Preview

Here we include a preview of the dataset, including the original spoken language terms, the automatically cleaned terms based on the original texts, the human annotation based on the original text, and the expanded texts based on the human annotation. We include one sample per puddle (as long as the puddle includes a sample that is less than 200 characters).

Due to incompatibility in displaying non-latin scripts, we exclude Korean, Chinese, Arabic, Urdu, Hebrew, Russian, Ukrainian, Pashto, Japanese, Hindi, Amharic, Thai, Polish, Greek, Maltese, Bulgarian, Nepali, and Vietnamese.

| pid | eid | Lang | Original | Cleaned (from Original) | Annotation | Expanded (from Annotation) |
|---|---|---|---|---|---|---|
| 4 | 101 | en | `[glasses]` | `[glasses]` | `[Glasses]` | `[Glasses, Eyeglasses, Spectacles, Optical frames]` |
| 5 | 296 | en | `[gesture 1, heck with it]` | `[heck with it, gesture 1]` | `[heck with it]` | `[Heck with it, Forget it, Never mind, Screw it, Drop it, Let it go]` |
| 11 | 92 | en | `[cookie, biscuit, https://www.youtube.com/watch?v=jyOh9Ss7Dzs]` | `[cookie, biscuit]` | `[cookie, biscuit]` | `[Cookie, Biscuit, Cracker, Snack, Treat, Sweet, Baked good]` |
| 13 | 2 | zh-CN | `[epigrafe]` | `[]` | `[]` | `[]` |
| 16 | 2829 | es | `[soñar, SWS-TAG[SignLists:Dict Sept15 Honduras,SignLists:New Dict,Dibujos:Dibujos Grupo1]]` | `[soñar]` | `[soñar]` | `[soñar, soñando]` |
| 17 | 2 | en | `[CBS 1, Deaf Hands Story CBS 1 (as retold by Stuart)]` | `[CBS 1, Deaf Hands Story CBS 1]` | `[]` | `[]` |
| 19 | 1947 | pl | `[sobie, siebie, to oneself; oneself]` | `[to oneself, oneself, siebie, sobie]` | `[sobie, siebie, to oneself, oneself]` | `[sobie, siebie]` |
| 20 | 49 | fr | `[amygdalectomie]` | `[amygdalectomie]` | `[amygdalectomie]` | `[Amygdalectomie]` |
| 21 | 624 | en | `[VetDr04]` | `[]` | `[]` | `[]` |
| 22 | 8 | fr | [ADN, ADN-définition] | [ADN] | [] | [] |
| 23 | 17 | no | [Gullhår 12, Vi tre går en tur i skogen.] | [Vi tre går en tur i skogen.] | [Vi tre går en tur i skogen.] | [Vi tre går en tur i skogen.] |
| 24 | 4 | no | [få (ikke mange)] | [få] | [få, ikke mange] | [få, ikke mange] |
| 25 | 1268 | en | [wrong, accidental, by mistake] | [by mistake, accidental, wrong] | [wrong, accidental, by mistake] | [Wrong, Incorrect, Inaccurate, Mistaken, Accidental, Unintentional, By mistake, By accident, In error] |
| 26 | 261 | de | [Städte Teil 2, Spiel: "Stadt, Land, Fluss"] | [Spiel: Stadt, Land, Fluss, Städte Teil 2] | [] | [] |
| 27 | 14 | de | [Beispielsatz index, Thomas kauft ein Auto. Es ist billig.] | [Thomas kauft ein Auto. Es ist billig.] | [Thomas kauft ein Auto. Es ist billig.] | [Thomas kauft ein Auto. Es ist billig.] |
| 28 | 11584 | en | [455] | [455] | [455] | [455, Four hundred fifty-five] |
| 29 | 4 | de | [Vater, father] | [Vater, father] | [Vater, father] | [Vater, Papa, Papi, Erzeuger] |
| 30 | 30 | da | [R, fingerspelling] | [R] | [R] | [R] |
| 31 | 173 | mt | [Kugin, Cousin, Dizzjunarju ta' Affarijiet ta' Kuljum\n\nVolum: FAMILJA] | [Cousin, Kugin] | [Kugin, Cousin] | [Kugin, Kuzin] |
| 32 | 9 | en | [God is with us, The biblical meaning of Emmanuel] | [God is with us, Emmanuel] | [God is with us] | [God is with us] |
| 33 | 409 | pt | [expressão-facial, expressão-facial] | [expressão-facial] | [expressão-facial, expressão-facial] | [Expressão facial, Rosto, Expressão no rosto] |
| 35 | 6 | en | [Arkansas, US State, (n) a state in the United States.] | [Arkansas] | [Arkansas] | [Arkansas] |

| 36 | 156 | cs | [O perníkové chaloupce 2] | [O perníkové chaloupce 2] | [] | [] |
|----|------|-----|---------------------------|---------------------------|-----|-----|
| 37 | 9 | cs | [podtřída] | [] | [] | [] |
| 41 | 1765 | es | [Grace, .LSC vocab Personas] | [Grace] | [Grace] | [Grace] |
| 42 | 23 | en | [brother] | [brother] | [brother] | [Brother, Sibling, Bro, Buddy, Mate, Pal, Comrade, Fellow] |
| 43 | 1396 | fr | [cinéma] | [cinéma] | [cinéma] | [cinéma] |
| 44 | 5617 | nl | [Jelle, jelle] | [jelle, Jelle] | [Jelle, jelle] | [Jelle] |
| 45 | 3 | es | [walk] | [caminar] | [walk] | [caminar, andar, pasear] |
| 46 | 11173 | pt | [alfabeto] | [alfabeto] | [] | [] |
| 47 | 10094 | fr | [trésorier, trésorière, nom] | [trésorier, trésorière] | [trésorier, trésorière] | [trésorier, trésorière] |
| 48 | 3891 | de | [glcklich-2] | [glcklich] | [Glücklich, Happy] | [Glücklich, Froh, Fröhlich, Zufrieden] |
| 49 | 1267 | fr | [jeu SignEcriture, 3-11-4] | [] | [] | [] |
| 50 | 2 | it | [VAUD, canton Suisse] | [VAUD] | [VAUD] | [VAUD] |
| 51 | 801 | es | [pensamiento] | [pensamiento] | [pensamiento] | [pensamiento, reflexión, idea, concepción, cogitación] |
| 52 | 1007 | sk | [displej (IMoTeSP)] | [displej] | [displej] | [displej, obrazovka] |
| 53 | 12013 | de | [für, hier:  für 2010 (obewohl kein Bonativ?)] | [für, hier:  für 2010] | [für] | [für] |
| 54 | 1222 | eo | [ist, G@17] | [ist] | [ist] | [estas, estis, estos] |
| 55 | 34 | es | [tocar] | [tocar] | [tocar] | [tocar, reproducir, interpretar, ejecutar] |
| 56 | 1605 | ca | [geografía] | [geografía, geografia, geography] | [geografía] | [geografia, geografies] |
| 57 | 173 | fi | [TAVATA] | [TAVATA] | [TAVATA] | [TAVATA, Tavata] |

| 58 | 1375 | fr | [surnom] | [surnom] | [surnom] | [surnom, sobriquet, pseudo, surnommer] |
|---|---|---|---|---|---|---|
| 59 | 274 | en | [boy, Theme: family details, son, theme: cards] | [boy, son] | [boy, son] | [Boy, Son, Child, Youngster, Lad, Kid, Offspring, Male child] |
| 60 | 12 | en | [seven] | [seven] | [seven] | [Seven, 7] |
| 62 | 35 | en | [A, fingerspelling] | [A] | [A] | [A, One] |
| 63 | 677 | it | [(passato) incontrare] | [incontrare] | [incontrare] | [incontrare, incontrarsi] |
| 65 | 65 | es | [Uno, Uno (one)] | [Uno, one] | [Uno, one] | [Uno, 1] |
| 66 | 3 | ms | [Malaysia] | [Malaysia] | [Malaysia] | [Malaysia] |
| 67 | 1598 | es | [banarse, bathe] | [bañarse] | [banarse, bathe] | [bañarse, ducharse] |
| 68 | 68 | nl | [AANGIFTE / AANGEVEN ()] | [AANGIFTE, AANGEVEN] | [AANGIFTE, AANGEVEN] | [AANGIFTE, AANGEVEN, Melding, Melden, Aangifte doen] |
| 69 | 679 | no | [4, fire, tall] | [fire, tall, 4] | [4] | [4, Fire] |
| 70 | 33 | en | [I] | [I] | [I] | [I, Me, Myself] |
| 71 | 67 | es | [flor, flower] | [flor, flower] | [flor, flower] | [flor, flora] |
| 72 | 325 | fil | [mountain] | [mountain] | [mountain] | [bundok] |
| 73 | 72 | sv | [Ä, fingerspelling] | [Ä, ä] | [Ä] | [Ä, A med ring över] |
| 74 | 2477 | sl | [pojutrišnjem, day after tomorrow] | [pojutrišnjem, day after tomorrow] | [pojutrišnjem, the day after tomorrow] | [pojutrišnjem, pojutrišnjem dnevu] |
| 76 | 39 | es | [y] | [y] | [y] | [y, e] |
| 77 | 45 | en | [Science] | [Science] | [Science] | [Science, Scientific, Sciences] |
| 79 | 10 | sw | [see] | [see] | [see] | [see] |
| 80 | 883 | pt | [barriga] | [barriga] | [Barriga, Stomach] | [Barriga, Estômago] |
| 81 | 36 | fr | [Les cinq frères chinois 02] | [] | [] | [] |
| 82 | 115 | sq | [Itali, italy] | [Itali, Italy] | [Itali, italy] | [Itali, Italia] |

| 89 | 260 | sk | [chlapec] | [chlapec] | [chlapec] | [chlapec, chalan, kluk, chlapčenský] |
|---|---|---|---|---|---|---|
| 90 | 34 | tr | [o, he/she/it; er/sie/es] | [o, he/she/it, er/sie/es] | [o, he, she, it, er, sie, es] | [o, he, she, it, er, sie, es] |
| 91 | 28 | ar | [Goldilocks Page 1] | [] | [] | [] |
| 93 | 527 | es | [SignoEscritura Reading Lessons pg. 20] | [] | [] | [] |
| 94 | 25 | ca | [patinatge artístic] | [patinatge artístic] | [patinatge artístic] | [Patinatge artístic, Patinatge artístic sobre gel] |
| 96 | 13 | de | [Noah 15] | [] | [] | [] |
| 98 | 35 | nl | [android] | [android] | [android] | [Android] |
| 99 | 16 | ja | [texto11] | [] | [] | [] |
| 100 | 1 | am | [texto11] | [texto11] | [] | [] |
| 104 | 1181 | ar | [Bonjour] | [Bonjour] | [] | [] |
| 105 | 338 | en | [exit, leave, out] | [exit, out, leave] | [exit, leave, out] | [Exit, Leave, Out, Quit, Depart, Go away, Withdraw, Egress, Vacate] |
| 111 | 192 | en | [why] | [why] | [why] | [Why, For what reason, What is the reason, What is the purpose, What is the cause] |
| 113 | 1 | ht | [Zebra] | [Zebra] | [Zebra] | [Zebra] |
| 114 | 2190 | pt | [prova] | [prova] | [prova] | [prova, teste, exame] |
| 115 | 24 | pt | [bebe12] | [] | [] | [] |
| 116 | 1348 | pt | [Dentes superiores tocando a língua] | [Dentes superiores tocando a língua] | [] | [] |
| 117 | 8 | pt | [Isabel Morais, Nome Gesto] | [] | [] | [] |
| 119 | 404 | es | [World Explorers Part 1 pg. 06] | [] | [] | [] |

| 120 | 17 | es | [Tapa abriendo] | [Tapa abriendo] | [Tapa abriendo] | [Tapa abriendo] |
|-----|-----|-----|-----|-----|-----|-----|
| 122 | 3 | hu | [Ország, Country, Land] | [Ország, Country, Land] | [Ország, Country, Land] | [Ország, Országok] |
| 123 | 1 | hu | [raus, im ärgerlichen Befehlston] | [raus] | [raus] | [raus] |
| 124 | 15 | fr | [corbeau] | [corbeau] | [corbeau] | [corbeau, corneille, corbin] |
| 125 | 16 | en | [The farmer is in his den, part 6, Theme: practice makes perfect DVD, The dog wants a bone, the dog wants a bone, E I A D O, the dog wants a bone.] | [The farmer is in his den, part 6, The dog wants a bone, the dog wants a bone, E I A D O, the dog wants a bone.] | [The dog wants a bone, the dog wants a bone, E I A D O, the dog wants a bone.] | [The dog wants a bone, the dog wants a bone, E I A D O, the dog wants a bone.] |
| 126 | 319 | ar | [Luc4:23 (LSF)] | [] | [] | [] |
| 128 | 41 | mw | [teacher] | [teacher] | [teacher] | [teacher] |
| 129 | 5284 | gn | [Mejorar 2] | [Mejorar, 2] | [Mejorar] | [Mejorar] |
| 131 | 36 | is | [stór (pf.1)] | [stór] | [stór] | [stór, stórt] |
| 132 | 12 | ro | [luni, Transilvania Semne, Monday, Montag] | [Monday, Montag, luni] | [luni, Monday, Montag] | [luni] |
| 135 | 89 | es | [h, Letra (consonante)] | [h] | [h] | [h] |
| 137 | 14 | es | [San Salvador, san salvador, Capital (El Salvador), Geografia.] | [San Salvador] | [San Salvador, san salvador] | [San Salvador] |
| 143 | 84 | es | [abierto] | [abierto] | [abierto] | [abierto, desbloqueado, libre, sin restricciones] |
| 147 | 1154 | mt | [Taken from...Ghaqda, Mehud mit-test tal-Ghaqda Bibblika ..etc] | [Mehud mit-test tal-Ghaqda Bibblika ..etc] | [] | [] |
| 148 | 17 | sl | [Sveti Filip, prosi za nas!] | [Sveti Filip, prosi za nas!] | [Sveti Filip, prosi za nas!] | [Sveti Filip, prosi za nas!] |

| 151 | 10009 | en | [Matthew15v07 NLT, You hypocrites!  Isaiah was right when he prophesied about you, for he wrote,\n\nMatthew15v7 NLT] | [Verse 7:  You hypocrites!  Well did Isaiah prophesy of you, saying:, Verse 7:  Ye hypocrites, well did Esaias prophesy of you, saying,] | [Verse 7:  You hypocrites!  Well did Isaiah prophesy of you, saying,, Verse 7:  Ye hypocrites, well did Esaias prophesy of you, saying,] | [] |
| 152 | 10173 | en | [Proverbs 24:28, Pr 24:28 ¶ Be not a witness against thy neighbour without cause; and deceive [not] with thy lips.\n\nDo not witness against neighbor for no reason and do not deceive people.] | [Be not a witness against your neighbor without cause; and deceive not with your lips., Do not witness against neighbor without cause; and deceive not with your lips.] | [Don't be a witness against your neighbor without cause.  Don't deceive with your lips., Be not a witness against thy neighbour without cause; and deceive not with thy lips.] | [] |

### 6.3.7 Data Quality Experiments

| Dataset | Training Pairs | Vocab | Sockeye | | Fairseq | | OpenNMT | | Keras (mT5) | |
|---|---|---|---|---|---|---|---|---|---|---|
| | | | BLEU | chrF | BLEU | chrF | BLEU | chrF | BLEU | chrF |
| Original | 521,390 | 6,016 | 0.2 | 8.4 | 0.18 | 4.74 | 0.69 | 9.21 | 0.07 | 6.39 |
| Cleaned | 357,574 | 5,200 | **22.32** | **28.63** | 1.1 | **7.59** | **30.6** | **22.46** | **6.02** | 12.35 |
| Expanded | 1,027,418 | 5,976 | 0.55 | 7.22 | **1.26** | 6.52 | 13.38 | 13.0 | 2.99 | **12.49** |

Table 6.8: Evaluation of the usability of our data for machine translation.

To evaluate the quality of our cleaning and expansion, we test its effect on machine translation. We train machine translation models on the original data, on the cleaned data, and on the expanded data, in an imbalanced multilingual setting that contains all of the puddles on SignBank. For this comparison, we focus on the *signed-to-spoken* direction, since automatic evaluation of spoken language text is well established. For a development set, in each data scenario, we consider the first 3000 entries. For our test set, we use our manually annotated data from §6.3.3. In the source text, we include tags to indicate the source and target language for the translation. We use sacreBLEU 2.3.1 (Post, 2018), to evaluate BLEU[18] (Papineni et al., 2002) and chrF[19] (Popović, 2016b).

This comparison is only made to evaluate the quality of the different datasets. Thus, for every framework, we use the default training settings and avoid attempting to optimize with smaller models or different architecture. We posit that better test-set performance in a given framework indicates higher data quality. While we believe that this effect should be highly potent for the *spoken-to-signed* translation direction, it is not evaluated in this work since there are no human-validated automatic metrics to evaluate SignWriting output.

**Sockeye / Fairseq / OpenNMT**   In pre-processing, the SignWriting text is tokenized using §6.3.8, and the spoken language text is tokenized using BPE (Sennrich et al., 2016b) with 3000 merges. For the cleaned dataset, this results in a smaller vocabulary than for the original dataset since some unigrams are fil-

---

[18] BLEU = case:mixed|eff:no|tok:13a|smooth:exp
[19] chrF = case:mixed|eff:yes|nc:6|nw:0|space:no

tered out. Model training is early-stopped on validation `chrF` score (Sockeye), `BLEU` (Fairseq), and accuracy (OpenNMT) with a patience of 10 epochs.

**Keras (Chollet et al., 2015)** To address the effect of clean data on pre-trained language models, we fine-tune *mT5-small* (Xue et al., 2021) using Keras and HuggingFace Transformers (Wolf et al., 2020). In this setting, both the source and target texts are tokenized using the *mT5* tokenizer. Since our source data is extremely out-of-domain to the original language model training, we do not expect to see improvements from the pre-trained language model. The model is fine-tuned for up to 20 epochs, early stopped on validation loss.

### 6.3.8  Tokenization

We tokenize the FSW sequences into discrete tokens. For example, the American Sign Language sign for "Hello" is represented as:

```
M518x529S14c20481x471S27106503x489
```

This representation is a sequence of graphemes, each with a structure of a symbol , modifiers , and an `<x,y> position` (Table 6.9).

|     | symbol | modifiers | x | y |
| --- | --- | --- | --- | --- |
| 1. | M | - | 518 | 529 |
| 2. | S14c | 2 0 | 481 | 471 |
| 3. | S271 | 0 6 | 503 | 489 |

Table 6.9: Tokenized structure for the ASL sign for 'Hello'.

From this structure, we treat each component as a separate token. We further remove predictable and redundant symbols, such as `M`, `x` and the size of the box, resulting in the following sequence:

`M` `p518` `p529` `S14c` `c2` `r0` `p481` `p471` `S271` `c0` `r6` `p503` `p489`

This tokenization process simplifies the complex FSW strings, creating a small vocabulary of 1182 tokens for our NMT framework (4 boxes, 656 symbols, 6 plane modifiers, 16 rotation modifiers, and 500 positions).

### 6.3.9   Results

Table 6.8 shows that despite the different frameworks, pre-trained models, unoptimized modeling, and imbalanced multilingual translation scenarios, performance on the cleaned data is consistently better compared to the original data. This establishes our cleaned data as more useful for signed-to-spoken machine translation.

In the *signed-to-spoken* translation direction, the use of our expanded data is dubious. If our cleaned data is of perfectly good quality, our expansion can only add noise by introducing multiple targets for the same source. However, since we know that our cleaned data is not perfect, we hypothesize that the additional noise from the data expansion smooths out the noise in the imperfect data, by introducing more overlaps between identical translations, thus drowning the noise. This is very difficult to evaluate. As we vary the target texts in many dimensions (gender, formality, capitalization, script, and form), uncontrolled translation of the test set into the original distribution of these dimensions is improbable, even when disregarding noise coming from wrong expansions. This is reflected in the results. Using the expanded data for pre-training our Sockeye model, then fine-tuning on the cleaned data gets the model back to the target distribution, with state-of-the-art results of 31.39 `BLEU` and 31.97 `chrF`.

We compare these results to the previous work. Specifically, we query the API endpoint made available by Jiang et al. (2023a) to translate our test set. To some extent, this is an unfair comparison, since they likely saw these exact translation sources in training and since we are evaluating more languages than their model was trained on. And yet, their method achieves 5.03 `BLEU` and 18.92 `chrF` on our test set. Despite their optimization in modeling, our optimization in data quality makes up for simple modeling.

### 6.3.10 Conclusions

This work introduces a methodology for data cleaning and expansion for low-resource settings such as sign language translation. Its main contribution is the introduction of *SignBank+*, a cleaner and more expansive sign language translation dataset than *SignBank*. The data and the code for the baseline models are publically available on <https://github.com/sign-language-processing/signbank-plus>.

### 6.3.11 Future Work

We encourage future work to expand on our efforts and create *SignBank++*. The *clean* and *expand* steps can be executed with more, and better language models. Quality estimation filtering methods can be created to filter out text pairs likely to not be parallel. Additionally, optimizing the input representation, by encoding SignWriting as images (Dosovitskiy et al., 2021), reducing the token count, or standardizing phoneme order, all of which could improve translation performance. Finally, robust evaluation metrics for spoken-to-signed translation should be created and validated with human judgments.

# Chapter 7

# Sign Language Production

## 7.1 Baseline ([Moryossef et al., 2023b](#))

We explore wether or not estimated skeletal poses are viable for use in sign lan- guage translation. We introduce a sign language production baseline that smartly stitches dictionary entries. This baseline is very limited, and the limitations are throughly discussed. A large part of this section was independently published as "An Open-Source Gloss-Based Baseline for Spoken to Signed Language Translation".

Sign language translation systems are complex and require many components. As a result, it is very hard to compare methods across publications. We present an open-source implementation of a text-to-gloss-to-pose-to-video pipeline approach, demonstrating conversion from German to Swiss German Sign Language, French to French Sign Language of Switzerland, and Italian to Italian Sign Language of Switzerland. We propose three different components for the text-to-gloss translation: a lemmatizer, a rule-based word reordering and dropping component, and a neural machine translation system. Gloss-to-pose conversion occurs using data from a lexicon for three different signed languages, with skeletal poses extracted from videos. To generate a sentence, the text-to-gloss system is first run, and the pose representations of the resulting signs are stitched together.

### 7.1.1 Introduction

Sign language plays a crucial role in communication for many deaf individuals worldwide. However, producing sign language content is often a challenging, laborious, and time-consuming process, requiring skilled translators/interpreters for effective communication. Recent technological advancements have led to the development of automatic sign language translation systems, which have the potential to increase accessibility for the deaf community.

One of the critical issues in this field is the lack of a reproducible and reliable baseline. Without a baseline, it is challenging to measure the progress and effectiveness of new methods and systems. Additionally, the absence of such a baseline makes it difficult for new researchers to enter the field, hampers comparative evaluation, and discourages innovation.

Addressing this gap, this paper presents an open-source implementation of a text-to-gloss-to-pose-to-video pipeline approach for sign language translation, extending the work of Stoll et al. (2018, 2020). Our main contribution is the development of an open-source, reproducible baseline that can aid in making sign language translation systems more available and accessible, particularly in resource-limited settings. This open-source approach allows the community to identify issues, work together on improving these systems, and facilitates research into novel techniques and strategies for sign language translation.

Our approach involves three alternatives for text-to-gloss translation, including a lemmatizer, a rule-based word reordering and dropping component, and a neural machine translation (NMT) system. For gloss-to-pose conversion, we use lexicon-acquired data for three signed languages, including Swiss German Sign Language, Swiss French Sign Language, and Swiss Italian Sign Language. We extract skeletal poses using a state-of-the-art pose estimation framework, and apply a series of improvements to the poses, including cropping, concatenation, and smoothing, before applying a smoothing filter.

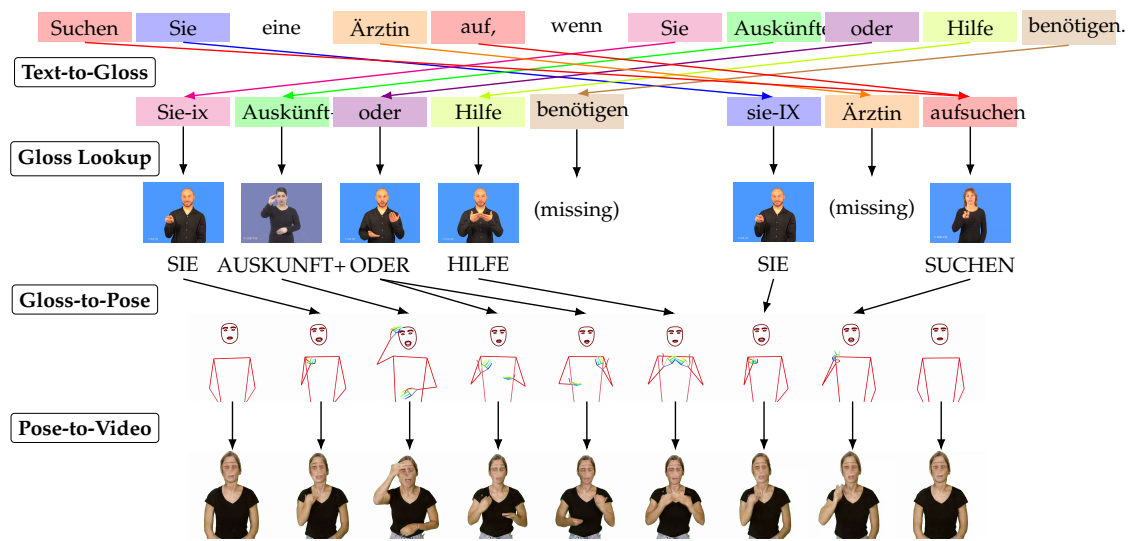

Figure 7.1: The figure depicts the entire pipeline of the proposed text-to-gloss-to-pose-to-video approach for sign language translation. Starting with a German sentence, the system applies text-to-gloss translation, for example, using a rule-based word reordering and dropping component. The resulting gloss sequence is used to search for relevant videos from a lexicon of Swiss German Sign Language (DSGS). The poses of each relevant video are then extracted and concatenated in the gloss-to-pose step to create a pose sequence for the sentence, which is then transformed back to a (synthesized) video using the pose-to-video model. The figure demonstrates the transformation of the sentence "Suchen Sie eine Ärztin auf, wenn Sie Auskünfte oder Hilfe benötigen." ('Seek out a doctor if you need information or assistance.') to a sequence of glosses, the search for relevant videos for each gloss, the concatenation of pose videos, and the final video output.

## 7.1.2   Background

Sign language translation can be accomplished in various ways. In this section, we focus on the pipeline approach that involves text-to-gloss, gloss-to-pose, and, optionally, pose-to-video techniques.  The text-to-gloss technique translates spoken language text into sign language glosses, which are then converted into a sequence of poses by gloss-to-pose techniques, and into a photorealistic video using pose-to-video techniques.

This pipeline offers the benefit of preserving the content of the sentence, while exhibiting a tendency for verbosity and a lower degree of fluency.  In this section, we explore each of the pipeline components comprehensively and examine recent progress in sign language translation utilizing these methods.

**Text-to-Gloss**

Text-to-gloss, an instantiation of sign language translation, is the task of translating between a spoken language text and sign language glosses.  It is an appealing area of research because of its simplicity for integrating into existing NMT pipelines, despite recent works (Yin and Read, 2020a; Müller et al., 2023) claiming that glosses are an inefficient representation of sign language, and that glosses are not a complete representation of signs (Pizzuto et al., 2006).  Zhao et al. (2000) used a Tree Adjoining Grammar (TAG)-based system to translate English sentences to American Sign Language (ASL) gloss sequences.  They parsed the English text and simultaneously assembled an ASL gloss tree, using Synchronous TAGs (Shieber and Schabes, 1990; Shieber, 1994), by associating the ASL elementary trees with the English elementary trees and associating the nodes at which subsequent substitutions or adjunctions can occur. Synchronous TAGs have been used for translation between spoken languages (Abeillé et al., 1991), but this was the first application to a signed language.

For the automatic translation of gloss-to-text, Othman and Jemni (2012) identified the need for a large parallel sign language gloss and spoken language text corpus. They developed a part-of-speech-based grammar to transform En-glish sentences from the Gutenberg Project ebooks collection (Lebert, 2008) into American Sign Language gloss. Their final corpus contains over 100 million synthetic sentences with 800 million words and is the most extensive English-ASL gloss corpus we know of. Unfortunately, it is hard to attest to the quality of the corpus, as the authors did not evaluate their method on real English-ASL gloss pairs.

Egea Gómez et al. (2021) presented a syntax-aware transformer for this task, by injecting word dependency tags to augment the embeddings inputted to the encoder. This involves minor modifications in the neural architecture leading to negligible impact on computational complexity of the model. Testing their model on the RWTH-PHOENIX-Weather-2014T (Camgöz et al., 2018), they demonstrated that injecting this additional information results in better translation quality.

**Gloss-to-Pose**

Gloss-to-Pose, subsumed under the task of sign language production, is the task of producing a sequence of poses that adequately represent a sequence of signs written as gloss.

To produce a sign language video, Stoll et al. (2018) constructed a lookup table between glosses and sequences of 2D poses. They aligned all pose sequences at the neck joint of a reference skeleton and grouped all sequences belonging to the same gloss. Then, for each group, they applied dynamic time warping and averaged out all sequences in the group to construct the mean pose sequence. This approach suffers from not having an accurate set of poses aligned to the gloss and from unnatural motion transitions between glosses.

To alleviate the downsides of the previous work, Stoll et al. (2020) constructed a lookup table of gloss to a group of sequences of poses rather than creating a mean pose sequence. They built a Motion Graph (Min and Chai, 2012), which is a Markov process used to generate new motion sequences that are representative of natural motion, and selected the motion primitives (sequence of poses) per gloss with the highest transition probability. To smooth that sequence and reduce unnatural motion, they used a Savitzky–Golay motion transition smoothing filter (Savitzky and Golay, 1964). Moryossef et al. (2023b) re-implemented their approach and made it open-source.

Huang et al. (2021) used a new non-autoregressive model to generate a sequence of poses for a sequence of glosses. They argued that existing models like Saunders et al. (2020a) are prone to error accumulation and high inference latency due to their autoregressive nature. Their model performs gradual upsampling of the poses, by starting with a pose including only two joints in the first layer, and gradually introducing more keypoints. They evaluated their model on the Phoenix-14T dataset (Forster et al., 2014) using Dynamic Time Warping (DTW) (Berndt and Clifford, 1994) to align the poses before computing Mean Joint Error (DTW-MJE). They demonstrated that their model outperforms existing methods in terms of accuracy and speed, making it a promising approach for fast and high-quality sign language production.

### Pose-to-Video

Pose-to-Video, also known as motion transfer or skeletal animation in the field of robotics and animation, is the conversion of a sequence of poses to a video. This task is the final "rendering" of sign language in a visual modality.

Chan et al. (2019) demonstrated a semi-supervised approach where they took a set of videos, ran pose estimation with OpenPose (Cao et al., 2019), and learned an image-to-image translation (Isola et al., 2017) between the rendered skeleton and the original video. They demonstrated their approach on human dancing, extracting poses from a choreography and rendering any person as if *they* were dancing. They predicted two consecutive frames for temporally coherent video results and introduced a separate pipeline for a more realistic face synthesis, although still flawed.

Wang et al. (2018) suggested a similar method using DensePose (Güler et al., 2018) representations in addition to the OpenPose (Cao et al., 2019) ones. They formalized a different model, with various objectives to optimize for, such as background-foreground separation and temporal coherence by using the previ-ous two timestamps in the input.

Using the method of Chan et al. (2019) on "Everybody Dance Now", Giró-i Nieto (2020) asked, "Can Everybody Sign Now?" and investigated if people could understand sign language from automatically generated videos. They conducted a study in which participants watched three types of videos: the original signing videos, videos showing only poses (skeletons), and reconstructed videos with realistic signing. The researchers evaluated the participants' understanding after watching each type of video. Results revealed a preference for reconstructed videos over skeleton videos. However, the standard video synthesis methods used in the study were not effective enough for clear sign language translation. Participants had trouble understanding the reconstructed videos, suggesting that improvements are needed for better sign language translation in the future.

As a direct response, Saunders et al. (2020b) showed that like in Chan et al. (2019), where an adversarial loss was added to specifically generate the face, adding a similar loss to the hand generation process yielded high-resolution, more photo-realistic continuous sign language videos. To further improve the hand image synthesis quality, they introduced a keypoint-based loss function to avoid issues caused by motion blur.

In a follow-up paper, Saunders et al. (2021) introduced the task of Sign Language Video Anonymisation (SLVA) as an automatic method to anonymize the visual appearance of a sign language video while retaining the original sign language content. Using a conditional variational autoencoder framework, they first extracted pose information from the source video to remove the original signer appearance, then generated a photo-realistic sign language video of a novel appearance from the pose sequence. The authors proposed a novel style loss that ensures style consistency in the anonymized sign language videos.

**Sign Language Avatars**

**JASigning** is a virtual signing system that generates sign language performances using virtual human characters. This system evolved from the earlier

SiGMLSigning system, which was developed during the ViSiCAST (Bangham et al., 2000; Elliott et al., 2000) and eSIGN (Zwitserlood et al., 2004) projects, and later underwent further development as part of the Dicta-Sign project (Matthes et al., 2012; Efthimiou et al., 2012).

Originally, JASigning relied on Java JNLP apps for standalone use and integration into web pages. However, this approach became outdated due to the lack of support for Java in modern browsers. Consequently, the more recent CWA Signing Avatars (CWASA) system was developed, which is based on HTML5, utilizing JavaScript and WebGL technologies.

SiGML (Signing Gesture Markup Language) (Elliott et al., 2004) is an XML application that enables the transcription of sign language gestures. SiGML builds on HamNoSys, and indeed, one variant of SiGML is essentially an encoding of HamNoSys manual features, accompanied by a representation of non-manual aspects. SiGML is the input notation used by the JASigning applications and web applets. A number of editing tools for SiGML are available, mostly produced by the University of Hamburg.

The system parses the English text into SiGML before mapping it onto a 3D signing avatar that can produce signing. CWASA then uses a large database of pre-defined 3D signing avatar animations, which can be combined to form new sentences. The system includes a 3D editor, allowing users to create custom signing avatars and animations.

**PAULA (Davidson, 2006)** is a computer-based sign language avatar, initially developed for teaching sign language to hearing adults. The avatar is a 3D model of a person with a sign vocabulary that is manually animated. It takes an ASL utterance as a stream of glosses, performs syntactic and morphological modifications, decides on the appropriate phonemes and timings, and combines the results into a 3D animation of the avatar. Over the years, several techniques were used to make the avatar look more realistic.

Over the years, several advancements have been made to enhance the realism and expressiveness of the PAULA avatar, such as refining the eyebrow motion to appear more natural (Wolfe et al., 2011), combining emotion and co-occurring facial nonmanual signals (Schnepp et al., 2012, 2013), improving smoothness while avoiding robotic movements (McDonald et al., 2016), and facilitating simultaneity (McDonald et al., 2017). Other developments include interfacing with sign language notation systems like AZee (Filhol et al., 2017), enhancing mouthing animation (Johnson et al., 2018; Wolfe et al., 2022), multi-layering facial textures and makeup (Wolfe et al., 2019), and applying adverbial modifiers (Moncrief, 2020, 2021).

Additional improvements to PAULA focus on making the avatar more life-like by relaxing wrist orientations and other extreme "mathematical" angles (Filhol and McDonald, 2020), refining hand shape transition, relaxation, and collision (Baowidan, 2021), implementing hierarchical transitions (McDonald and Filhol, 2021), creating more realistic facial muscle control (McDonald et al., 2022), and supporting geometric relocations (Filhol and McDonald, 2022).

**SiMAX (Sign Time GmbH, 2020)** is a software application developed to transform textual input into 3D animated sign language representations. Utilizing a comprehensive database and the expertise of deaf sign language professionals, SiMAX ensures accurate translations of both written and spoken content. The process begins with the generation of a translation suggestion, which is subsequently reviewed and, if necessary, modified by deaf translators to ensure accuracy and cultural appropriateness. These translations are carried out by a customizable digital avatar that can be adapted to reflect the corporate identity or target audience of the user. This approach offers a cost-effective alternative to traditional sign language video production, as it eliminates the need for expensive film studios and complex video technology typically associated with such productions.

**Image and Video Generation Models** Most recently in the field of image and video generation, there have been notable advances in methods such as Style-Based Generator Architecture for Generative Adversarial Networks (Karras et al., 2018, Karras et al. (2020),Karras et al. (2021)), Variational Diffusion

Models (Kingma et al., 2021), High-Resolution Image Synthesis with Latent Diffusion Models (Rombach et al., 2021), High Definition Video Generation with Diffusion Models (Ho et al., 2022), and High-Resolution Video Synthesis with Latent Diffusion Models (Blattmann et al., 2023). These methods have significantly improved image and video synthesis quality, providing stunningly realistic and visually appealing results.

However, despite their remarkable progress in generating high-quality images and videos, these models trade-off computational efficiency. The complexity of these algorithms often results in slower inference times, making real-time applications challenging. On-device deployment of these models provides benefits such as lower server costs, offline functionality, and improved user privacy. While compute-aware optimizations, specifically targeting hardware capabilities of different devices, could improve the inference latency of these models, Chen et al. (2023) found that optimizing such models on top-of-the-line mobile devices such as the Samsung S23 Ultra or iPhone 14 Pro Max can decrease per-frame inference latency from around 23 seconds to around 12.

ControlNet (Zhang and Agrawala, 2023) recently presented a neural network structure for controlling pretrained large diffusion models with additional input conditions. This approach enables end-to-end learning of task-specific conditions, even with a small training dataset. Training a ControlNet is as fast as fine-tuning a diffusion model and can be executed on personal devices or scaled to large amounts of data using powerful computation clusters. ControlNet has been demonstrated to augment large diffusion models like Stable Diffusion with conditional inputs such as edge maps, segmentation maps, and keypoints. One of the applications of ControlNet is pose-to-image translation control, which allows the generation of images based on pose information. Although this method has shown promising results, it still requires retraining the model and does not inherently support temporal coherency, which is important for tasks like sign language translation.

In the near future, we can expect many works on controlling video diffusion models directly from text for sign language translation. These models will likely generate visually appealing and realistic videos. However, they may still make mistakes and be limited to scenarios with more training data available. Developing models that can accurately generate sign language videos from text or pose information while maintaining visual quality and temporal coherency will be essential for advancing the field of sign language production.

### 7.1.3 Method

In this section, we provide an overview of our text-to-gloss-to-pose-to-video pipeline, detailing the components and how they work together to convert input spoken language text into a sign language video. The pipeline consists of three main components: text-to-gloss translation, gloss-to-pose conversion, and pose-to-video animation. For text-to-gloss translation, we provide three different alternatives: a lemmatizer, a rule-based word reordering and dropping component, and a neural machine translation system. Figure 7.1 illustrates the entire pipeline and its components.

**Pipeline**

Below, we describe the structure of our pipeline, including the text-to-gloss translation, gloss-to-pose conversion, and pose-to-video animation components:

1. **Text-to-Gloss Translation:** The input (spoken language) text is first processed by the text-to-gloss translation component, which converts it into a sequence of glosses.

2. **Gloss-to-Pose Conversion:** The sequence of glosses generated from the previous step is then used to search for relevant videos from a lexicon of signed languages (e.g., DSGS, LSF-CH, LIS-CH). We extract the skeletal poses from the relevant videos using a state-of-the-art pre-trained pose estimation framework. These poses are then cropped, concatenated, and smoothed, creating a pose representation for the input sentence.

3. **Pose-to-Video Generation:** The processed pose video is transformed back into a synthesized video using an image translation model, based on a custom training of Pix2Pix.

**Implementation Details**

Our system accepts spoken language text as input and outputs an *.mp4* video file, or a binary *.pose* file, which can be handled by the *pose-format* library Moryossef et al. (2021a) in Python and JavaScript. The *.pose* file represents the sign language pose sequence generated from the input text. To make our system easy to use, we deploy it as an HTTP endpoint that receives text as input and outputs the *.pose* file. We provide a demonstration of our system using `https://sign.mt`, with support for the three signed languages of Switzerland.

We implement our pipeline using Python and package it using Flask, a lightweight web framework. This allows us to create an HTTP endpoint for our application, making it easy to integrate with other systems and web applications. Our system is deployed on a Google Cloud Platform (GCP) server, providing scalability and easy access. Furthermore, we release the source code of our implementation as open-source software, allowing others to build upon our work and contribute to improving the accessibility of sign language translation systems.

By implementing our system as an open-source Python application and deploying it as an HTTP endpoint, we aim to facilitate collaboration and improvements to sign language translation systems.

## 7.1.4   Text-to-Gloss

We explore three different components as part of text-to-gloss translation, including a lemmatizer, a rule-based word reordering and dropping component, and a neural machine translation (NMT) system.

**Lemmatizer**

We use the *Simplemma* simple multilingual lemmatizer for Python (Barbaresi, 2023) to reduce words to their base form (i.e., lemma), which is useful for our case, as it helps to preserve meaning while reducing the complexity of the input. This approach is limited by the use of the simplistic context-free lemmatizer, since no sense information is captured in the lemma, which causes ambiguity.

**Word Reordering and Dropping**

We generate near-glosses for sign language from spoken language text using a rule-based approach. The process from converting spoken language sentences into sign language gloss sequences can be naively summarized by a removal of word inflection, an omission of punctuation and specific words, and word reordering. To address these differences, we adopt the rule-based approach from Moryossef et al. (2021c) to generate near-glosses from spoken language: lemmatization of spoken words, PoS-dependent word deletion, and word order permutation. With their permission, we re-share these rules:

Specifically, we use spaCy (Montani et al., 2023) for lemmatization, PoS tagging, and dependency parsing. Unlike Simplelemma, the spaCy lemmatizer is language-specific and context-based. We drop words that are not content words (e.g., articles, prepositions), as they are largely unused in signed languages, but keep possessive and personal pronouns as well as nouns, verbs, adjectives, adverbs, and numerals. We devise a short list of syntax transformation rules based on the grammar of the sign language and the corresponding spoken language. We identify the subject, verb, and object in the input text, and reorder them to match the order used in the signed language. For example, for German-to-German Sign Language (*Deutsche Gebärdensprache*, DGS), we reorder SVO sentences to SOV, move verb-modifying adverbs and location words to the start of the sentence (a form of topicalization), and move negation words to the end.

The specific rules we use for German to DGS/DSGS are:

1. For each subject-verb-object triplet $(s, v, o) \in \mathcal{S}$, swap the positions of $v$ and $o$ in $\mathcal{S}$

2. Keep all tokens $t \in \mathcal{S}$ if $\textbf{PoS}(t) \in \{$noun, verb, adjective, adverb, numeral, pronoun$\}$

3. If **PoS**($t$) = adverb and **HEAD**($t$) = verb, move $t$ to the start of $S$

4. If **NER**($t$) = location, move $t$ to the start of $S$

5. If **DEP**($t$) = negation, move $t$ to the end of $S$

6. Lemmatize all tokens $t \in \mathcal{S}$

We first split each sentence into separate clauses and reorder them before we apply these rules to each clause. Reordering the clauses may be needed for conditional sentences where the conditional subordinate clause should precede the main clause, as in "if...then...". These rules allow us to transform spoken language text into near-glosses that more closely match the word order and structure of sign language. Overall, our rule-based approach provides a flexible and effective way to generate near-glosses for sign language from spoken language text, with the ability to incorporate language-specific rules to capture the nuances of different sign languages. This approach employs a more accurate lemmatizer, however, it still suffers from word sense ambiguity.

**Neural Machine Translation**

As an alternative to rule-based transformations of text to glosses, we train a neural machine translation (NMT) system.

**Data**   We use the Public DGS Corpus, a publicly available corpus of German Sign Language videos with annotated glosses (Hanke et al., 2020). We hold out a random sample of 1k training examples each for development and testing purposes. Table 7.1 overviews the number of sentence pairs in all splits.

We download and process release 3.0 of the corpus. To DGS glosses we apply the following modifications derived from the DGS Corpus transcription conventions (Konrad et al., 2022):

- Removing entirely two specific gloss types that cannot possibly help the translation: `$GEST-OFF` and `$$EXTRA-LING-MAN`.

| Partition | Available Languages | | | |
|---|---|---|---|---|
| | EN | DGS·DE | DGS·EN | DE |
| Train | 61912 | 61912 | 61912 | 61912 |
| Dev | 1000 | 1000 | 1000 | 1000 |
| Test | 1000 | 1000 | 1000 | 1000 |
| **Total** | **63912** | **63912** | **63912** | **63912** |

Table 7.1: Number of sentence pairs used for gloss models.
DGS·DE=original gloss transcriptions,
DGS·EN=DGS glosses translated to English.

- Removing *ad-hoc* deviations from citation forms, marked by `*`. Example: `ANDERS1*` → `ANDERS1`.

- Removing the distinction between type glosses and subtype glosses, marked by `^`. Example: `WISSEN2B^` → `WISSEN2B`.

- Collapsing phonological variations of the same type that are meaning-equivalent. Such variants are marked with uppercase letter suffixes. Example: `WISSEN2B` → `WISSEN2`.

- Deliberately keep numerals (`$NUM`), list glosses (`$LIST`) and finger alphabet (`$ALPHA`) intact, except for removing handshape variants.

See Table 7.2 for examples for this preprocessing step. Overall these simplifications should reduce the number of observed forms while not affecting the machine translation task. For other purposes such as linguistic analysis our preprocessing would of course be detrimental.

**Preprocessing**  Our preprocessing and model settings are inspired by OPUS-MT (Tiedemann and Thottingal, 2020). The only preprocessing step that we apply to all data is Sentencepiece segmentation (Kudo, 2018). We learn a shared vocabulary with a desired total size of 1k pieces.

We additionally preprocess DGS glosses in a corpus-specific way, informed by the DGS Corpus glossing conventions (Konrad et al., 2022). See Table 7.2 for examples for this preprocessing step. Overall the desired effect is to reduce the number of observed forms while not altering the meaning itself.

| | |
|---|---|
| **Before** | `$INDEX1 ENDE1^ ANDERS1* SEHEN1 MÜNCHEN1B* BEREICH1A*` |
| **After** | `$INDEX1 ENDE1 ANDERS1 SEHEN1 MÜNCHEN1 BEREICH1` |
| **Before** | `ICH1 ETWAS-PLANEN-UND-UMSETZEN1 SELBST1A* KLAPPT1*` |
| | `$GEST-OFF^ BIS-JETZT1 GEWOHNHEIT1* $GEST-OFF^*` |
| **After** | `ICH1 ETWAS-PLANEN-UND-UMSETZEN1 SELBST1 KLAPPT1` |
| | `BIS-JETZT1 GEWOHNHEIT1` |

Table 7.2: Examples for preprocessing of DGS glosses.

**Core model settings** We train NMT models with Sockeye 3 (Hieber et al., 2022). The models are standard Transformer models (Vaswani et al., 2017), except with some hyperparameters modified for a low-resource scenario. E.g., dropout rate is set to a high value of 0.5 for all dropout layers of the model (Sennrich and Zhang, 2019).

The NMT system itself is trained with three-way weight tying between the source embeddings, target embeddings matrix and softmax output (Press and Wolf, 2017).

We train a multilingual model, following the methodology described in Johnson et al. (2017) which inserts special tokens into all source sentences to indicate the desired target language. For comparison, we also train bilingual systems that can translate in only one direction each. Our automatic evaluation confirms that one multilingual system leads to higher translation quality than individual bilingual systems.

We perform an automatic evaluation of translation quality. We measure translation quality with BLEU (Papineni et al., 2002) and CHRF (Popović, 2016a), computed with the tool SacreBLEU (Post, 2018). See Table 7.3 for all SacreBLEU signatures.

Whenever gloss output is evaluated we disable BLEU's internal tokenization, as advocated by Müller et al. (2023). Earlier works did not consider this detail and therefore our BLEU scores may appear low in comparison.

| | |
|---|---|
| **BLEU with internal tokenization** | `BLEU+case.mixed+numrefs.1+smooth.exp+tok.13a+version.1.4.14` |
| **BLEU without internal tokenization** | `BLEU+case.mixed+numrefs.1+smooth.exp+tok.none+version.1.4.14` |
| **CHRF** | `chrF2+numchars.6+space.false+version.1.4.14` |

Table 7.3: SacreBLEU signatures for evaluation metrics.

| | DGS→DE | DE→DGS |
|---|---|---|
| Bilingual | 28.610 | - |
| Bilingual | - | 32.920 |
| Multilingual: all DE and DGS directions | 28.210 | 34.760 |

Table 7.4: CHRF scores of the multilingual translation system compared to bilingual systems.

Finally, because DGS glosses are preprocessed in a corpus-specific way (see above), they are evaluated against a preprocessed reference as well, since this process cannot be reversed after translation. This means that corpus-specific preprocessing for DGS glosses simplifies the translation task overall, compared to a system that predicts glosses in their original forms.

Table 7.4 reports the translation quality of our machine translation systems, as measured by CHRF. The table shows that one multilingual system that can translate between DGS and German leads to higher translation quality than two bilingual systems.

**Language Dependent Implementation**

In this paper, we study three sign languages: LIS-CH, LSF-CH and DSGS. For LIS-CH and LSF-CH we always apply our simple lemmatizer for the text-to-gloss step. The lemmatizer-only component is universally applicable to many more languages. However, it is worth noting that this approach does not capture the full spectrum of syntactic and morphological changes necessary in going from a spoken to a signed language, which leads to suboptimal translations.

For DSGS, we explored different options for text-to-gloss, comparing the lemmatizer, rule-based system and NMT system. We observed that the glosses output by the NMT system are less accurate than rule-based reordering. A potential explanation for this is that the system is trained on German Sign Language (DGS) data. Due to the inherent differences between DGS and DSGS, using the NMT system could result in inaccurate translations or out-of-lexicon glosses. Furthermore, we found that the NMT system is not robust to out-of-domain text or capitalization differences, which further limits its applicability in these scenarios.

In the end, for DSGS we opted to employ our rule-based system (§7.1.4), which has been tailored to accommodate the unique linguistic characteristics of DSGS, and produces the best results.

### 7.1.5 Gloss-to-Pose

Gloss-to-pose translation involves converting sign language glosses into a sequence of poses that adequately represent a sequence of signs.

We use the SignSuisse dataset (Schweizerischer Gehörlosenbund SGB-FSS, 2023), which consists of sign language videos in three different languages. We extract skeletal poses from these videos using Mediapipe Holistic (Grishchenko and Bazarevsky, 2020), a state-of-the-art pose estimation framework that estimates 3D coordinates of various landmarks on the human body, including the face, hands, and body. We preprocess the poses by ensuring that the `body` wrists are in the same location as the `hand` wrists, removing the legs, hands, and face from the body pose, and cropping the videos in the beginning and end to avoid returning to a neutral body position.

We concatenate the poses for each gloss by finding the best 'stitching' point that minimizes L2 distance. We then concatenate these poses, adding 0.2 seconds of 'padding' in between, before applying cubic smoothing on each joint to ensure smooth transitions between signs, and filling in missing keypoints. Finally, we apply a Savitzky-Golay motion transition smoothing filter (Savitzky and Golay, 1964), similar to Stoll et al. (2020), to reduce unnatural motion.

### 7.1.6    Pose-to-Video

We use a semi-realistic human-like avatar system to animate the poses generated by our approach. The avatar system is a Pix2Pix model (Isola et al., 2017) adjusted to operate on pose sequences, not individual images. With her permission, we use the likeness of Maayan Gazuli[1]. We use OpenCV (Bradski, 2000) to render the poses as images and feed them into the Pix2Pix model to generate realistic-looking video frames. The avatar system can run in real-time on supported devices and is integrated into `https://sign.mt` (Moryossef, 2023c). This system is far from the state of the art, however, we believe that the open-source nature of it will bring rapid improvements, like faster inference speed, and higher animation quality.

### 7.1.7    Future Work

Here we include several future work directions that we believe have the potential to further enhance the performance and user experience of our system for text-to-gloss-to-pose-to-video generation, and we look forward to exploring these possibilities in the future, together with the open-source community.

### 7.1.8    Qualitative Evaluation

To evaluate the effectiveness of our approach, we will conduct a study to gather first impressions from deaf users. We already recruited a group of deaf individuals and will ask them to use our system to translate text into sign language.

Each participant will be asked to provide feedback on the system after using it to translate five different sentences from German into DSGS. We will provide the sentences to the participants, and they will be asked to sign the translations generated by our system. After each sentence, the participant will be asked to provide feedback on the accuracy of the translation, the quality of the poses and/or synthesized video, and the overall usability of the system.

---

[1] `https://nlp.biu.ac.il/~amit/datasets/GreenScreen/`

**Gloss Sense Disambiguation**

The current approach to text-to-gloss translation relies on a simple lemmatizer and a rule-based word reordering and dropping component, which can lead to ambiguity in the glosses produced. In the future, we can enhance our system by incorporating gloss sense disambiguation to better capture the intended meaning of the input text. Our NMT approach responds with gloss IDs from the MeineDGS corpus, which already are sense-disambiguated. Annotation of our sign language lexicon with senses will allow us to retrieve the relevant sense.

**Handling Unknown Glosses**

Where we encounter a gloss that does not exist in our lexicon, we propose exploring alternative methods to generate a video for it. One possible solution is to leverage another lexicon that includes a written representation of the gloss in question (e.g., SignWriting Sutton (1990) or HamNoSys Prillwitz and Zienert (1990)), or to employ a neural machine translation system to translate the individual concept to a writing system. Utilizing the capabilities of machine translation to embed words, we can perform a fuzzy match, addressing issues such as synonyms.

Additionally, for named entities such as proper nouns and place names that are not covered by our current gloss-to-pose conversion system, we could revert to fingerspelling them.

Once we have the written representation, we can use a system like Ham2Pose Arkushin et al. (2023) to generate a single sign video from the writing. When combined with fingerspelling for named entities, this approach should enable greater coverage of the language.

**Handling Unknown Gloss Variations**

In situations where the required gloss variation is not present in the lexicon but a related gloss exists, we propose developing a system that can modify the known gloss to generate the desired variation. This would allow for better han-dling of unknown gloss variations and increase the accuracy of the information conveyed by the signing.

**Number Forms**    For words like *KINDER* (children), we may encounter glosses such as *KIND+*, which represent "child" in plural form. Assuming that we have *KIND* in our lexicon but not *KINDER*, a system could be developed to modify signs to plural forms, such as by repeating movements or incorporating specific handshapes or locations that indicate plurality in the target sign language. Conversely, if we only have the plural form of a gloss in our lexicon, the system could be designed to generate the singular form by removing or modifying the elements that indicate plurality.

**Part of Speech Conversion**    Another challenge arises when nouns or verbs exist in the lexicon, but their counterparts do not. For instance, if *HELFEN* (to help) is present in the dictionary as a verb, but *HILFE* (help) does not exist as a noun, a system could be designed to modify signs from one part of speech to another, such as from verb to noun or noun to verb. This system could potentially involve morphological or movement modifications, depending on the linguistic rules of the target sign language.

### Post-editing Pose Sequences

The current approach generates a sequence of poses that represent a sign language sentence. We believe that there is also room for improvement in terms of the fluency and naturalness of the generated sequence. Exploring the use of automatic post-editing techniques is necessary. One such approach could identify datasets that include sentences and gloss sequences, such as the Public DGS Corpus, then, using our gloss-to-pose approach generate a pose sequence with poses from the lexicon, and could learn a diffusion model between the synthetic and real pose sequences.

### 7.1.9 Conclusions

We presented an implementation of a text-to-gloss-to-pose-to-video pipeline for sign language translation, focusing on Swiss German Sign Language, Swiss French Sign Language, and Swiss Italian Sign Language. Our approach comprises three main components: text-to-gloss translation, gloss-to-pose conversion, and pose-to-video animation.

We explained the structure of our system and discussed its limitations, as well as future work directions to address them. These directions have the potential to improve our system, and we look forward to exploring them in collaboration with the open-source community.

The main contribution of this paper is the creation of a reproducible baseline for spoken to signed language translation. The system should serve as a baseline for comparison with more sophisticated sign language translation systems and can be improved upon by the community. You can try our system for the three signed languages of Switzerland on `https://sign.mt`.

## 7.2   Translation (Jiang et al., 2023a)

Our sign language production pipeline relies critically on the translation from spoken language text to SignWriting. Further technical insights into our bi-directional translation models are covered in Jiang et al. (2023a) and additional background can be found in Section 6.3.

While we delve into the analysis and comparison of models for translating from SignWriting to spoken language in Section 6.3, we abstain from such scrutiny in the opposite direction. The primary constraint here is the absence of established metrics to assess the quality of SignWriting translations, which complicates and raises the cost of performance evaluation.

We encourage researchers in the sign language domain to develop suitable evaluation frameworks for this issue. The ability to automatically gauge the similarity between a reference and a hypothesis in SignWriting is a precondition for furthering research in this field.

## 7.3 Production (Arkushin et al., 2023)

In the course of this thesis, I had the honor of overseeing the research detailed in Arkushin et al. (2023). Together, we developed a language-agnostic framework for sign language production, leveraging written phonetic representations of various signed languages. This section serves as a brief overview of the existing landscape of sign language production. It aims not only to contextualize the current state of the field but also to highlight the innovative impact of our contributions within this larger framework.

### 7.3.1 Signed Text to Pose

Arkushin et al. (2023) proposed Ham2Pose, a model to animate HamNoSys into a sequence of poses. They first encode the HamNoSys into a meaningful "context" representation using a transform encoder, and use it to predict the length of the pose sequence to be generated. Then, starting from a still frame they used an iterative non-autoregressive decoder to gradually refine the sign over $T$ steps, In each time step $t$ from $T$ to 1, the model predicts the required change from step $t$ to step $t-1$. After $T$ steps, the pose generator outputs the final pose sequence. Their model outperformed previous methods like Saunders et al. (2020c), generating more realistic sign language sequences.

### 7.3.2 Spoken Text to Pose

Text-to-Pose, also known as sign language production, is the task of producing a sequence of poses that adequately represent a spoken language text in sign language, as an intermediate representation to overcome challenges in animation. Most efforts use poses as an intermediate representation to overcome the challenges in generating videos directly, with the goal of using computer animation or pose-to-video models to perform video production.

Saunders et al. (2020c) proposed Progressive Transformers, a model to translate from discrete spoken language sentences to continuous 3D sign pose se-quences in an autoregressive manner. Unlike symbolic transformers (Vaswani et al., 2017), which use a discrete vocabulary and thus can predict an end-of-sequence (EOS) token in every step, the progressive transformer predicts a *counter* $\in [0, 1]$ in addition to the pose. In inference time, *counter* $= 1$ is considered the end of the sequence. They tested their approach on the RWTH-PHOENIX-Weather-2014T dataset using OpenPose 2D pose estimation, uplifted to 3D (Zelinka and Kanis, 2020), and showed favorable results when evaluating using back-translation from the generated poses to spoken language. They further showed (Saunders et al., 2020a) that using an adversarial discriminator between the ground truth poses and the generated poses, conditioned on the input spoken language text, improves the production quality as measured using back-translation.

To overcome the issues of under-articulation seen in the above works, Saunders et al. (2020b) expanded on the progressive transformer model using a Mixture Density Network (MDN) (Bishop, 1994) to model the variation found in sign language. While this model underperformed on the validation set, compared to previous work, it outperformed on the test set.

Zelinka and Kanis (2020) presented a similar autoregressive decoder approach, with added dynamic-time-warping (DTW) and soft attention. They tested their approach on Czech Sign Language weather data extracted from the news, which is not manually annotated, or aligned to the spoken language captions, and showed their DTW is advantageous for this kind of task.

Xiao et al. (2020) closed the loop by proposing a text-to-pose-to-text model for the case of isolated sign language recognition. They first trained a classifier to take a sequence of poses encoded by a BiLSTM and classify the relevant sign, then proposed a production system to take a single sign and sample a constant length sequence of 50 poses from a Gaussian Mixture Model. These components are combined such that given a sign class $y$, a pose sequence is generated, then classified back into a sign class $\hat{y}$, and the loss is applied between $y$ and $\hat{y}$, and not directly on the generated pose sequence. They evaluate their approach on the CSL dataset (Huang et al., 2018) and show that their generated pose sequences almost reach the same classification performance as the refer-ence sequences.

Due to the need for more suitable automatic evaluation methods for generated signs, existing works resort to measuring back-translation quality, which cannot accurately capture the quality of the produced signs nor their usability in real-world settings. Understanding how distinctions in meaning are created in signed language may help develop a better evaluation method.

## 7.4 Animation

I consider the task of animating/rendering poses into 3D avatars or photorealistic videos out-of-scope for this thesis for brevity. Section 7.1 covers a simple approach for the task, as well as the background for this step (§7.1.2). I urge computer vision researchers to improve on this baseline method, develop stronger on-device animation models, and release them publicly.

# Part III

# Discussion and Implications

"if I have seen further it is by standing on the shoulders of giants."

— Isaac Newton

# Chapter 8

# Sign Language Translation Application (Moryossef, 2023c)

This chapter presents *sign.mt*, an open-source application pioneering real-time multilingual bi-directional translation between spoken and signed languages. Harnessing state-of-the-art open-source models, this tool aims to address the communication divide between the hearing and the deaf, facilitating seamless translation in both spoken-to-signed and signed-to-spoken directions.

Promising reliable and unrestricted communication, *sign.mt* offers offline functionality, crucial in areas with limited internet connectivity. It further enhances user engagement by offering customizable photo-realistic avatars, thereby encouraging a more personalized and authentic user experience.

Licensed under CC BY-NC-SA 4.0, *sign.mt* signifies an important stride towards open, inclusive communication. The app can be used, and modified for personal and academic uses. It features a translation API, fostering integration into a range of applications. However, it is by no means a finished product.

We invite the NLP community to contribute towards the evolution of *sign.mt*. Whether it be the integration of more refined models, the development of innovative pipelines, or user experience improvements. Available at `https://sign.mt`, it stands as a testament to what we can achieve together, as we strive to make communication accessible to all.

# 8.1   Motivation

Sign language translation applications are crucial tools for enabling communication between individuals who are deaf or hard of hearing and those who communicate through spoken language. However, the complexity of developing sign language translation applications goes beyond handling mere text. These applications must be able to process and generate videos, demanding additional considerations like compute capabilities, accessibility, usability, working with large files, and platform support.

*sign.mt*, standing for **Sign** Language **M**achine **T**ranslation, was conceived as a response to these challenges. Current research in the field of sign language translation is fragmented and somewhat nebulous, with different research groups focusing on various aspects of the translation pipeline or on specific languages. Moreover, the high costs associated with server-side deployment and the complexity of client-side implementations often deter the development of interactive demonstrations for newly proposed models.

By providing a comprehensive application infrastructure that integrates the essential features around the translation process, *sign.mt* serves as a dynamic proof-of-concept. It aims to streamline the integration of new research findings into the application, sidestepping the overhead typically associated with implementing a full-stack application. When a research group develops a new model or improves a pipeline, they can integrate their advancements into the app swiftly, focusing only on their model. This approach allows researchers to deploy the app in a branch, testing their models in a practical environment. If the license allows and the models show an improvement, they can contribute their models to the main codebase. This is the first tool of its kind, diverging significantly from closed-source commercial applications.

Further, *sign.mt* serves as a multilingual platform, thus unifying the fragmented research landscape. It enables the concurrent running of models from different research groups for the supported languages, providing users with state-of-the-art translation capabilities for each language. Through this, *sign.mt* not only enhances accessibility and communication but also fuels continuous innovation in sign language translation research.

## 8.2   Implementation

Sign language translation presents unique challenges that set it apart from text-based translation. While text-based translation operates entirely within the textual domain for both input and output, sign language translation involves cross-modal transformation – from text to video and vice versa. This demands distinct implementations not only in functionality but also in the user interface.

It is essential to emphasize that the specific models utilized within various pipelines are deliberately modular and interchangeable. Our current choice of models for each module or task is primarily opportunistic, driven by availability rather than performance metrics or user evaluations. The app serves as a dynamic orchestrator, seamlessly coordinating among these models to deliver an integrated user experience. The platform's design accommodates the likelihood that researchers may wish to experiment with different models or fine-tune existing pipelines, without being constrained by rigid implementation details.

### 8.2.1   Spoken-to-Signed Translation

For spoken-to-signed translation, the process begins with an input of spoken language text. Optionally, we allow audio input, which is first transcribed into spoken language text using on-device Speech-to-Text (STT) technology.

When the input language is unknown, this textual input undergoes Spoken Language Identification (using `cld3` (Salcianu et al., 2016)), which detects the language of the provided text. This is crucial for choosing the appropriate model for subsequent translation steps. Simultaneously, the text is optionally normalized (using `ChatGPT` (OpenAI, 2022)) - this includes fixing capitalization, punctuation, grammatical errors, or misspellings, which we have found to enhance the performance of subsequent translation stages.

The language-identified and potentially normalized text is then translated into SignWriting (Sutton, 1990). Here, our system leverages real-time client-side machine translation (Bogoychev et al., 2021) to translate the grammatical structures and lexicon of spoken languages into the visual-gestural modality of sign languages (Jiang et al., 2023a; Moryossef and Jiang, 2023).

The SignWriting output is then converted into a pose sequence (Inspired by Arkushin et al. (2023)), an ordered set of human poses that represent the signed sentence. This pose sequence is the input for the rendering engine, with three options: Skeleton Viewer (Minimalistic visualization of the skeletal pose (Moryossef et al., 2021a)) Human GAN (Pix2Pix (Isola et al., 2017; Shi et al., 2016) image-to-image model, generating a realistic human avatar video), and a 3D Avatar (Neural model to translate between pose positions and rigged rotations, generating a stylized 3D character performing the signs).

These different outputs provide users with a choice on how they prefer to view the translation, catering to a broad range of preferences and use cases. The skeleton viewer is useful for developers to see the raw output, as well as for low-compute users. The 3D Avatar is useful in mixed reality applications, where it can be integrated in the environment, and the Human GAN is useful for high-compute users, facilitating a natural interaction.

Through this pipeline (Figure 8.1), *sign.mt* is capable of real-time translation from spoken language audio (or text) into sign language video, further democratizing communication across modalities.

Currently, while we don't have a fully functional SignWriting to pose animation model, we have created a baseline model as an interim solution (Moryossef et al., 2023b). This model performs dictionary-based translation from the spoken language text directly to poses, bypassing the SignWriting stage. However, it's important to note that there are numerous common cases in sign languages that this baseline model cannot handle adequately yet. We have made the baseline model open-source, and it is available for further improvements and contributions from the community at https://github.com/ZurichNLP/spoken-to-signed-translation. We hope that this approach will stimulate further research and development in this area, allowing for the integration of more sophisticated and accurate models in future iterations of the application.

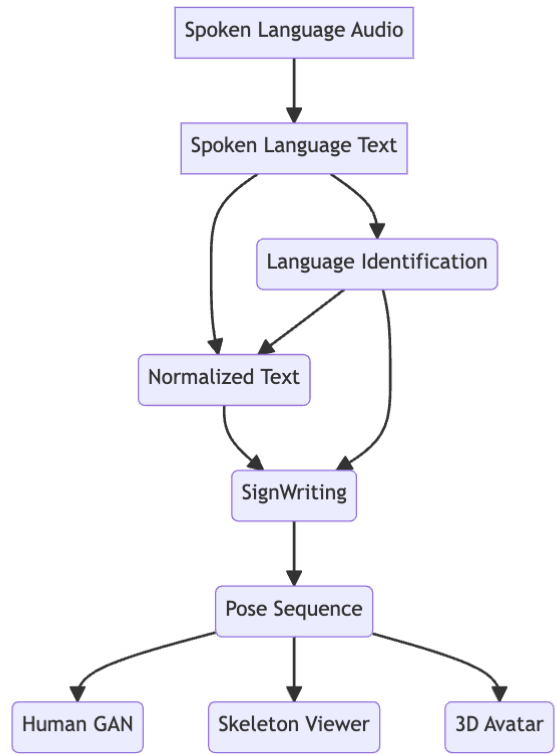

Figure 8.1: The Spoken-to-Signed translation pipeline.

## 8.2.2 Signed-to-Spoken Translation

For signed-to-spoken translation, the source is a video (either by the user uploading a pre-existing sign language video or using the camera to record a live sign language video). Our current pipeline takes the video, and using Mediapipe Holistic (Grishchenko and Bazarevsky, 2020) pose estimation extracts the full body pose from each frame.

This pose information is then fed into a Segmentation module (Moryossef et al., 2023a), which segments distinct signs within the continuous signing flow, as well as phrase boundaries. The segmented signs are subsequently lexically transcribed using SignWriting (Sutton, 1990), a comprehensive system for transcribing sign languages visually.

This SignWriting transcription serves as the textual input for the translation model, which translates it into corresponding spoken language text (Jiang et al., 2023a; Moryossef and Jiang, 2023). This text is then optionally converted into spoken language audio using on-device Text-to-Speech (TTS), providing an auditory output for the user.

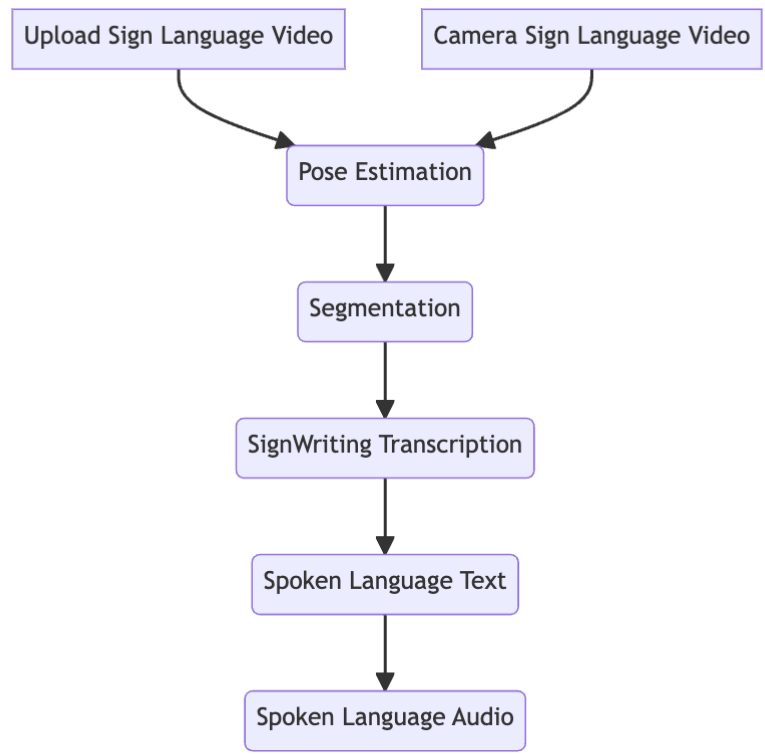

Figure 8.2: The Signed-to-Spoken translation pipeline.

Through this pipeline (Figure 8.2), *sign.mt* can take a sign language video and output corresponding spoken language text or audio in real-time. The offline functionality of the app ensures that this feature remains accessible even in areas with limited connectivity, provided that the models were pre-loaded on the device.

## 8.3 User Engagement

The impact of *sign.mt* can be measured by its widespread and consistent usage, highlighting the tremendous growth potential as the app continues to slowly improve.

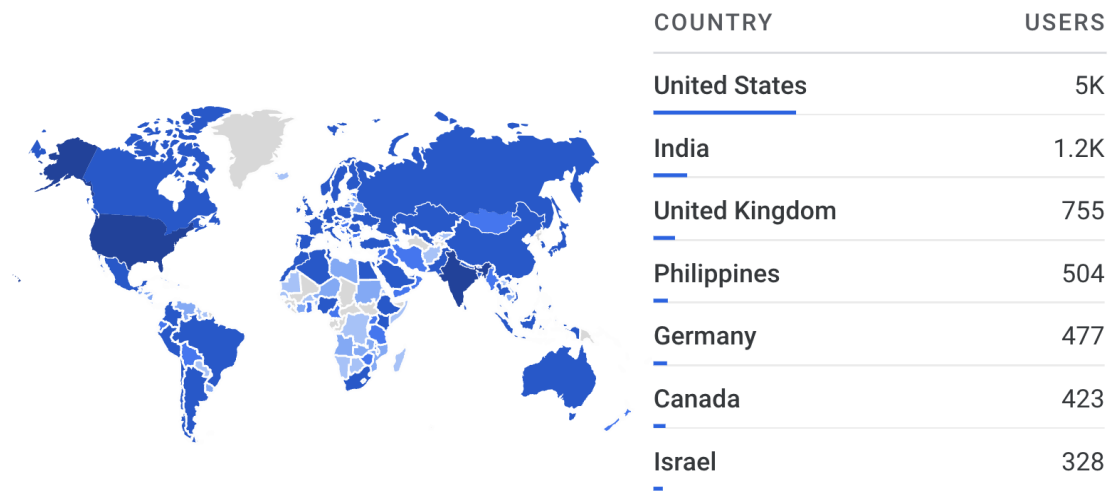

| COUNTRY | USERS |
|---|---|
| United States | 5K |
| India | 1.2K |
| United Kingdom | 755 |
| Philippines | 504 |
| Germany | 477 |
| Canada | 423 |
| Israel | 328 |

Figure 8.3: Distribution of *sign.mt* users across the world, over the last year.

Figure 8.3 depicts the global adoption of *sign.mt*, with users distributed across multiple countries. None of these top user countries are home to the core developer of the app.

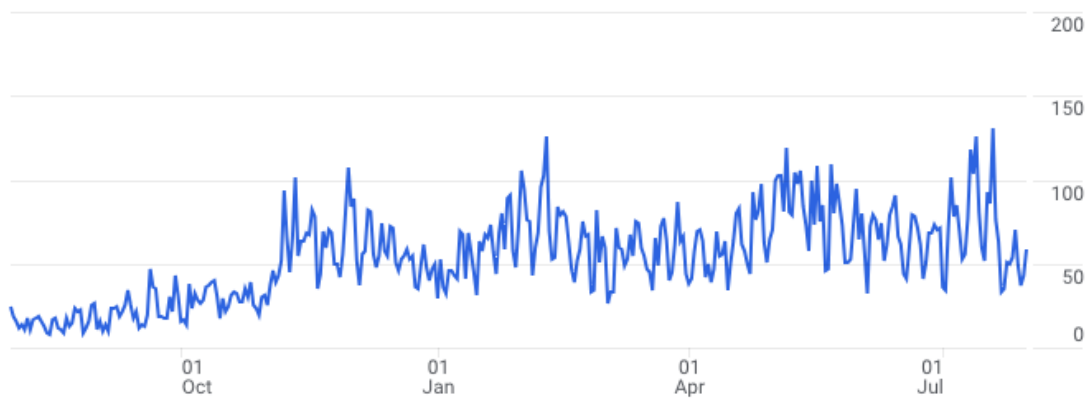

Figure 8.4: Growth of *sign.mt* users over the last year.

As shown in Figure 8.4, *sign.mt* demonstrates slow but consistent user growth (by Google Analytics), indicative of its reliability and sustained relevance.

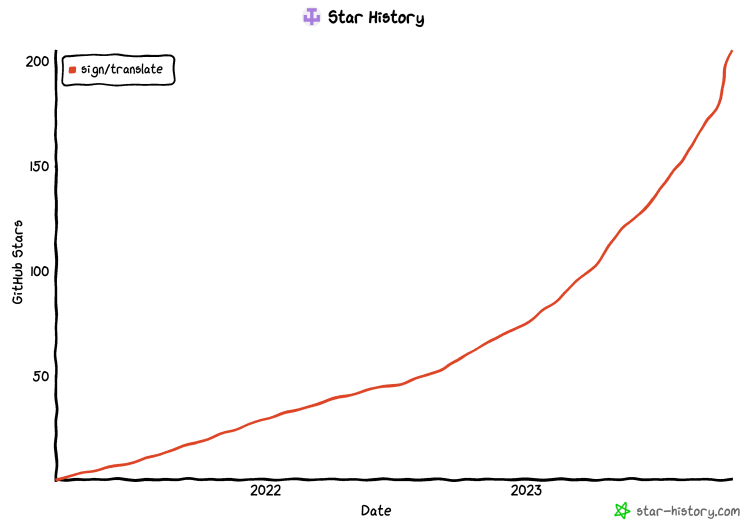

Figure 8.5: Number of stars for the repository over time.

Further validation of the community interest in *sign.mt* is evidenced by the increasing number of stars for its repository, reaching 151 stars as of August 1st, 2023 (Figure 8.5).

Public interest in *sign.mt* is further supported by Google Search Console metrics (Figure 8.6), showing a significant increase in impressions and clicks over the past six months: 3.75K clicks (up from 1.56K), and 106K impressions (up from 24.4K). Despite the absence of a marketing team and a single maintainer, *sign.mt* has managed to carve a niche for itself in the realm of NLP tools, reiterating its significance and impact.

## 8.4  Distribution

The code for *sign.mt* is openly accessible and available for contribution on GitHub at `https://github.com/sign/translate`, under CC BY-NC-SA 4.0. Open sourcing with a permissive license encourages the continuous refinement

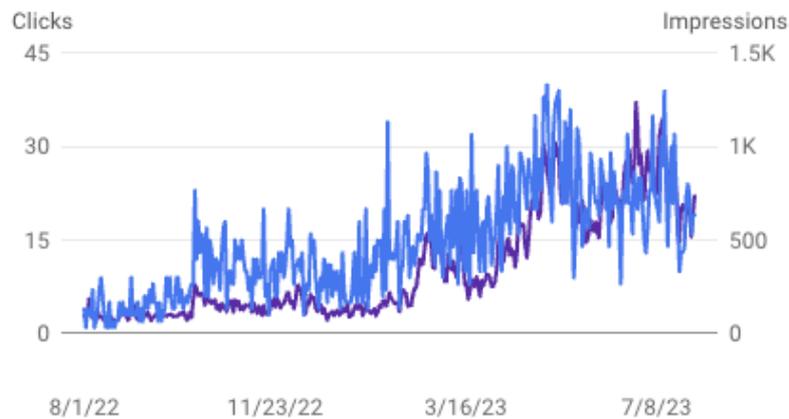

Figure 8.6: Google Search Console metrics showing increasing interest in *sign.mt*. (Clicks in blue)

and enhancement of the app through contributions from the wider developer and research communities.

The web application is freely accessible at `https://sign.mt`, designed with a responsive layout to cater to both desktop and mobile devices. Adhering to the design principles native to each platform, the application ensures an intuitive and user-friendly experience across all devices. With localization being a critical aspect of accessibility, the app interface supports 104 languages. Contributors can add their language or enhance the support for existing languages.

In addition to the web application, native builds for iOS and Android devices are also provided through the GitHub repository. While these are currently in development, the plan is to make them available to download on the respective app stores as they reach stability, thereby extending the reach of *sign.mt* to a wider audience.

## Limitations

As an evolving open-source project, *sign.mt* still faces several challenges and limitations. At present, the app does not provide complete support for every component of the translation pipeline. Notably, the SignWriting-to-pose ani-mation model does not currently exist, and instead, we use a simple dictionary lookup approach (Moryossef et al., 2023b). Although it serves as an interim solution, it is insufficient for handling signed languages. We eagerly anticipate and encourage contributions from the research community to fill this gap with more advanced models.

Although the app aspires to be a multilingual platform, the availability of models for different languages is currently fragmented. We rely on the research community to develop and contribute models for different languages. The support for each language, therefore, depends on the respective models available, leading to varying degrees of effectiveness across languages. For example, the SignWriting translation module works reasonably well for English/American Sign Language, German/German Sign Language and Portuguese/Brazilian Sign Language translations, and much worse for all other language pairs. Another example is the dictionary-based baseline only working on languages where dictionaries are available.

Due to the client-side deployment, we are restricted to using relatively smaller models. This inevitably leads to trade-offs in terms of translation accuracy and quality. While the offline functionality ensures accessibility in low connectivity areas, the constraint on model size is challenging.

The video processing components, including pose estimation and video rendering, are computationally intensive. This demands significant computational power, limiting the app's performance on devices with lesser computing capabilities. Optimizing these components further to ensure a smoother user experience across a wider range of devices is a challenge, often met with using lower-end models to achieve smoothness at the cost of accuracy.

Despite these limitations, *sign.mt* serves as a robust foundation upon which future advancements can be built. It continues to evolve in response to the feedback of the wider community, consistently striving towards the goal of facilitating accessible, inclusive communication.

# Chapter 9

# Implications for Spoken Languages (Moryossef, 2023b)

This chapter explores the critical but often overlooked role of non-verbal cues, including co-speech gestures and facial expressions, in human communication and their implications for Natural Language Processing (NLP). We argue that understanding human communication requires a more holistic approach that goes beyond textual or spoken words to include non-verbal elements. Borrowing from advances in sign language processing, we propose the development of universal automatic gesture segmentation and transcription models to transcribe these non-verbal cues into textual form. Such a methodology aims to bridge the blind spots in spoken language understanding, enhancing the scope and applicability of NLP models. Through motivating examples, we demonstrate the limitations of relying solely on text-based models. We propose a computationally efficient and flexible approach for incorporating non-verbal cues, which can seamlessly integrate with existing NLP pipelines. We conclude by calling upon the research community to contribute to the development of universal transcription methods and to validate their effectiveness in capturing the complexities of real-world, multi-modal interactions.

# 9.1 Introduction

Human speech is typically accompanied by a dynamic combination of co-speech gestures and facial expressions, together forming an integral part of human communication. These non-verbal cues, far from being random or merely accessory, provide additional layers of meaning, clarify intention, emphasize points, regulate conversation flow, and facilitate emotional connection. They enrich our interactions and help convey complex or nuanced information that words alone might not capture.

Co-speech gestures refer to the hand and body movements accompanying spoken discourse; they supplement verbal communication by offering additional information, such as object size or shape; they emphasize and make abstract concepts tangible, like gesturing upwards to signify an increase; they control the conversation flow, signaling a speaker's intent, inviting listener interaction, or showing that the speaker is in thought or pause; and lastly, they compensate for the limitations of spoken language, especially in high-stakes or noisy environments, by providing an alternative mode of conveying complex or nuanced information.

Facial expressions during speech significantly contribute to communication by indicating the speaker's emotions, and providing insight into their feelings about the topic; they can emphasize certain aspects of the discourse, with actions like raised eyebrows signifying surprise or importance; they offer social cues, with expressions like a smile suggesting friendliness or a serious look indicating sincerity; they help clarify verbal meaning, especially in ambiguous situations, for example, a confused expression might denote misunderstanding; finally, they enhance interpersonal connection by helping to build rapport, expressing empathy, and conveying cues of understanding and engagement; altogether, facial expressions, like gestures, add complexity and depth to verbal communications.

The field of Natural Language Processing (NLP) has become highly effective in understanding language directly from text. However, understanding speech, with its imperfect and noisy signals, remains a more complex challenge. Text-based language models have proven highly scalable, thanks largely to the compressible nature of text and its abundant availability in semi-anonymous forms. Yet, these models fundamentally ignore the rich layers of meaning added by non-verbal cues, a significant aspect of human communication. This means that while we have become adept at parsing text, we are missing out on the nuanced interplay of speech and gesture that characterizes in-person communication. Despite some promising work in generating co-speech gestures from audio (Ginosar et al., 2019; Bhattacharya et al., 2021; Liu et al., 2022), these gestures are often treated as accessory to speech rather than integral components, and thus, they do not always contribute the correct or intended information. As such, an understanding and integration of non-verbal cues remain an important frontier for further exploration in NLP.

Spoken language understanding, we propose, can benefit immensely from the advances in sign language processing. We advocate for the implementation of universal automatic gesture segmentation and transcription models that can transcribe co-speech gestures into textual input. This could be a pioneering step towards integrating the richness of non-verbal cues directly into the NLP models. By including transcribed gestures, the models would bridge the blind spots in spoken language understanding. This is a bidirectional process; Just as spoken language models can learn from sign language processing, the insights from the transcription of spoken language gestures can also inform and enhance sign language processing, due to iconicity, and metaphors. Ultimately, this holistic approach would result in a more nuanced and comprehensive understanding of human communication, bringing us closer to the complexities and richness of real-world, multi-modal interactions.

## 9.2 Stereotypical Language Variation

Non-verbal forms of communication are subject to significant cultural variability, shaped by a complex interplay of historical, societal, and cultural factors.

In Mediterranean cultures, non-verbal communication is prevalent and vi-brant. People in this region often use expressive gestures and maintain close personal space when communicating. Italian, for instance, is renowned for its extensive use of gestures. Italians often use their hands and bodies expressively to illustrate their points or emotions, and there is a broad range of specific gestures that carry particular meanings, often comprehensible even without accompanying speech.

In contrast, Japanese communication tends to incorporate fewer and more subtle non-verbal cues. A bow, a nod, or a slight tilt of the head can convey a myriad of meanings depending on the context, demonstrating respect, agreement, or understanding. Meanwhile, in Nordic cultures, such as Swedish or Finnish, non-verbal cues are typically used sparingly. The communication style tends to be direct and understated, with less emphasis on gestures and more focus on verbal content.

Overall, these stereotypical examples highlight the diverse ways in which languages around the world incorporate non-verbal cues into communication. This diversity emphasizes the importance of cultural understanding and sensitivity in interpreting and engaging in cross-cultural communication research, and data collection and annotation.

## 9.3 Motivating Examples

Non-verbal cues can act to affirm and reinforce the spoken words, thereby strengthening the communicated message. They can also undermine the verbal message, creating a contradiction between what is being said and the speaker's true intent or feelings. For NLP research to understand speech, it can not rely solely on audio (or textual transcription) to understand the intent of the speaker.

For example, saying 'Perfect' while making a circle with the thumb and index finger often emphasizes approval and satisfaction. Similarly, nodding while saying 'Yes' reinforces affirmation, underscoring the speaker's understanding or agreement. On the other hand, saying 'OK' while rolling one's eyes, can suggest that the speaker doesn't find the situation truly satisfactory, despite the verbal agreement. Similarly, stating "I'm not mad" while frowning or clenching fists suggests that the speaker is upset, contradicting their verbal assertion.

### 9.3.1  Sentiment Analysis

To demonstrate the limitations of text-based sentiment analysis, consider the following hypothetical dialogue between a couple, where the man is utilizing passive-aggressive communication. In each turn, we also present the sentiment score as predicted by the Google Cloud Natural Language API Demo[1], where scores range between $[-1, 1]$.

```
Woman: How is it going? (0)
Man:   I am fine. (0.74)
       [crosses his arms]

Woman: Did you enjoy dinner? (0.55)
Man:   It was fine. (0.92)
       [avoids eye contact,
       lips pressed tightly]

Woman: Is something wrong?
       You seem distant. (-0.78)
Man:   No, nothing's wrong. (0.53)
       [shakes his head slightly,
       exhales loudly]

Woman: Are you sure? (0)
Man:   I said I'm fine. (0)
       [rolls eyes, turns away]
```

While all the man's responses register as neutral to positive, his body language—avoiding eye contact, pressing his lips tightly together, shaking his head, exhaling loudly, rolling his eyes, and turning away—signals that he may actually be upset, frustrated, or disengaged. By neglecting body language and other contextual clues, current models miss out on a significant layer of human communication, particularly in emotionally charged or complex dialogues. Such a holistic approach could provide a more nuanced and accurate understanding of the emotional context, thus enriching machine-human interactions.

---

[1]https://cloud.google.com/natural-language

### 9.3.2    Machine Translation

While existing in many other languages, Italian stereotypically gives us many examples of gestures conveying meaning, where the verbal part is often dropped altogether, making it even more similar to signed languages.

Table 9.1 showcases a toy example of a conversation between two Italians using only gestures, without speech. It is transcribed using SignWriting (Sutton, 1990) to demonstrate that anonymous non-verbal transcription can be done in a low-bandwidth manner and that it can be reproduced and understood by people trained at reading SignWriting.

## 9.4    Methodology

Machine learning techniques that focus solely on text have gained predominance due to several key factors: the abundance of readily available text data, the potential for semi-anonymous data collection and processing, the very high bandwidth-to-overhead ratio as a word consumes only a few bytes compared to kilobytes or more for a second of speech or video, and the ease with which text can be viewed, edited, and corrected.

Previous efforts have attempted to include other modalities like images (Razavi et al., 2019), videos (Yan et al., 2021), or audio through the use of techniques like VQ-VAEs (van den Oord et al., 2017). However, these approaches often significantly increase the context size, are not transferable across different systems, and generally require the original signal (like a video) to be sent for processing.

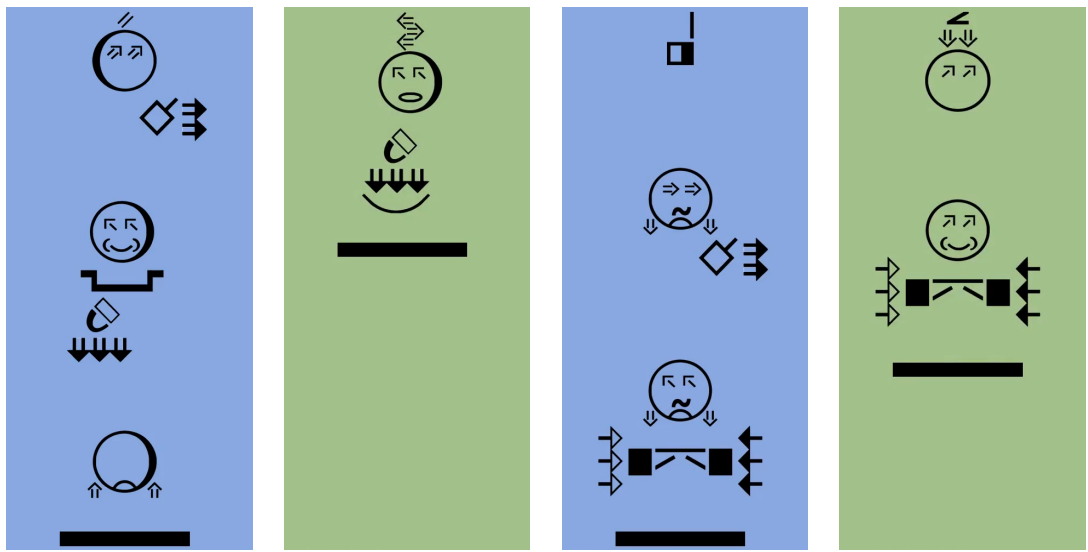

| wth with these two?? | I know! What the hell? | Did you hear they're together?? | Oh yeah. They're definitely together! |

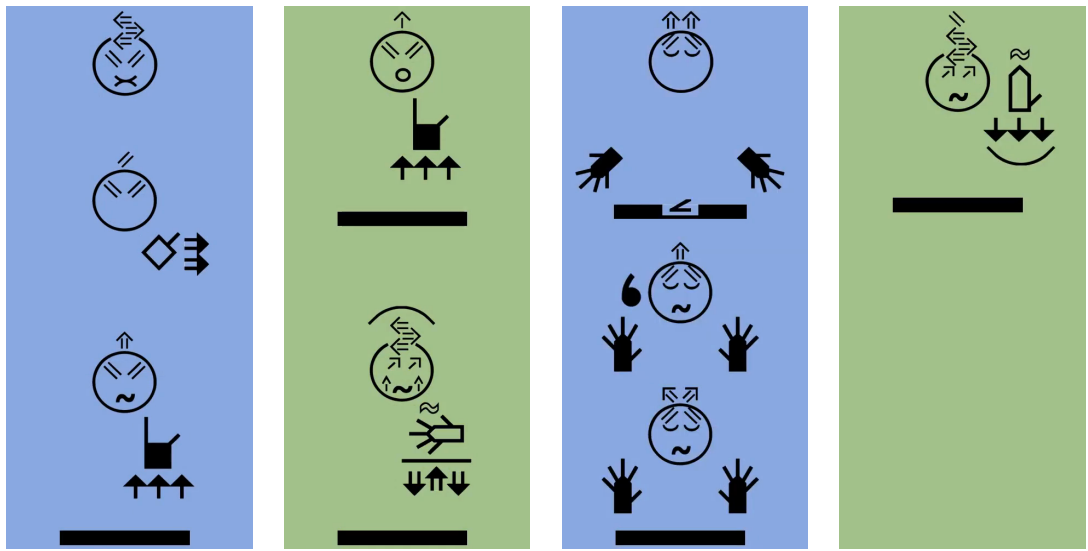

| I shouldn't be saying this but... she's cheating on him | She's cheating on him?? Woahhh | Yeah! But it's none of my business | I can't believe it |

Table 9.1: "How to gossip in Italian" by the Pasinis, transcribed in Sutton SignWriting by Sutthikhun Phaengphongsai https://www.youtube.com/watch?v=7V-GniCQFkE, demonstrating a conversation between two Italians using only gestures, without speech.

In contrast, our proposal offers a more flexible, universal, and efficient way to incorporate non-verbal cues directly as text.

### 9.4.1  Proposal

We propose adopting a universal transcription system for body language, much like the written system used for spoken languages. This system would transcribe gestures, facial expressions, and other non-verbal cues into textual form. The advantages of this approach are numerous:

**Flexibility in Transcription**  Different programs can decide on their own transcription methods, taking into account local variations and context.

**Computational Efficiency**  Text-based methods require significantly lower computational resources compared to image or video processing. (Notoriously, GPT-4 was released without image upload support, since inference on a single image takes upwards of 20 seconds)

**Compatibility with Existing Models**  As the body language would be transcribed into discrete tokens, it can fit seamlessly into existing large language models without any modification.

**Anonymity**  Transcription acts as a form of biometric anonymization, removing the need to share actual video or images.

**Explainability**  The textual transcription provides a more transparent input, making the language modeling process more understandable.

**Seamless Integration**  The proposed methodology does not require any significant changes to existing NLP pipelines. It simply acts as an additional layer of data for better understanding and disambiguation. You can include it, or not.

### 9.4.2   Implementation

To successfully integrate non-verbal cues when processing spoken language, we advocate the following steps:

1. Capture both video and audio during speech.

2. Use sign language segmentation models to identify boundaries of individual gestures.

3. Transcribe these gestures into a textual notation system like SignWriting.

4. Use speech-to-text models to transcribe the spoken language, identifying the boundaries where each word is expressed.

5. If word boundaries are not directly accessible, a re-alignment model can be used to approximate these boundaries.

6. Combine both speech and gesture transcriptions into a single text string, where gestures provide additional context to the spoken words.

This approach can be thought of as analogous to incorporating additional context, such as gender, into machine translation (Moryossef et al., 2019). By training on a large dataset that includes unmarked sentences, the model may develop certain biases. Introducing a smaller dataset with contextual information can help the model learn correlations between language and specific contexts. During inference, one has the option to either provide just the text for a more generalized output or include additional contextual tags for a more accurate and targeted output.

## 9.5   Conclusions

This chapter underscores the fundamental role of non-verbal cues, such as co-speech gestures and facial expressions, in human communication. While strides have been made in the realm of Natural Language Processing for understanding textual content, a holistic approach that integrates the rich layers of non-verbal information is significantly lacking. This shortfall not only hampers the comprehension of spoken language but also limits our ability to construct nuanced, context-aware NLP models.

The key to advancing in this frontier may lie in borrowing techniques and insights from sign language processing. We advocate for the adaptation and implementation of universal automatic gesture segmentation and transcription models that can transcribe co-speech gestures into textual input. Such an approach would be a pivotal step in bridging the gap between text-based and real-world, spoken interactions, thereby enriching both the scope and applicability of NLP models.

When processing spoken language content, researchers should adopt a more holistic lens, one that goes beyond words and phrases to also consider nonverbal cues. Existing universal segmentation and transcription models used in sign language can serve as invaluable resources for this purpose, as they offer the ability to transcribe gestures directly as text.

We call upon researchers in spoken language processing to contribute to the development of universal gesture transcription methods. Furthermore, we encourage the academic community to construct challenge sets specifically tailored to validate the utility of these transcription methods in capturing the complexities of non-verbal communication. These steps are not merely supplementary but are central to achieving a more comprehensive understanding of human communication in its full richness and complexity.

# Chapter 10

# Conclusions

This thesis introduces a novel pipeline to including signed languages in Natural Language Processing. Preliminary work showed that the currently predominant gloss-based pipeline is problematic, and contemporary research shows that we are far from solving this problem in an end-to-end manner. Therefore, **this thesis proposes the introduction of lexical sign language notations as the pivot between the video and text modalities**. It addresses both directions - *signed-to-spoken* translation (Chapter 6) and *spoken-to-signed* translation (Chapter 7), and addresses the various sub-tasks of each pipeline to some extent. We believe that *solving* each one of these sub-tasks requires deeper studies and entire theses can be written on each.

While this thesis does not, by all means, solve the task of sign language translation, we believe that it introduces the first viable pipeline to address both directions, not only in a real-time multilingual setting. This pipeline relies on little, high-quality sign language transcription data in SignWriting, which since it is a universal (multilingual) representation of signed languages, weakens the necessity of large annotated datasets for every signed language. We believe that the implications of this thesis go beyond signed languages, into the realm of spoken language processing and even action recognition. Exhaustive movement transcription opens new and exciting avenues in all of these fields, the same way normalized spoken language writing systems have revolutionized

Natural Language Processing.

We call for a complete separation of sign language processing between the fields of Natural Language Processing and Computer Vision. In Natural Language Processing, researchers should explore the translation from SignWriting to spoken language text and vice-versa, dealing directly with the language content as written. They should be aware that transcription of real signing data is noisy and does not necessarily map into well-formed dictionary forms. In Computer Vision, researchers should explore the transcription of sign language videos into SignWriting, without understanding the meaning, in a language-agnostic setting. Further, they should explore the animation of SignWriting sequences into videos, with the use of 3D avatars or photo-realistic models.

## Caution for Future Researchers

In this thesis, we propose using SignWriting as a pivot between spoken language text and sign language video, in both directions. We take a specific pipeline approach that utilizes more intermediate steps in order to explore this pipeline as it is possible in our times. However, we believe that as this pipeline gets used, it will make more data available (e.g., transcribing existing sign language videos will create more parallel data between spoken language text and SignWriting), and allow for end-to-end approaches in SLP. We encourage these improvements and the reduction of steps in the pipeline.

Ideally, we believe that end-to-end sentence-level video-to-SignWriting transcription systems, and SignWriting-to-video animation systems will prevail over the inclusion of pose estimation and sign segmentation, but, we caution against treating the SignWriting pivot the same way, and attempting end-to-end video-to-text and text-to-video systems. These systems will be fundamentally hard to evaluate, challenging to modify, and language-dependent. Breaking away from the SignWriting pivot would also break the separation between Natural Language Processing and Computer Vision, and could lead to sub-par research.

# Chapter 11

# תקציר

שפות סימנים משמשות כאמצעי תקשורת חיוני עבור מיליוני אנשים חירשים וכבדי שמיעה ברחבי העולם.  תוך שימוש במודאליות חזותית־תנועתית, הם מעבירים מבנים לשוניים מורכבים באמצעות ניסוחים ידניים בשילוב עם אלמנטים לא ידניים כמו הבעות פנים ותנועת גוף. למרות העושר הלשוני והחשיבות התרבותית שלהן, שפות סימנים נדחקו לעתים קרובות לשוליים על ידי ההתקדמות האחרונה בטכנולוגיות של בינה מלאכותית ממוקדת טקסט, כגון תרגום מכונה ומודלים של שפות גדולות. דחיקה זו מגבילה את הגישה לטכנולוגיות אלו עבור אוכלוסייה משמעותית, ומשאירה אותם מאחור בהתקדמות בבינה מלאכותית מבוססת שפה.

עיבוד שפת סימנים הוא תחום בינתחומי המורכב מעיבוד שפה טבעית וראייה ממוחשבת. הוא מתמקד בהבנה חישובית, תרגום והפקה של שפות סימנים. גישות מסורתיות הוגבלו לעתים קרובות על ידי שימוש במערכות מבוססות גלוס שהן ספציפיות לשפה וגם לא מתאימות ללכידת הטבע הרב־ממדי של שפת הסימנים. מגבלות אלו הפריעו להתפתחות טכנולוגיה המסוגלת לעבד שפות סימנים ביעילות.

תזה זו שואפת לחולל מהפכה בתחום עיבוד שפת הסימנים על ידי הצעת פרדיגמה פשוטה שיכולה לגשר על פער טכנולוגי זה.  אנו מציעים להשתמש ב־SignWiring, מערכת תמלול אוניברסלית של שפות סימנים, כדי לשמש כקישור מתווך בין המודאליות החזותית־מחוותית של שפות הסימנים לבין ייצוגים לשוניים מבוססי טקסט.

בניגוד לגישות מבוססות גלוס, הפרדיגמה שלנו באמצעות SignWiring נועדה ללכוד במדויק את ההיבטים הרב־ממדיים והבלתי תלויים בשפה של שפות סימנים. זה מאפשר יצירת מסגרת מאוחדת וניתנת להרחבה שיכולה להכיל את המגוון הלשוני העשיר שנמצא בשפות סימנים שונות ברחבי העולם.

אנו תורמים ספריות ומשאבים בסיסיים לקהילת עיבוד שפת הסימנים, ובכך

מאפשרים חקירה מעמיקה יותר של משימות התרגום וההפקה של שפת הסימנים. משימות אלו מקיפות את התרגום של שפת הסימנים מוידאו לטקסט בשפה דבורה ולהיפך. באמצעות הערכות אמפיריות, אנו מבססים את היעילות של שיטת התמלול שלנו כציר לאפשר מחקר מהיר וממוקד יותר, שיכול להוביל לתרגומים טבעיים ומדויקים יותר במגוון שפות.

הפרדיגמה שלנו קובעת גבול ברור בין עיבוד שפה טבעית לראייה ממוחשבת בהקשר הרחב יותר של עיבוד שפת הסימנים. חלוקה זו משקפת את ההפרדה הקיימת בין עיבוד שפה טבעית ועיבוד אותות בתחום טכנולוגיות השפה הדבורה. על ידי כך, אנו פותחים את הדלת למאמצי מחקר מדוייקטים יותר בכל תת־דיסציפלינה, ובכך מעשירים את המערכת האקולוגית של טכנולוגיות ומתודולוגיות הזמינות עבור עיבוד שפת הסימנים.

האופי האוניברסלי של הפרדיגמה המבוססת על התמלול שלנו גם סולל את הדרך ליישומים רב־לשוניים בזמן אמת בעיבוד שפת הסימנים, ובכך מציע גישה כוללת וגישה יותר לטכנולוגיית שפה. זהו צעד משמעותי לקראת נגישות אוניברסלית, המאפשר טווח רחב יותר של טכנולוגיות שפה מונעות בינה מלאכותית לכלול את קהילת החירשים וכבדי השמיעה.

לסיכום, עבודת גמר זו מציגה גישה חדשה לעיבוד שפת הסימנים, כזו שמטרתה לקבוע סטנדרט חדש לטכנולוגיות שפה כוללניות, בזמן אמת ורב־לשוניות. על ידי גישור על הפער הקיים בין בינה מלאכותית ממוקדת טקסט לעולם החזותי־תנועתי של שפת הסימנים, אנו תורמים באופן משמעותי להפיכת בינה מלאכותית מבוססת שפה לנגישה אוניברסלית.

# תוכן עניינים

# תודות

לאלו שתמיד האמינו בי,
ולאלה שאהבתי ואיבדתי.

# עיבוד שפת הסימנים
# מרובה שפות בזמן אמת

## עמית מוריוסף
### המחלקה למדעי המחשב